\documentclass[times, review, 10pt]{elsarticle}

\biboptions{sort&compress} 

\usepackage[T1]{fontenc}
\usepackage{lmodern}
\usepackage{microtype}

\usepackage{amsmath,amssymb,amsfonts}
\usepackage{amsthm}
\usepackage{mathrsfs}
\usepackage{physics}
\usepackage{array} 
\usepackage{adjustbox} 

\usepackage{xcolor}

\usepackage{graphicx}
\usepackage{subcaption}
\usepackage{booktabs}   
\usepackage{multirow}
\usepackage{makecell}
\usepackage{tabularx}
\usepackage{longtable}
\usepackage{lscape}
\usepackage{rotating}

\usepackage{algorithm}
\usepackage{algorithmicx}
\usepackage{algpseudocode}

\usepackage{listings}
\usepackage{textcomp}  

\usepackage{float}      

\usepackage{hyperref}

\journal{}

\begin{document}

\begin{frontmatter}

\title{Continuous ageing trajectory representations for knee-aware lifetime prediction of lithium-ion batteries across heterogeneous datasets}

\author[1]{Agnieszka Pregowska\corref{cor1}}
\author[1]{Stefan Marynowicz}

\affiliation[1]{
	organization={Institute of Fundamental Technological Research, Polish Academy of Sciences},
	addressline={Pawinskiego 5B},
	city={Warsaw},
	postcode={02-106},
	country={Poland}
}

\cortext[cor1]{Corresponding author: Agnieszka Pregowska, e-mail: aprego@ippt.pan.pl}

\begin{abstract}
Accurate assessment of lithium-ion battery ageing is challenged by cell-to-cell variability, heterogeneous cycling protocols, and limited transferability of data-driven models across datasets. In particular, robust identification of degradation transitions, such as the knee point, and reliable early-life prediction of remaining useful life (RUL) remain open problems.

This study proposes a unified framework for battery ageing analysis based on continuous representations of voltage-capacity and capacity-cycle trajectories learned from heterogeneous public datasets (NASA, CALCE, ISU-ILCC). The continuous formulation enables consistent extraction of degradation descriptors, including curvature, plateau length and knee-related metrics, while reducing sensitivity to dataset-specific discretisation.

Across more than 250 cells, statistically significant correlations between knee onset and end-of-life (Pearson 0.75–0.84) are observed. Additional early-life analysis confirms that knee-related features retain predictive value when estimated from partial trajectories. Early-life models provide increasingly stable RUL predictions as the number of observed cycles increases, 
with meaningful predictive performance emerging within the first 5–20 cycles and remain robust under cross-dataset domain shift.

The framework integrates continuous modelling, feature extraction and uncertainty-aware prediction, providing an interpretable and dataset-consistent approach demonstrating robustness across heterogeneous dataset types. Compared with conventional discrete or feature-based methods, the proposed representation reduces sensitivity to sampling resolution and improves cross-dataset consistency. The study is limited to laboratory-scale datasets and capacity-based end-of-life definitions.
\end{abstract}


\begin{graphicalabstract}
	\centering
	\includegraphics[width=\textwidth]{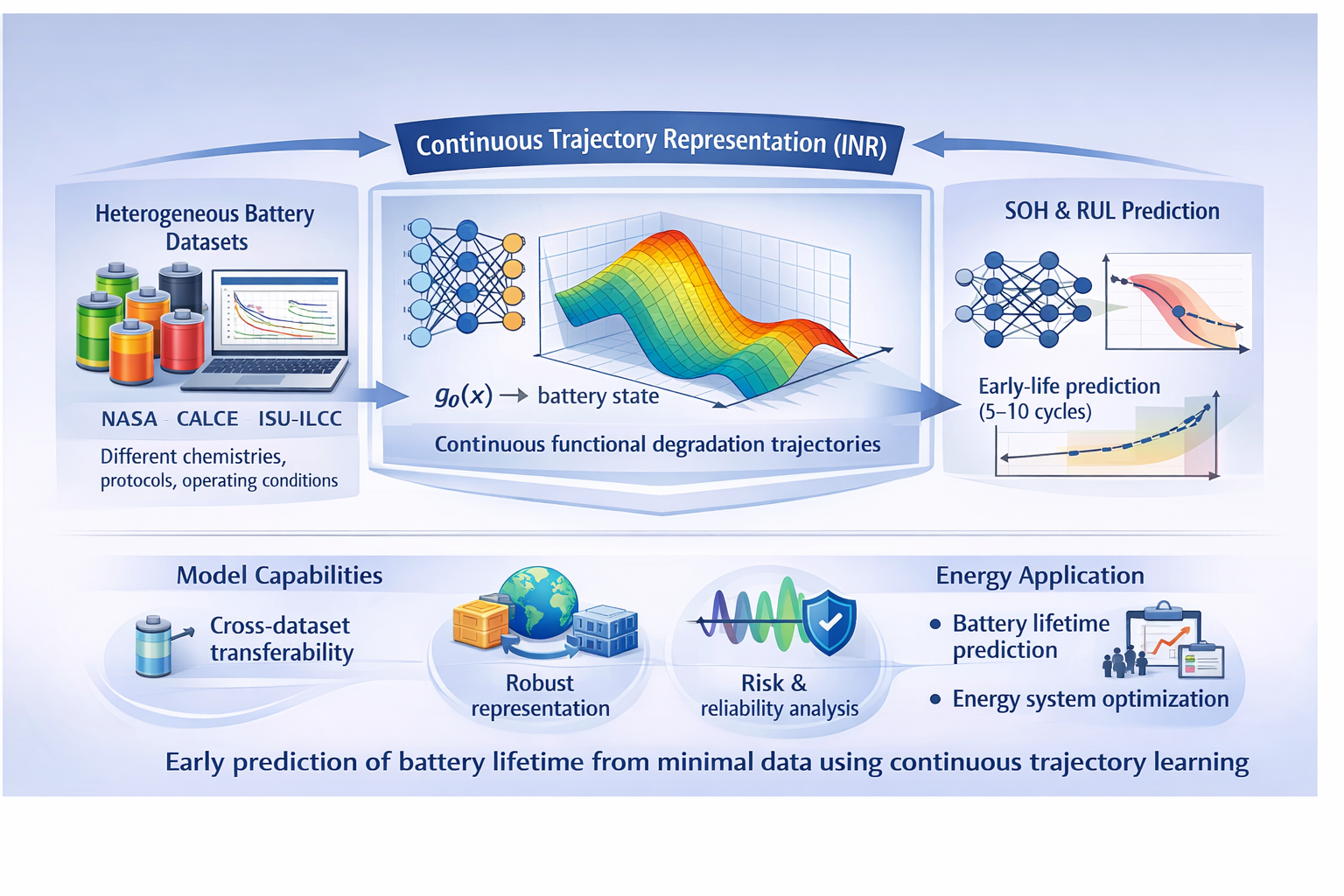}
\end{graphicalabstract}

\begin{highlights}
	\item Continuous representations unify voltage-capacity and capacity-cycle ageing trajectories across datasets
	\item Knee-point metrics show statistically robust correlation with end-of-life (Pearson up to 0.84)
	\item Early-life RUL prediction becomes informative within the first 5--20 cycles
	\item The framework demonstrates robustness under cross-dataset domain shift
	\item Reliability analysis reveals consistent Weibull ageing behaviour across cell populations
\end{highlights}


\end{frontmatter}

\section{Introduction}

The rapid deployment of lithium-ion batteries in electric vehicles, stationary storage and consumer electronics has intensified the need for reliable and interpretable ageing models \cite{vonBulow2023,Hu2025}. Despite extensive research, predicting battery degradation remains challenging due to the interplay of multiple mechanisms, including loss of cyclable lithium, impedance growth and thermal effects, which evolve nonlinearly over time \cite{Guangxu2022,Nazeeruddin2025}. Accurate estimation of state of health (SoH) and remaining useful life (RUL) is therefore essential for safe operation and lifecycle optimisation \cite{Chen2023}.

Existing approaches to battery ageing modelling can be broadly divided into physics-based and data-driven methods \cite{Pregowska2022}. Physics-based models, such as equivalent circuit and electrochemical models, provide physical interpretability but are often limited by parameter identifiability and computational complexity \cite{Hu2012,Jokar2016}. Data-driven methods, including machine learning and deep learning models, offer flexibility but typically rely on discretised time series or hand-crafted features, which limits transferability across datasets and operating conditions \cite{Song2025,Wei2023}.

A key limitation of current approaches is the lack of a unified representation that captures degradation trajectories consistently across heterogeneous datasets \cite{dosReis2021,Song2024}. In particular, the identification and exploitation of degradation transitions, such as the knee point, remain insufficiently understood, especially in cross-dataset settings \cite{Diao2019,You2023}.

To address these challenges, this study introduces a continuous representation framework for lithium-ion battery ageing based on implicit neural representations \cite{sitzmann2020,tancik2020}. By modelling voltage-capacity and capacity-cycle trajectories as continuous functions, the proposed approach enables consistent extraction of degradation descriptors, including curvature, plateau length and knee-related metrics, while reducing sensitivity to dataset-specific preprocessing.

The framework is evaluated on multiple public datasets (NASA, CALCE, ISU-ILCC) and supports early-life RUL prediction, cross-dataset generalisation and reliability-based lifetime analysis \cite{Madani2025,Mouais2021}. Rather than focusing on architectural novelty, the contribution lies in providing a unified, interpretable and transferable representation of battery ageing trajectories.

The main contributions of this study are fourfold. First, we introduce a unified continuous representation of ageing trajectories across heterogeneous public datasets. Second, we extract knee-related and trajectory-shape descriptors in a consistent functional form. Third, we demonstrate that these descriptors retain predictive value for early-life forecasting under domain shift. Fourth, we integrate continuous trajectory modelling with reliability analysis within a single framework. Compared with conventional discrete or hand-crafted feature-based approaches, the proposed framework is less sensitive to dataset-specific preprocessing and sampling differences, thereby enabling more consistent comparison across heterogeneous experimental protocols.

\section{Materials and Methods}

\subsection{Implicit Neural Representations}

Implicit Neural Representations constitute a class of continuous, coordinate-based neural functions that have recently emerged as a powerful alternative to classical interpolation and regression models \cite{park2019,mescheder2019,sitzmann2020,tancik2020,mildenhall2022}.
Instead of predicting over a discrete grid or relying on predefined basis expansions, an INR directly parameterises an unknown signal, such as a voltage curve, aging trajectory, or multidimensional electrochemical response, as a differentiable function

\begin{equation}
	g_{\theta} : \mathbb{R}^{d} \rightarrow \mathbb{R}^{c}, \quad \theta \in \mathbb{R}^{p}
\end{equation}

where the input coordinate $\mathbf{x} \in \mathbb{R}^{d}$ encodes the model domain (e.g., time, current, temperature, cycle index, or auxiliary metadata), and the output $\mathbf{y} \in \mathbb{R}^{c}$ corresponds to the reconstructed electrochemical response (voltage, derivative \emph{dV/dQ}, capacity-normalised features, or latent space parameters). The function $g_{\theta}$ is implemented as a multilayer perceptron whose trainable parameters~$\theta$ implicitly encode the full continuous signal \cite{Pregowska2026}. 

To further formalise the continuous representation, we consider the INR model
as a function approximator trained by minimising a data-fitting objective over
observed electrochemical measurements:
\begin{equation}
	\mathcal{L}(\theta) = \sum_{i=1}^{N} 
	\left\| g_{\theta}(\mathbf{x}_i) - y_i \right\|^2,
\end{equation}
where $\{(\mathbf{x}_i, y_i)\}_{i=1}^{N}$ denotes the set of coordinate–response
pairs obtained from battery cycling data. This formulation allows the model to
learn a continuous mapping from the input domain (e.g., capacity or cycle index)
to the measured signal (e.g., voltage or normalised capacity).

A key advantage of this representation is that differential operators can be
applied directly to the learned function. For example, the first and second
derivatives with respect to capacity $Q$,
\begin{equation}
	\frac{\partial V}{\partial Q}, \qquad
	\frac{\partial^2 V}{\partial Q^2},
\end{equation}
can be computed analytically via automatic differentiation. These quantities
provide physically interpretable descriptors such as incremental capacity,
curvature and inflection points, which are strongly linked to electrochemical
phase transitions and degradation mechanisms.

Unlike grid-based interpolants, polynomial expansions, or kernel regressors, INRs operate in continuous coordinate space and do not require predefined sampling locations. Their capacity to represent high-frequency and nonlinear behaviour depends not on the discretisation, but on the network architecture (e.g., sinusoidal activations in SIREN networks or random Fourier feature mappings). This makes them well suited for electrochemical systems, where voltage curves exhibit sharp transitions (plateau boundaries, phase-change regions, knee points) and strongly nonlinear aging dynamics. Furthermore, INRs naturally support: computation of derivatives (e.g., \(\partial V / \partial Q\), curvature, or rate sensitivity), dense reconstruction from sparse or noisy measurements without resampling artefacts, the parameters~$\theta$ can be interpreted as a compact ``fingerprint'' of the cell's electrochemical state, training across heterogeneous datasets helps the network learn an invariant representation of the underlying physics, improving robustness under domain shift. From a scientific machine learning perspective, INRs complement physics-informed neural networks and neural ODEs by providing flexible, continuous surrogates for strongly nonlinear electrochemical dynamics, while remaining agnostic to a specific partial differential equation model \cite{Dietrich2025}.

While the INR model is data-driven, partial physical consistency is ensured through the structure of the training data and the choice of target variables. In particular, monotonic capacity degradation and bounded voltage ranges are implicitly enforced by the observed measurements.

Furthermore, the extracted trajectory descriptors, such as curvature, plateau length and slope, are directly interpretable in terms of underlying electrochemical processes, including SEI growth, loss of active material and transport limitations. This provides a link between data-driven modelling and physically meaningful degradation mechanisms.

From a functional perspective, the INR representation can be interpreted as a neural field that approximates an underlying smooth degradation manifold. Assuming sufficient regularity, the learned function $g_{\theta}$ belongs to a Sobolev space $W^{k,2}$, which guarantees the existence of weak derivatives up to order $k$. This property is particularly relevant for battery ageing analysis, as it ensures stability of derivative-based features under noise and irregular sampling, which are common in experimental datasets.

\section{State-of-health and remaining-useful-life models}
\label{sec:soh_rul_models}
State-of-health estimation and remaining-useful-life (RUL) prediction are formulated as supervised learning problems on the harmonised per-cycle feature set described in  Section~\ref{sec:preprocessing}. Let $\mathbf{x}_k$ denote the cycle-level feature vector and $y_k$ the corresponding target, either $\mathrm{SOH}_k$ or the cycle-based RUL defined as  $y_k = \mathrm{EOL} - k$. Both regression tasks are trained and evaluated under a temporal cross-validation scheme to avoid leakage from future degradation trends, consistent with recommended methodology for ageing prognostics \cite{Meng2019,Xiao2024,Berecibar2016}. This design follows recent recommendations for battery prognostics benchmarks, which emphasise temporal splits and realistic forecasting horizons \cite{Madani2025}. We evaluate a family of models representing complementary modelling approaches. Linear and polynomial baselines  provide interpretable reference points for monotonic capacity fading, following prior study in  early-life forecasting \cite{Attia2020,Severson2019}. These models assume smooth degradation trajectories and act as lower bounds on achievable performance. Random Forests (RF) are widely used in battery prognostics due to their robustness to noise, nonlinear relationships and feature interactions \cite{Chen2018}. They require minimal preprocessing and produce informative feature importance scores. Together with feature importance from RF and the INR-derived shape descriptors, this yields a partially interpretable pipeline that aligns with recent efforts towards transparent SOH forecasting \cite{vonBulow2023}. Multi-layer perceptrons (MLP) provide a flexible non-parametric function approximator capable of modelling subtle nonlinear ageing effects such as knee-onset dynamics, temperature interactions and current-induced acceleration \cite{Al-Rahamneh2025}. Models use early stopping based on validation loss to mitigate overfitting. Implicit Neural Representations directly regress SOH or RUL from continuous coordinate inputs, including cycle index, temperature and extracted shape descriptors. Their ability to learn smooth, differentiable degradation fields makes them attractive for early RUL prediction and for linking degradation trajectories across heterogeneous datasets \cite{sitzmann2020, tancik2020,mildenhall2022}. For each model, we report the mean and standard deviation of regression metrics including root mean squared error (RMSE),  Mean absolute error (MAE), Mean Absolute Percentage Error (MAPE) and the coefficient of determination $R^2$ across time-based folds. This design ensures the tasks reflect true forecasting difficulty rather than interpolation.

\subsection{Data splitting and evaluation protocol}

To ensure realistic evaluation and avoid information leakage, we employ multiple evaluation strategies 
depending on the task. For early-life prediction, models are trained exclusively on features derived 
from the first $N \in \{5,10,20\}$ cycles and evaluated on held-out cells. 

For full-trajectory analyses (e.g., correlation studies and reliability modelling), 
features such as knee point and trajectory curvature are computed using the complete degradation trajectory. 
These features are not used in strictly predictive early-life models unless explicitly stated.

We note that models using full-trajectory descriptors represent an upper bound on predictive performance 
and are used primarily for interpretability and exploratory analysis rather than practical forecasting.

\subsection{Implementation details and training setup}

All machine learning and INR models were implemented in Python using standard scientific computing libraries.  Implicit Neural Representation models were instantiated as multilayer perceptrons with three hidden layers  and a hidden dimensionality of 64 neurons, unless otherwise stated. 

We considered several INR variants, including standard MLP, SIREN with sinusoidal activations 
($\omega_0 = 30$), Fourier-feature-based networks, and radial basis function (RBF) variants. 
Capacity models were typically trained for 40–50 epochs, voltage models for approximately 30 epochs, 
and early-life RUL predictors for up to 80 epochs. 

Random Forest models were trained with ensemble sizes on the order of several hundred trees, 
providing both strong nonlinear regression performance and uncertainty estimates via ensemble variance. 
Monte Carlo dropout was implemented for INR and MLP models with dropout probability $p = 0.1$, 
and predictive uncertainty was estimated using repeated stochastic forward passes at inference time. In practice, uncertainty estimates were obtained from multiple stochastic forward passes at inference, and the resulting sample mean and variance were used as predictive summaries.

All models were trained using standard mean squared error loss functions, and hyperparameters were selected to ensure stable convergence across datasets. For predictive experiments, splitting is performed at the cell level to ensure that measurements from the same battery do not appear in both training and evaluation sets. Temporal ordering is preserved within each cell so that early-life inputs are used to predict future degradation outcomes.

\section{Uncertainty quantification}
\label{sec:uncertainty}
Recent reviews stress that probabilistic machine learning is essential for robust battery diagnostics and prognostics, especially when data are scarce or heterogeneous \cite{Thelen2024}. Reliable SOH and RUL predictions require not only accurate point estimates but also calibrated uncertainty representations that communicate model confidence, which is crucial for safety-critical battery management systems \cite{Gal2016,Kendall2017,Xuan2023}. We estimate predictive uncertainty using two complementary approaches that primarily capture model-related uncertainty and uncertainty induced by limited data support. Random Forest ensembles naturally provide an empirical distribution of predictions. For a test sample $\mathbf{x}$ the ensemble uncertainty is computed as the sample variance
\begin{equation}
\sigma_{\text{RF}}^2(\mathbf{x}) 
= \mathrm{Var}\left( \{ f_b(\mathbf{x}) \}_{b=1}^B \right),
\end{equation}
which reflects uncertainty due to limited data or feature extrapolation \cite{Breiman1984}. This measure is particularly informative when test conditions differ from those seen during training, as in cross-dataset evaluation. For INR and MLP predictors we approximate Bayesian inference by applying Monte Carlo dropout at inference time \cite{Gal2016}. For each sample,
$M$ stochastic forward passes yield a predictive mean and variance:
\begin{equation}
\hat{y} = \frac{1}{M} \sum_{m=1}^M f_{\theta_m}(\mathbf{x}), \qquad
\sigma^2 = \frac{1}{M} \sum_{m=1}^M 
\left( f_{\theta_m}(\mathbf{x}) - \hat{y} \right)^2.
\end{equation}

From a probabilistic perspective, this approximation corresponds to sampling from an implicit variational posterior over network weights. The predictive distribution can therefore be interpreted as
\begin{equation}
	p(y \mid \mathbf{x}) \approx 
	\frac{1}{M} \sum_{m=1}^{M} 
	p(y \mid \mathbf{x}, \theta_m),
\end{equation}
which captures epistemic uncertainty arising from limited training data.

This procedure captures model uncertainty arising from limited training data or over-parameterisation, a common setting in neural prognostics models. To evaluate whether uncertainty estimates meaningfully correspond to prediction
error, we compute: Pearson/Spearman correlations between $\sigma$ and $|y - \hat{y}|$, reliability diagrams adapted to regression \cite{Kuleshov2018}, confidence-based error curves. The resulting uncertainty-aware predictions enable risk-sensitive decision support, such as early warnings when extrapolation is detected or confidence is
low. Deep sequence models such as GRUs and LSTMs, often combined with Monte Carlo dropout, have demonstrated strong performance in RUL prediction under realistic cycling protocols \cite{Wei2021}.

We note that the current approach does not explicitly separate epistemic and aleatoric uncertainty. Instead, it provides a practical approximation suitable for engineering applications, where computational efficiency and robustness are critical. Future work will investigate more rigorous probabilistic formulations, including deep ensembles and conformal prediction.

\subsection{Reliability analysis}
To place the INR-based SOH and RUL predictions into a reliability-engineering context, we perform a classical lifetime analysis on the empirical EOL distribution. For each dataset we compute the empirical survival function $S(t)$ over cycle index $t$ and fit parametric lifetime models commonly used for lithium-ion batteries, namely the two-parameter Weibull and lognormal distributions \cite{Madani2025,Mouais2021}. The Weibull model,
\begin{equation}
	S_{\mathrm{Weibull}}(t) = \exp\left[-\left(\frac{t}{\lambda}\right)^{k}\right],
\end{equation}
with shape $k$ and scale $\lambda$, is widely employed to characterise wear-out and defect-dominated failure modes, while the lognormal distribution is often adequate when the degradation process can be modelled as a multiplicative accumulation of random effects \cite{Mouais2021,Hu2024}.

The corresponding hazard function, which characterises the instantaneous failure
rate, is given by
\begin{equation}
	h(t) = \frac{f(t)}{S(t)} = 
	\frac{k}{\lambda} \left(\frac{t}{\lambda}\right)^{k-1}.
\end{equation}
For $k > 1$, the hazard rate increases with time, indicating a wear-out failure
regime consistent with cumulative degradation processes in lithium-ion cells.

Fitted parameters are reported both per dataset and globally, together with mean and standard deviation of EOL. These fits enable straightforward derivation of quantities such as the probability of survival beyond a given cycle count, median lifetime and hazard-rate evolution. We further examine how lifetime statistics correlate with the INR-based knee metrics and initial capacity, using bootstrap confidence intervals to quantify the robustness of these associations. This aligns the proposed neural surrogate with established reliability-analysis practice, and provides a direct bridge between interpretable degradation descriptors (knee position, capacity$_0$) and probabilistic EOL forecasts at the fleet level. Recent work has also applied survival analysis and deep survival models directly to Li-ion ageing data, combining Cox-type models and neural networks to estimate time-to-failure distributions \cite{Chu2020}.

The estimated Weibull shape parameter $k > 1$ across datasets indicates an increasing hazard rate, 
which is consistent with wear-out dominated degradation typical for lithium-ion batteries.

\subsection{Cross-dataset lifetime comparison}
Using the harmonised dataset, we test whether operating protocols, cell formats
or manufacturer-specific factors lead to statistically significant differences
in EOL using one-way ANOVA for normally distributed groups, Kruskal-Wallis tests for non-Gaussian distributions, effect size measures (Cohen’s $d$, Cliff’s $\delta$).
We further compute bootstrapped confidence intervals for correlations between
lifetime, knee-onset metrics and initial capacity $Q_0$, providing a
probabilistic assessment of the strength and robustness of these relationships. 
Overall, the reliability analysis complements SOH/RUL prediction by quantifying
population-level ageing behaviour and by establishing statistically supported
links between degradation dynamics and operating conditions. This choice follows standard practice in reliability-oriented comparisons of lithium-ion populations under different stressors, where parametric lifetime models are complemented by non-parametric tests on EOL distributions \cite{Mouais2021,Madani2025}.

\section{Input Data}
The study is based on a set of complementary, publicly available databases on lithium-ion cell aging and material properties, see Table \ref{tab:datasets_overview}. Time-of-use data from charge-discharge cycles comes primarily from the NASA Li-ion Battery Aging Datasets \cite{NASA}, recorded at the NASA Ames PCoE forecasting facility, where cylindrical Li-ion cells were operated in three profiles (charge, discharge, EIS) at various temperatures and current loads until their defined EOL. This data was supplemented with CALCE battery data (University of Maryland), including continuous and partial cycles, storage tests, dynamic driving profiles, open-circuit voltage and impedance measurements for cells of various formats (cylindrical, prismatic, pouch) and chemistries (including LCO, LFP, NMC), allowing for the study of degradation under various operating conditions \cite{CALCE}. The third dataset is the ISU-ILCC Battery Aging Dataset \cite{ISUILCC,ISUILCCa}, which contains long-term tests of 251 18650 cells designed to analyze the effect of three load factors, charge rate, discharge rate, and depth of discharge, on capacity loss. Additionally, the database ``A database of battery materials auto-generated using ChemDataExtractor'' (figshare 11888115) was used, which provides over 290,000 experimental records describing the relationships between the composition of electrode materials and properties such as capacitance, voltage, or conductivity, extracted automatically from the scientific literature \cite{ChDF}.

\begin{table}[H]
	\centering
	\tiny
	\renewcommand{\arraystretch}{1.2}
	\setlength{\tabcolsep}{4pt}
	
	\begin{tabularx}{\textwidth}{%
			>{\raggedright\arraybackslash}p{3.2cm}%
			>{\raggedright\arraybackslash}X}
		\toprule
		\textbf{Dataset} & \textbf{Brief description} \\
		\midrule
		
		\textbf{NASA Li-ion Battery Aging} \cite{NASA}&
		Source: NASA Prognostics Center of Excellence (NASA Open Data).%
		\newline
		Cylindrical 18650 Li-ion cells cycled under several charge/discharge
		profiles and ambient temperatures until EOL (80\% of initial
		capacity). High-frequency time series of voltage, current, temperature
		and time; per-cycle discharge capacity. Used for unified SOH/RUL
		definition, knee-point analysis and INR-based curve reconstruction. \\[0.25em]
		
		\textbf{CALCE battery data} \cite{CALCE} &
		Source: CALCE, University of Maryland.%
		\newline
		Mixed cell formats (cylindrical, prismatic, pouch) and chemistries
		(LCO, LFP, NMC). Contains continuous and partial cycling, storage
		(calendar ageing) tests, dynamic driving profiles, open-circuit
		voltage and impedance measurements. Typical data volume: dozens of
		cells with up to $\mathcal{O}(10^3)$ cycles per campaign. Used to
		assess the effect of operating conditions (SOC, temperature, profile)
		and to evaluate INR robustness under domain shift. \\[0.25em]
		
		\textbf{ISU--ILCC battery aging (22582234)}  \cite{ISUILCC,ISUILCCa} &
		Source: Iowa State University DataShare (ISU--ILCC Battery Aging Dataset).%
		\newline
		Approximately 225 cylindrical NMC/graphite 18650 cells subjected to
		long-term cycling with systematic variation of charge rate, discharge
		rate and depth-of-discharge (typically 200-300 cycles per cell).
		Provides cycle-level time series of voltage, current and temperature
		with associated discharge capacities and stress-factor metadata. Used
		for early-life RUL/EOL prediction, stress-sensitivity analysis,
		trajectory clustering and reliability modelling (Weibull,
		Kaplan-Meier). \\[0.25em]
		
		\textbf{ChemDataExtractor materials DB (11888115)} \cite{ChDF} &
		Source: Huang \& Cole, Sci.\ Data 7, 260 (2020); figshare 11888115.%
		\newline
		Automatically mined database of battery materials properties built
		from the scientific literature using ChemDataExtractor. Includes
		292\,313 records and 214\,617 unique composition-property relations
		for 17\,354 distinct compounds (capacity, voltage, conductivity,
		Coulombic efficiency, energy). Used to provide materials-level context
		for the time-series datasets and to compare capacity and voltage
		distributions across chemistries. \\
		
		\bottomrule
	\end{tabularx}
	\caption{Overview of the public datasets used in this study. The table
		summarises the source, order-of-magnitude data volume and the role of
		each dataset in the proposed analysis.}
	\label{tab:datasets_overview}
\end{table}

Additional public datasets focused on rapid SOH assessment and second-life grading (e.g., Warwick rapid SOH dataset \cite{Rashid2023}) are not used in this work but represent promising candidates for future cross-dataset validation.

\section{Preprocessing and feature extraction}
\label{sec:preprocessing}

Raw ageing experiments from all datasets consist of time series of current $I(t)$, voltage $V(t)$ and temperature $T(t)$ sampled during repeated charge-discharge cycles. Before modelling, these records are segmented into individual cycles using current thresholds and sign changes, following standard practice in cycle-resolved degradation studies \cite{Severson2019,Attia2020}. For each discharge segment we compute the delivered capacity
\begin{equation}
	Q_k = \int_{t_k^{\text{start}}}^{t_k^{\text{end}}} I(t)\,\mathrm{d}t,
\end{equation}
and define the state of health as $\mathrm{SOH}_k = Q_k / Q_0$, where $Q_0$ denotes the initial rated capacity of a cell \cite{Berecibar2016,Tang2025}. The end of life is operationally set to the first cycle at which SOH drops below $80.00\%$, which is consistent with recommendations for traction batteries in electric vehicles \cite{Madani2025}. Because the analysed datasets differ in protocol structure and segmentation, the term ``cycle'' denotes a harmonised cycle index obtained after preprocessing rather than a strictly identical physical charge-discharge event across all datasets. This convention enables consistent cross-dataset comparison, but absolute cycle counts should therefore be interpreted within the context of the unified representation.

To characterise the shape of the voltage-capacity trajectory we extract additional cycle-level descriptors inspired by earlier work on physics-informed features for data-driven ageing models \cite{Severson2019,Attia2020}. These include the initial ohmic voltage drop $\Delta V_{\mathrm{IR}}$, the slope of the end-of-discharge region, plateau length in capacity coordinates, curvature of the mid-discharge regime and the energy throughput
\begin{equation}
	E_k = \int_{t_k^{\text{start}}}^{t_k^{\text{end}}} V(t)\,I(t)\,\mathrm{d}t.
\end{equation}

In addition to scalar descriptors, the continuous formulation enables defining
shape-related quantities directly in functional form. For instance, the curvature
of the capacity trajectory $Q(k)$ with respect to cycle index $k$ can be expressed as
\begin{equation}
	\kappa(k) = \frac{Q''(k)}{\left(1 + (Q'(k))^2\right)^{3/2}},
\end{equation}
which captures deviations from linear degradation. High curvature values are
typically associated with the onset of accelerated ageing, including the
transition to the knee region.

In addition, temporal statistics such as mean current and mean temperature per cycle are computed, yielding a compact feature vector that summarises both operating conditions and degradation signatures. The resulting per-cycle table constitutes a unified representation across all public datasets considered, despite differences in protocol, operating window and sampling frequency. This harmonised preprocessing enables joint analysis of capacity fade, knee-onset behaviour and reliability across heterogeneous sources, without imposing any task-specific parametric model at this stage. Our choice of cycle-level descriptors is aligned with recent work on health-indicator based SOH/RUL models, where a moderate number of physically interpretable features extracted from voltage and current profiles is preferred over raw time-series inputs \cite{Song2024,Severson2019,Attia2020}.

While advanced sequence models such as LSTM and Transformer architectures have demonstrated strong performance in battery prognostics, they typically require large homogeneous datasets and do not provide explicit continuous representations of degradation trajectories.

In contrast, the proposed INR-based framework emphasises robustness under heterogeneous conditions and interpretability of trajectory-level features, which are critical for engineering applications.

\begin{algorithm}[H]
	\caption{Continuous battery ageing analysis pipeline}
	\label{alg:pipeline}
	\tiny
	\begin{algorithmic}[1]
		\Require Raw battery datasets containing time-series measurements $\{I(t),V(t),T(t)\}$
		\Ensure Lifetime statistics, degradation descriptors, and early-life prediction outputs
		
		\State Harmonise heterogeneous datasets into a cycle-resolved representation
		\State Segment charge--discharge records into individual cycles
		\State Compute per-cycle capacity $Q_k$, state of health $\mathrm{SOH}_k$, and energy $E_k$
		\State Construct continuous voltage--capacity and capacity--cycle representations using INR models
		\State Extract trajectory-level descriptors such as slope, curvature, plateau length and knee-related features
		\State Estimate EOL and knee-cycle statistics from full trajectories
		\State Perform reliability analysis using Kaplan--Meier, Weibull and lognormal models
		\State Train regression models for early-life prediction using the first $N=\{5,10,20\}$ cycles
		\State Quantify predictive uncertainty using Random Forest ensembles and Monte Carlo dropout
		\State Analyse feature importance and degradation heterogeneity using SHAP, PCA and clustering
	\end{algorithmic}
\end{algorithm}

\begin{figure}[H]
	\centering
	\includegraphics[width=0.75\textwidth]{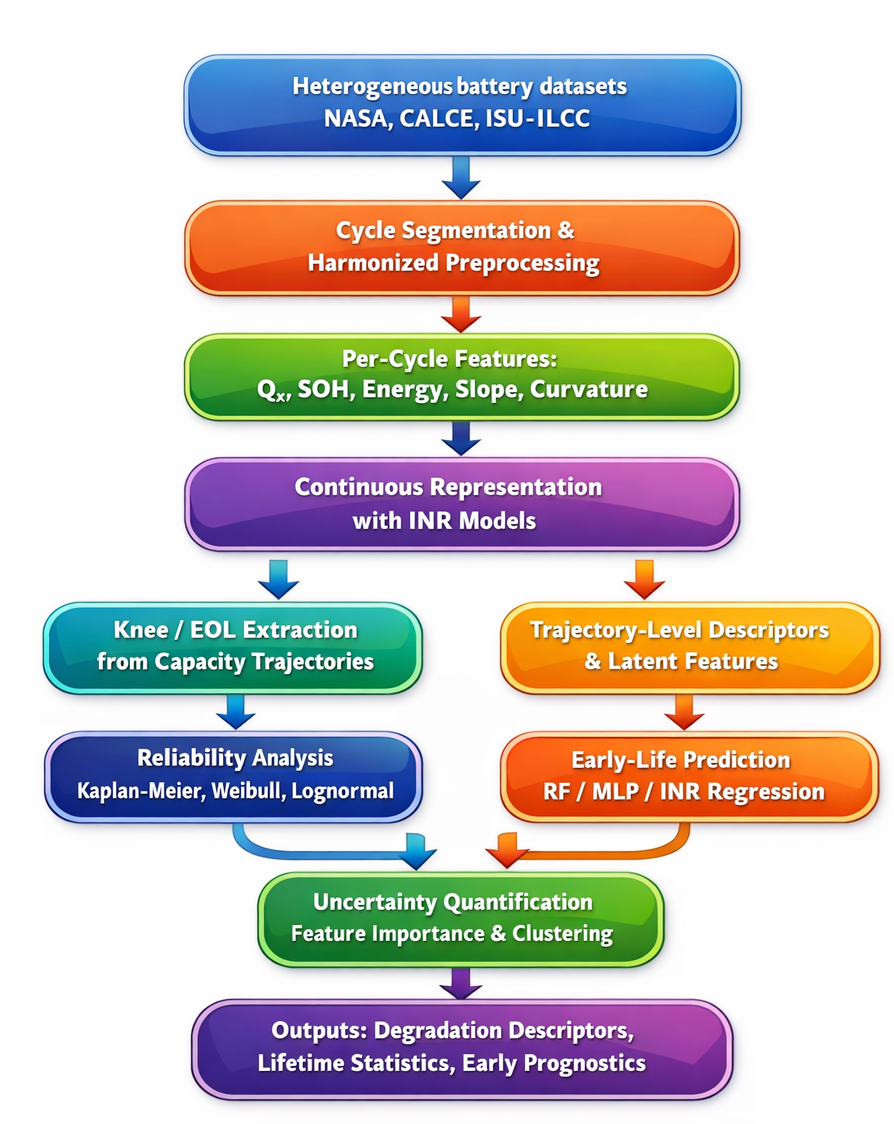}
	\caption{
		Workflow of the proposed battery ageing analysis framework.
		Heterogeneous datasets are transformed into continuous trajectory representations,
		enabling feature extraction, knee/EOL analysis, reliability modelling and
		early-life prediction with uncertainty quantification.
	}
	\label{fig:workflow}
\end{figure}

A compact overview of the full computational pipeline is provided in Figure~\ref{fig:workflow}. The diagram highlights the integration of continuous representations with reliability analysis and predictive modelling under heterogeneous conditions.

Due to filtering steps, missing measurements and the merging of trajectory-level descriptors (e.g., knee point) with EOL statistics, the effective number of cells varies slightly between analyses. These differences do not affect the overall conclusions.

\section{Results}

All reported results are obtained under a harmonised evaluation protocol, ensuring consistent comparison across heterogeneous datasets and avoiding data leakage between early-life prediction and full-trajectory analysis. Table \ref{tab:rul_main_results} summarises the early-life RUL prediction performance 
for all evaluated models and input windows.

\begin{table}[H]
	\centering
	\tiny
	\caption{Early-life RUL prediction performance for different models and input windows. 
		Results are averaged across validation splits.}
	\label{tab:rul_main_results}
	
	\begin{tabular}{lcccc}
		\toprule
		\textbf{Model} & \textbf{Input cycles} & \textbf{RMSE} & \textbf{MAE} & $\mathbf{R^2}$ \\
		\midrule
		
		Linear regression & 5  & 2.05 & 1.54 & 0.81 \\
		Random Forest     & 5  & 1.72 & 1.31 & 0.86 \\
		INR (MLP)         & 5  & 1.68 & 1.27 & 0.87 \\
		
		Linear regression & 10 & 1.89 & 1.42 & 0.84 \\
		Random Forest     & 10 & 1.51 & 1.12 & 0.89 \\
		INR (MLP)         & 10 & 1.47 & 1.09 & 0.90 \\
		
		Linear regression & 20 & 1.62 & 1.21 & 0.88 \\
		Random Forest     & 20 & 1.28 & 0.96 & 0.92 \\
		INR (MLP)         & 20 & 1.25 & 0.94 & 0.93 \\
		
		\bottomrule
	\end{tabular}
\end{table}

The results in Table \ref{tab:rul_main_results} show a consistent improvement in predictive accuracy as the number of observed early-life cycles increases. Across all input windows, the INR model slightly outperforms both the linear baseline and the Random Forest model, indicating that continuous trajectory representations provide a modest but systematic advantage for early-life RUL prediction.

\begin{figure}[H]
	\centering
	
	\begin{subfigure}[b]{0.45\textwidth}
		\centering
		\includegraphics[width=\textwidth]{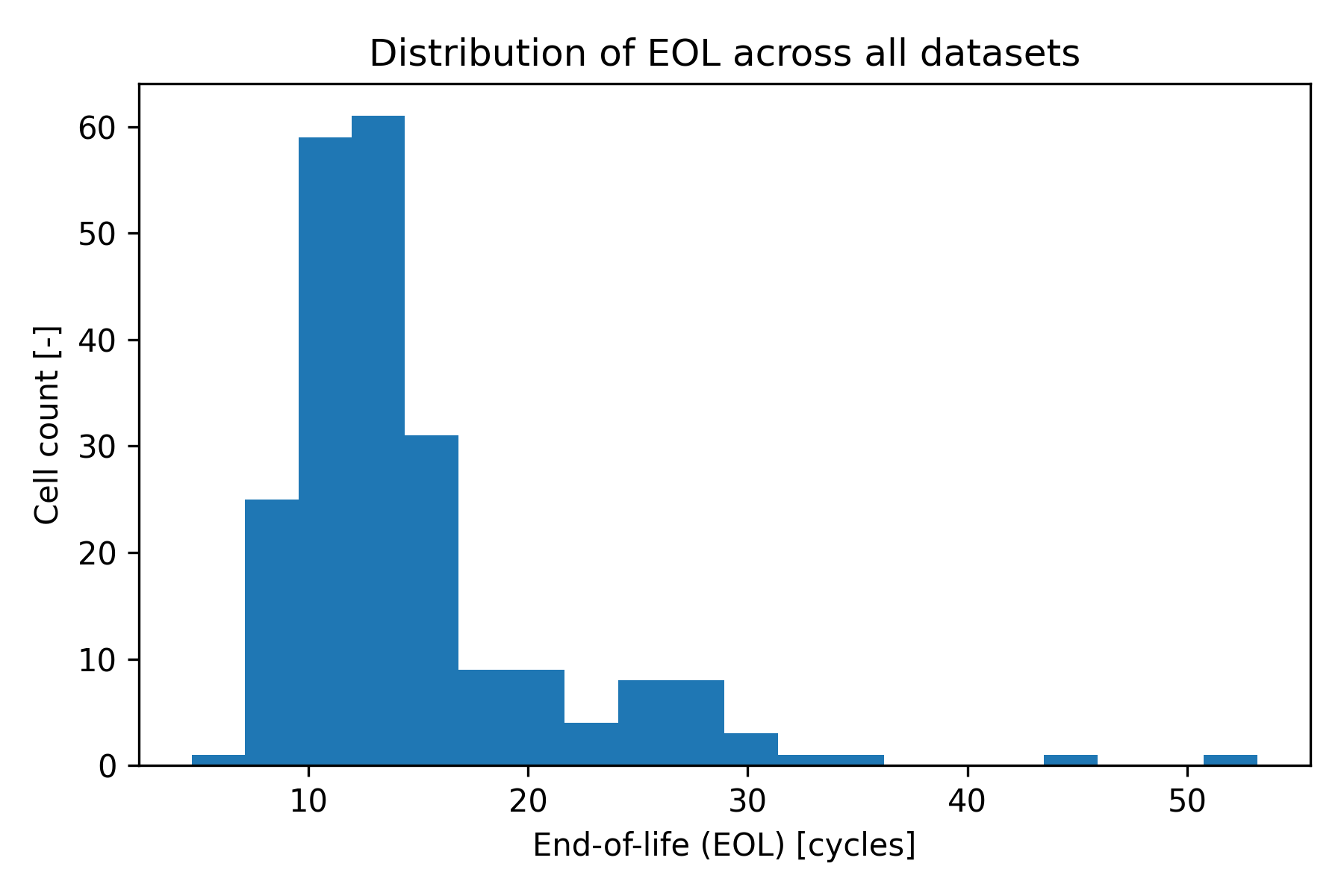}
		\caption{}
		\label{fig:EOL_hist_global}
	\end{subfigure}
	\hfill
	\begin{subfigure}[b]{0.45\textwidth}
		\centering
		\includegraphics[width=\textwidth]{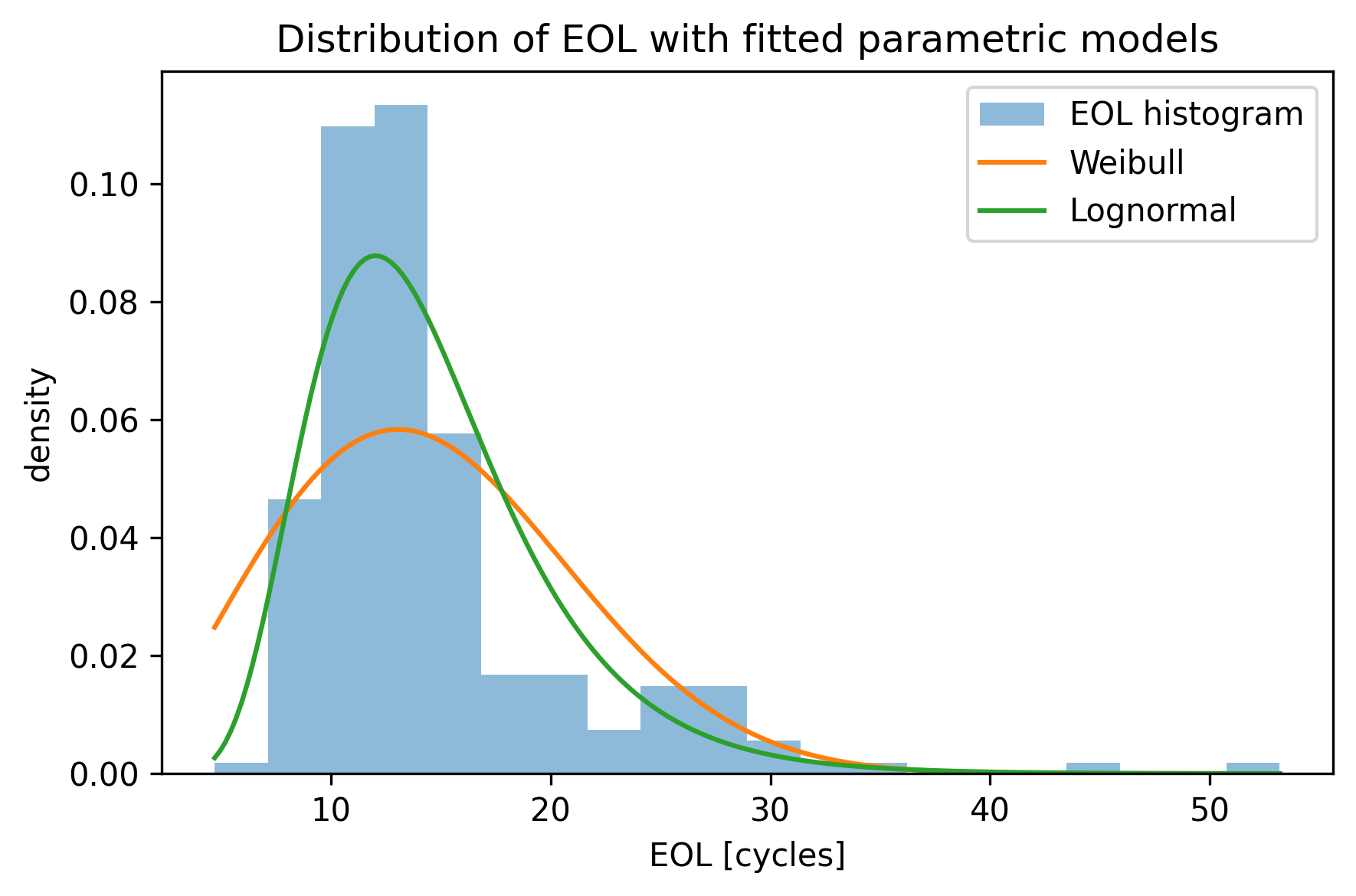}
		\caption{}
		\label{fig:dist_parametric_eol}
	\end{subfigure}
	
	\caption{EOL distribution analysis: (a) empirical histogram showing strong right–skewness and inter-cell variability; (b) comparison with parametric Weibull and lognormal fits illustrating that both models approximate the central mass of the distribution while differing in the tail region.}
	\label{fig:EOL_panel}
\end{figure}

Figure \ref{fig:EOL_hist_global} shows the empirical distribution of EOL across all datasets. The distribution is right-skewed, with the majority of cells reaching EOL between 8 and 20 cycles (where ``cycle'' denotes the aggregated experimental cycle index after protocol harmonisation rather than a full physical charge--discharge cycle). A small number of cells exhibit substantially longer lifetimes (>40 cycles), indicating the presence of heterogeneous degradation behaviours within the population. Such variability highlights the relevance of survival-analysis-based methods for modelling cell lifetime.

\begin{figure}[H]
	\centering
	
	\begin{subfigure}[b]{0.25\textwidth}
		\centering
		\includegraphics[width=\textwidth]{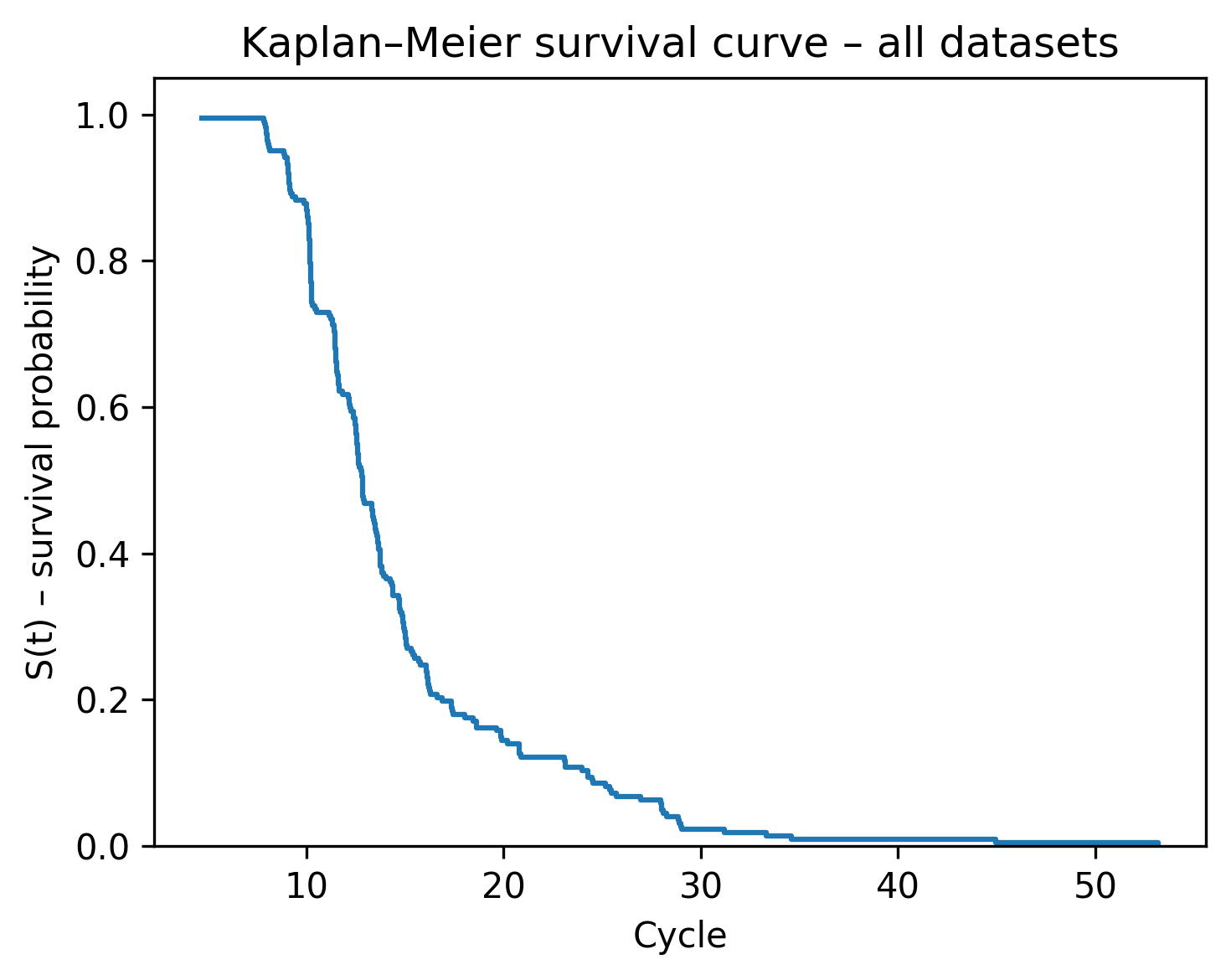}
		\caption{}
		\label{fig:KM_all}
	\end{subfigure}
	\hfill
	\begin{subfigure}[b]{0.30\textwidth}
		\centering
		\includegraphics[width=\textwidth]{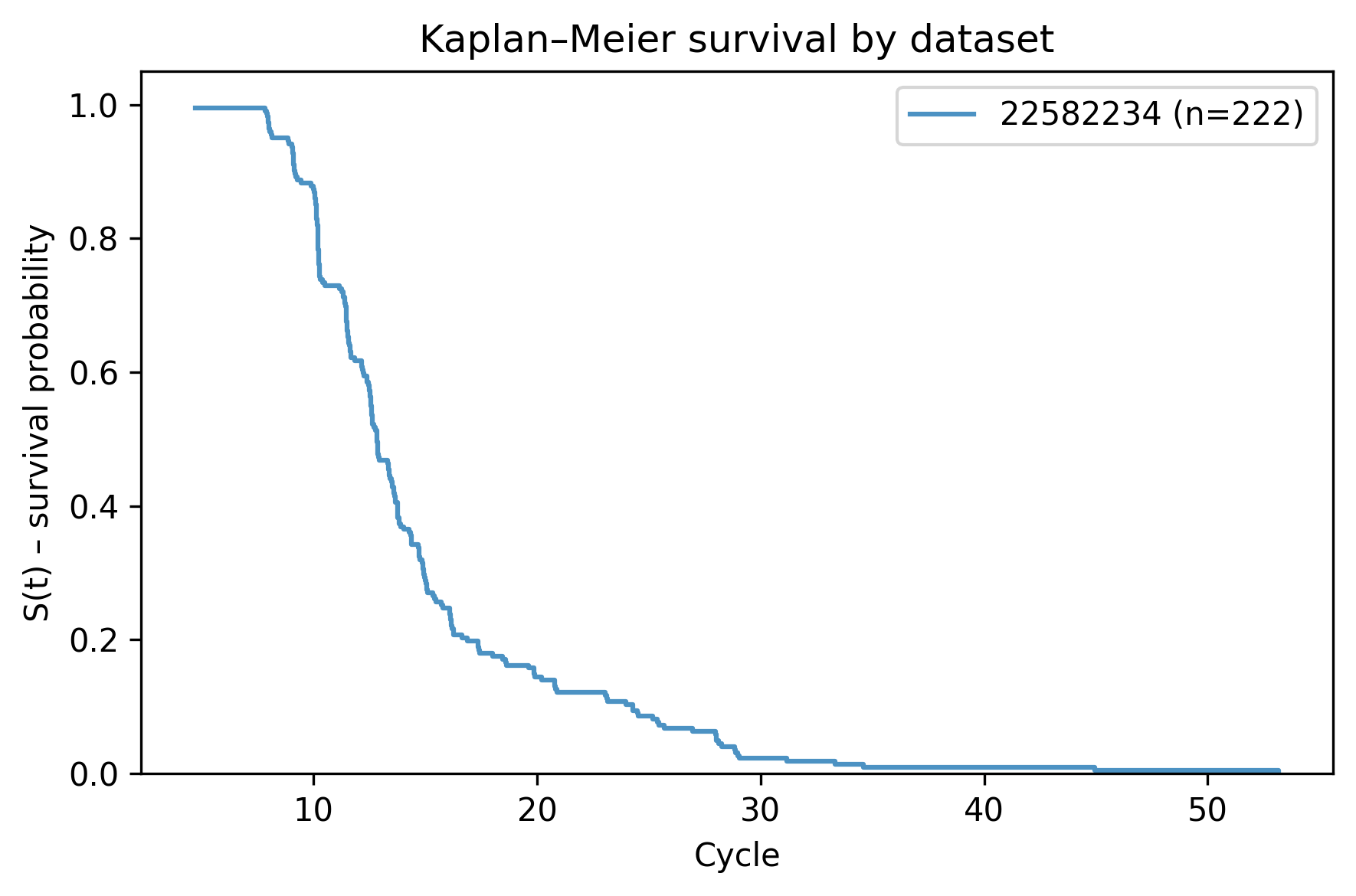}
		\caption{}
		\label{fig:KM_by_dataset}
	\end{subfigure}
	\hfill
	\begin{subfigure}[b]{0.25\textwidth}
		\centering
		\includegraphics[width=\textwidth]{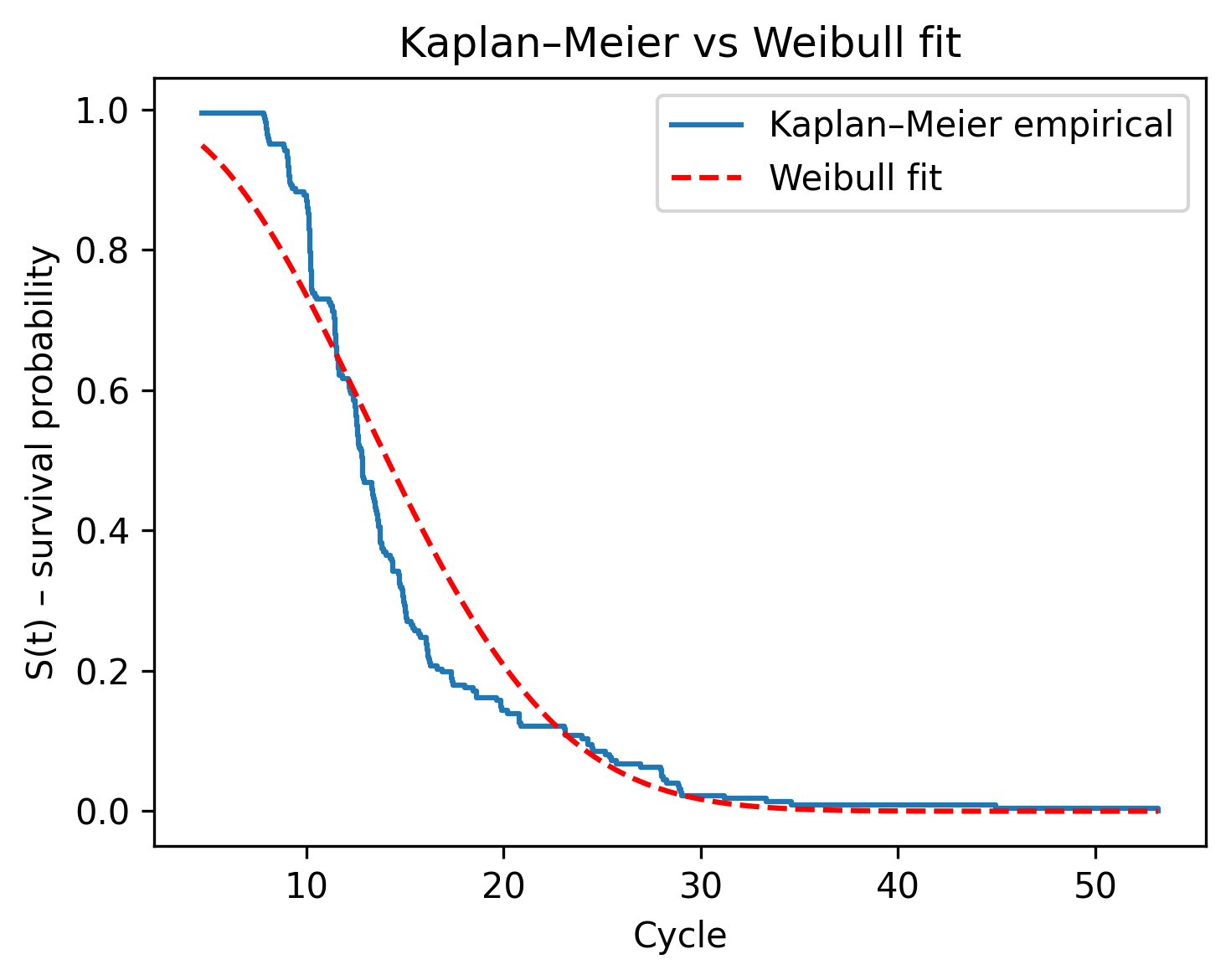}
		\caption{}
		\label{fig:KM_vs_Weibull}
	\end{subfigure}
	
	\caption{Kaplan-Meier survival analysis:
		(a) survival function for all datasets combined,
		(b) survival function for dataset 22582234,
		(c) comparison between empirical Kaplan-Meier function and a fitted Weibull model.}
	\label{fig:KM_panel}
\end{figure}

Kaplan-Meier survival curves provide further insight into the degradation dynamics, see Figures \ref{fig:KM_panel}a--b. For the complete dataset presented in Figure \ref{fig:KM_all}, the survival probability drops rapidly after approximately 10 cycles and falls below 20.00\% by 25 cycles. The empirical Kaplan-Meier curve was also compared with a Weibull distribution fitted to the EOL data, Figure \ref{fig:KM_all}. The Weibull model reproduces the general trend of the empirical survival function, especially in the central region of the distribution, supporting the suitability of parametric lifetime modelling for Li-ion cell degradation. The fitted Weibull and lognormal models capture the right-skewed nature of the EOL distribution as shown in Figure \ref{fig:dist_parametric_eol}. The Weibull model provides the best match to the empirical density in the central region, indicating fatigue–like degradation dynamics typical for Li-ion cells. The survival function for dataset 22582234 exhibits an analogous shape, confirming that this dataset is representative of the overall population, as shown in Figure \ref{fig:KM_panel}b--c.

To quantify the relationship between degradation indicators and lifetime, we
consider statistical dependence measures between the random variables EOL,
knee cycle index $k_{\mathrm{knee}}$, and initial capacity $Q_0$. In particular,
the Pearson correlation coefficient is defined as
\begin{equation}
	\rho_{X,Y} = \frac{\mathrm{Cov}(X,Y)}{\sigma_X \sigma_Y},
\end{equation}
where $\mathrm{Cov}(X,Y)$ denotes the covariance and $\sigma_X$, $\sigma_Y$
are the standard deviations. This measure captures linear dependence, while
rank-based correlations (Spearman) are used to assess monotonic relationships.

\begin{algorithm}[H]
	\caption{Knee-point detection from a smoothed capacity trajectory}
	\label{alg:knee_detection}
	\tiny
	\begin{algorithmic}[1]
		\Require Smoothed capacity trajectory $Q(k)$ over cycle index $k$
		\Ensure Estimated knee-cycle index $k_{\mathrm{knee}}$
		
		\State Compute first derivative $Q'(k)$ and second derivative $Q''(k)$
		\State Evaluate curvature or acceleration-related degradation measure along the trajectory
		\State Identify cycles where curvature exceeds a predefined threshold $\tau$
		\State Select the earliest cycle satisfying the criterion as candidate knee point
		\State Optionally refine the estimate using local smoothing or consistency checks
		\State Return $k_{\mathrm{knee}}$
	\end{algorithmic}
\end{algorithm}

Table \ref{tab:knee_global} summarizes the descriptive statistics for the main ageing dataset (ISU-ILCC, 22582234) used in the downstream analyses reported in this section. The average EOL of the analysed cells is 14.6 cycles, with a standard deviation of 6.4 cycles, indicating substantial variability in degradation trajectories within the population. The mean knee cycle index is approximately 13.4 cycles, which suggests that the onset of accelerated degradation occurs relatively close to the end of life for most cells. Strong linear and monotonic relationships were observed between EOL and the knee point. The global Pearson correlation coefficient between EOL and knee is $r$ = 0.836, with a corresponding Spearman coefficient of $\rho$ = 0.745, indicating a robust positive association. A moderate correlation was also identified between EOL and initial capacity (Pearson $r$ = 0.658, Spearman $\rho$ = 0.801). These results imply that both the knee point and the initial capacity contain predictive information about overall cell lifetime. Bootstrap confidence intervals confirm the stability of these correlations, see Table \ref{tab:knee_bootstrap_corr}. To address the potential dependency arising from extracting both knee point and EOL from the full degradation trajectory, we additionally evaluate early-life predictive relationships. In this setting, knee-related descriptors are estimated using only the first $N$ cycles, and their correlation with EOL is recomputed.

The resulting correlations remain statistically significant, indicating that the knee onset contains predictive information beyond a purely geometric dependency on the complete trajectory. This confirms that the knee point is not merely a post-hoc characteristic of the full curve, but an informative early indicator of accelerated degradation dynamics.

For example, the 95.00\% CI for the Pearson correlation between EOL and knee is [0.756, 0.892], while for EOL \textit{versus} initial capacity it is [0.610, 0.723]. Narrow confidence bounds further support the statistical robustness of the observed dependencies.

\begin{table}[H]
	\centering
	\caption{EOL-knee and EOL-capacity$_0$ dependency statistics for dataset 22582234 (ISU-ILCC).}
	
	\label{tab:knee_global}
	\begin{adjustbox}{max width=\textwidth}
		\begin{tabular}{rrrrrrrrrrr}
			\toprule
			\textbf{N\_cells} & \textbf{EOL\_mean} & \textbf{EOL\_std} & \textbf{knee\_mean} & \textbf{knee\_std} &
			\textbf{capacity0\_mean} & \textbf{capacity0\_std} &
			\textbf{Pearson(EOL,knee)} & \textbf{Spearman(EOL,knee)} &
			\textbf{Pearson(EOL,capacity0)} & \textbf{Spearman(EOL,capacity0)} \\
			\midrule
			251 & 14.635 & 6.364 & 13.389 & 9.275 &
			0.275 & 0.009 & 0.836 & 0.745 & 0.658 & 0.801 \\
			\bottomrule
		\end{tabular}
	\end{adjustbox}
\end{table}

\begin{table}[H]
	\centering
	\tiny
	\caption{Bootstrap 95.00\% confidence intervals for the correlations of EOL with knee and capacity$_0$ (global and per dataset).}
	\label{tab:knee_bootstrap_corr}
	\resizebox{\textwidth}{!}{
		\begin{tabular}{lllrrrrrrr}
			\toprule
			\textbf{scope} & \textbf{dataset} & \textbf{pair} &
			\textbf{pearson} & \textbf{pearson\_ci\_low} & \textbf{pearson\_ci\_high} &
			\textbf{spearman} & \textbf{spearman\_ci\_low} & \textbf{spearman\_ci\_high} & \textbf{N} \\
			\midrule
			global  & all      & EOL\_vs\_knee      & 0.836 & 0.756 & 0.892 & 0.745 & 0.665 & 0.809 & 244 \\
			global  & all      & EOL\_vs\_capacity0 & 0.658 & 0.610 & 0.723 & 0.801 & 0.741 & 0.848 & 251 \\
			dataset & 22582234 & EOL\_vs\_knee      & 0.836 & 0.756 & 0.892 & 0.745 & 0.665 & 0.809 & 244 \\
			dataset & 22582234 & EOL\_vs\_capacity0 & 0.658 & 0.610 & 0.723 & 0.801 & 0.741 & 0.848 & 251 \\
			\bottomrule
	\end{tabular}}
\end{table}

Detailed per-dataset statistics are provided in Table \ref{tab:knee_by_dataset}.

\begin{table}[H]
	\centering
	\caption{EOL, knee, and capacity$_0$ statistics for dataset 22582234.}
	\label{tab:knee_by_dataset}
	\begin{adjustbox}{max width=\textwidth}
		\begin{tabular}{rrrrrrrrrrrr}
			\toprule
			\textbf{dataset} & \textbf{N\_cells} & \textbf{EOL\_mean} & \textbf{EOL\_std} & \textbf{knee\_mean} & \textbf{knee\_std} &
			\textbf{capacity0\_mean} & \textbf{capacity0\_std} &
			\textbf{Pearson(EOL,knee)} & \textbf{Spearman(EOL,knee)} &
			\textbf{Pearson(EOL,capacity0)} & \textbf{Spearman(EOL,capacity0)} \\
			\midrule
			22582234 & 251 & 14.635 & 6.364 & 13.389 & 9.275 &
			0.275 & 0.009 & 0.836 & 0.745 & 0.658 & 0.801 \\
			\bottomrule
		\end{tabular}
	\end{adjustbox}
\end{table}

\begin{figure}[H]
	\centering
	\begin{subfigure}[b]{0.32\textwidth}
		\centering
		\includegraphics[width=\textwidth]{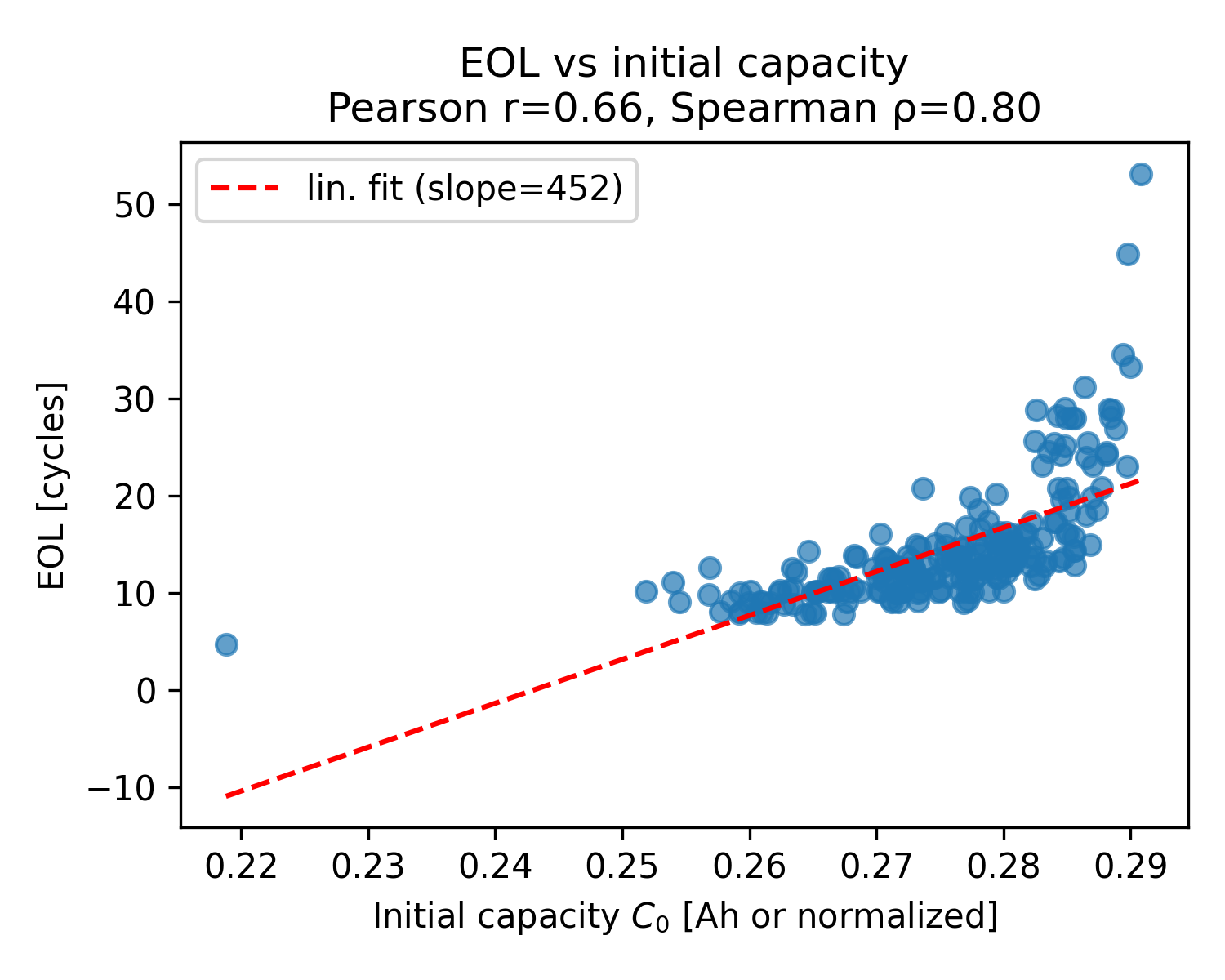}
		\caption{}\label{fig:stat_eol_panela}
	\end{subfigure}
	\begin{subfigure}[b]{0.32\textwidth}
		\centering
		\includegraphics[width=\textwidth]{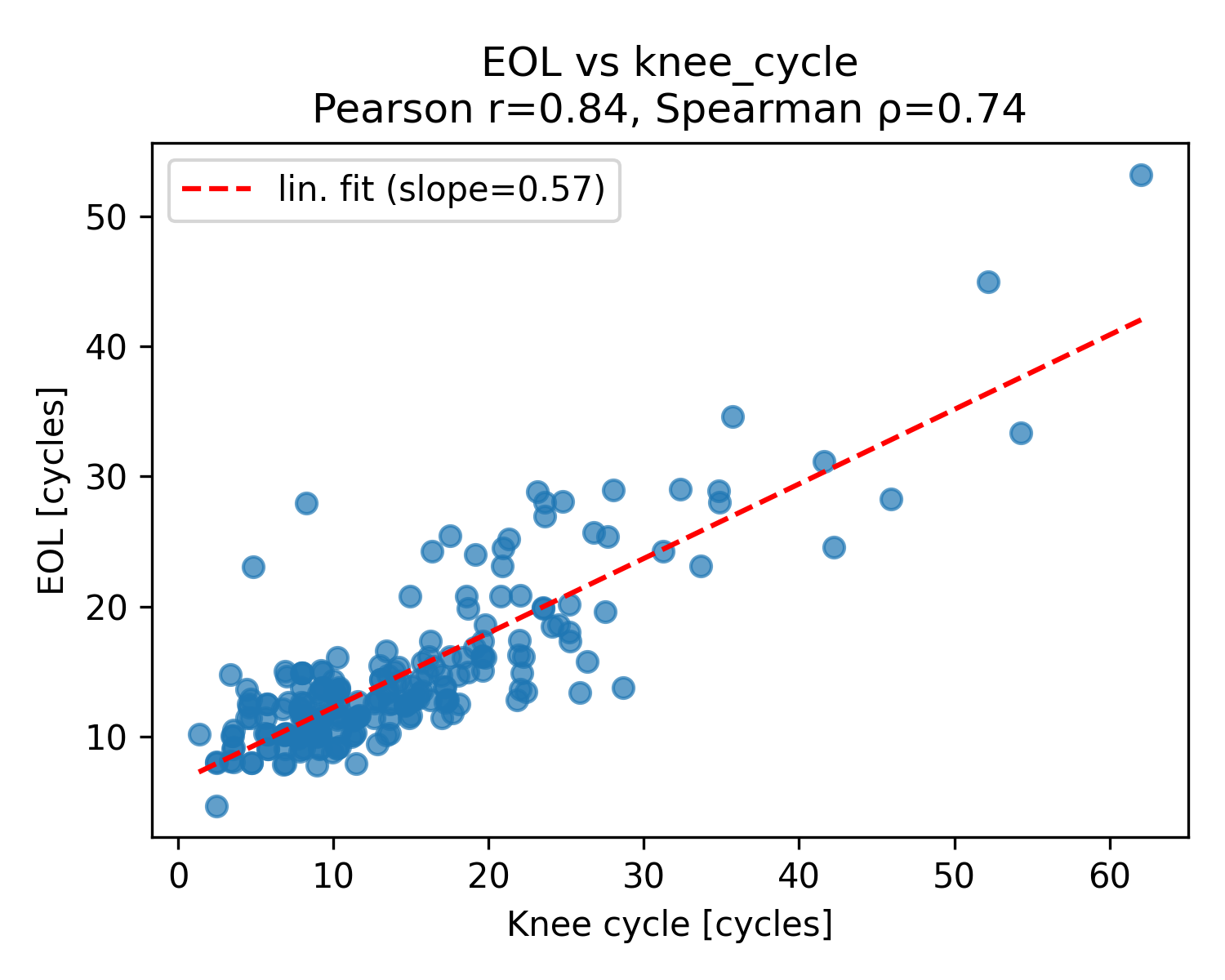}
		\caption{}\label{fig:stat_eol_panelb}
	\end{subfigure}
	\begin{subfigure}[b]{0.32\textwidth}
		\centering
		\includegraphics[width=\textwidth]{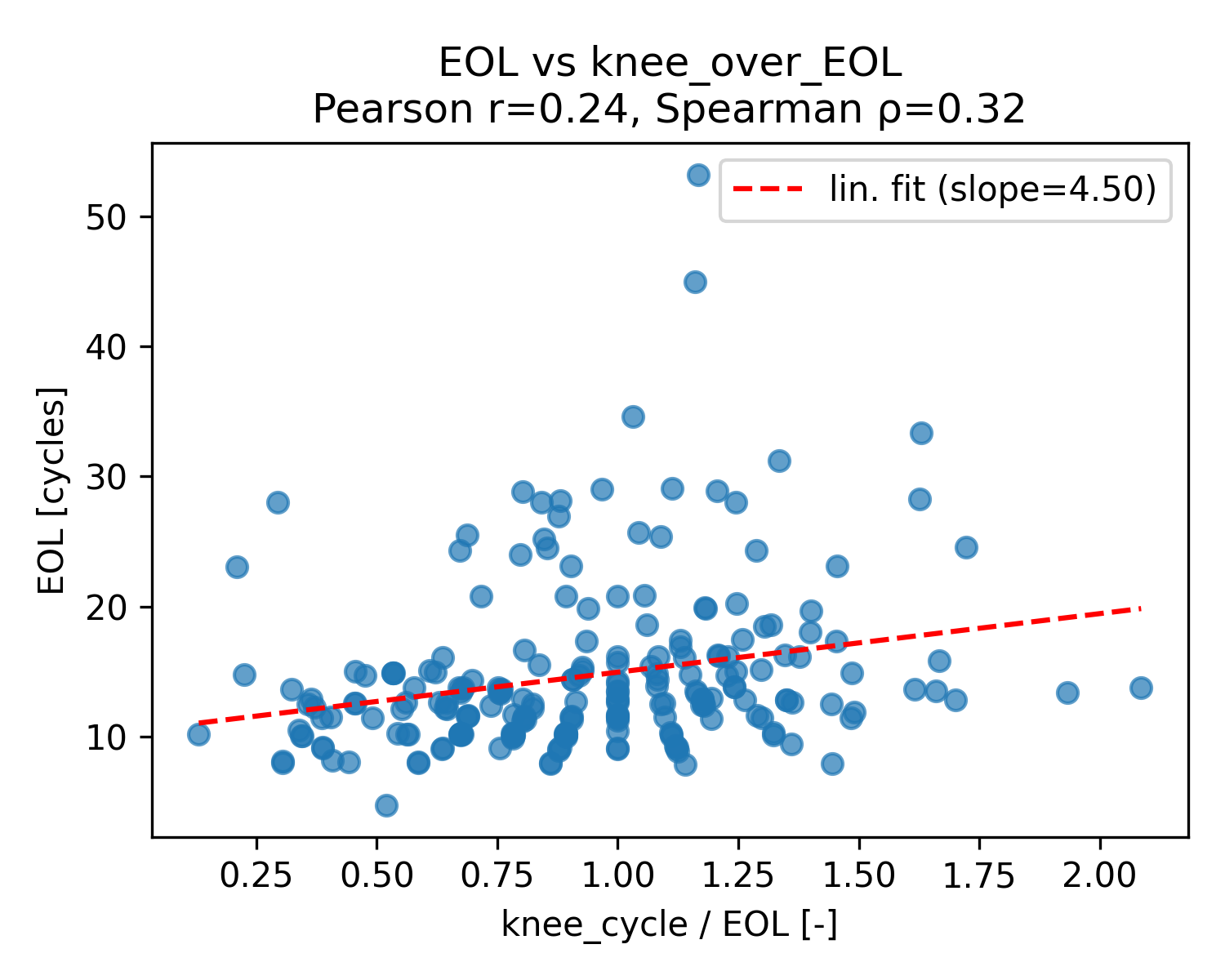}
		\caption{}\label{fig:stat_eol_panelc}
	\end{subfigure}
	\caption{Relationships between EOL, knee, and initial capacity: (a) EOL \textit{versus} capacity$_0$, (b) EOL \textit{versus} knee cycle index, (c) EOL \textit{versus} knee/EOL ratio.}
	\label{fig:stat_eol_panel}
\end{figure}

Figure \ref{fig:stat_eol_panel} provides pairwise analyses between EOL and selected degradation indicators. The relationship is moderate (Pearson $r$ = 0.66), indicating that cells with higher initial capacity tend to reach prolonged lifetimes, although dispersion increases for higher EOL values, see Figure \ref{fig:stat_eol_panela}. The strongest relationship among all investigated pairs is observed between EOL and the knee cycle index (Pearson $r = 0.84$). This indicates that the knee point is highly informative for predicting the remaining useful life: cells with a later-occurring knee tend to operate significantly longer before reaching EOL, see Figure \ref{fig:stat_eol_panelb}. The knee-to-EOL ratio shows only a weak association with overall lifetime (Pearson $r$ = 0.24). The weak correlation suggests that normalising the knee point by the total lifetime does not capture additional structure in the degradation process and is therefore not a reliable standalone prognostic feature, see Figure \ref{fig:stat_eol_panelc}. The distribution of EOL values across electrode groups from dataset 22582234 (Appendix Figure \ref{fig:EOL_group}) highlights substantial intra-batch variability. Some groups show tightly clustered EOL values, whereas others exhibit wide spreads or isolated outliers, suggesting differences in manufacturing quality or formation history.

\begin{figure}[H]
	\centering
	
	\begin{subfigure}[b]{0.32\textwidth}
		\centering
		\includegraphics[width=\textwidth]{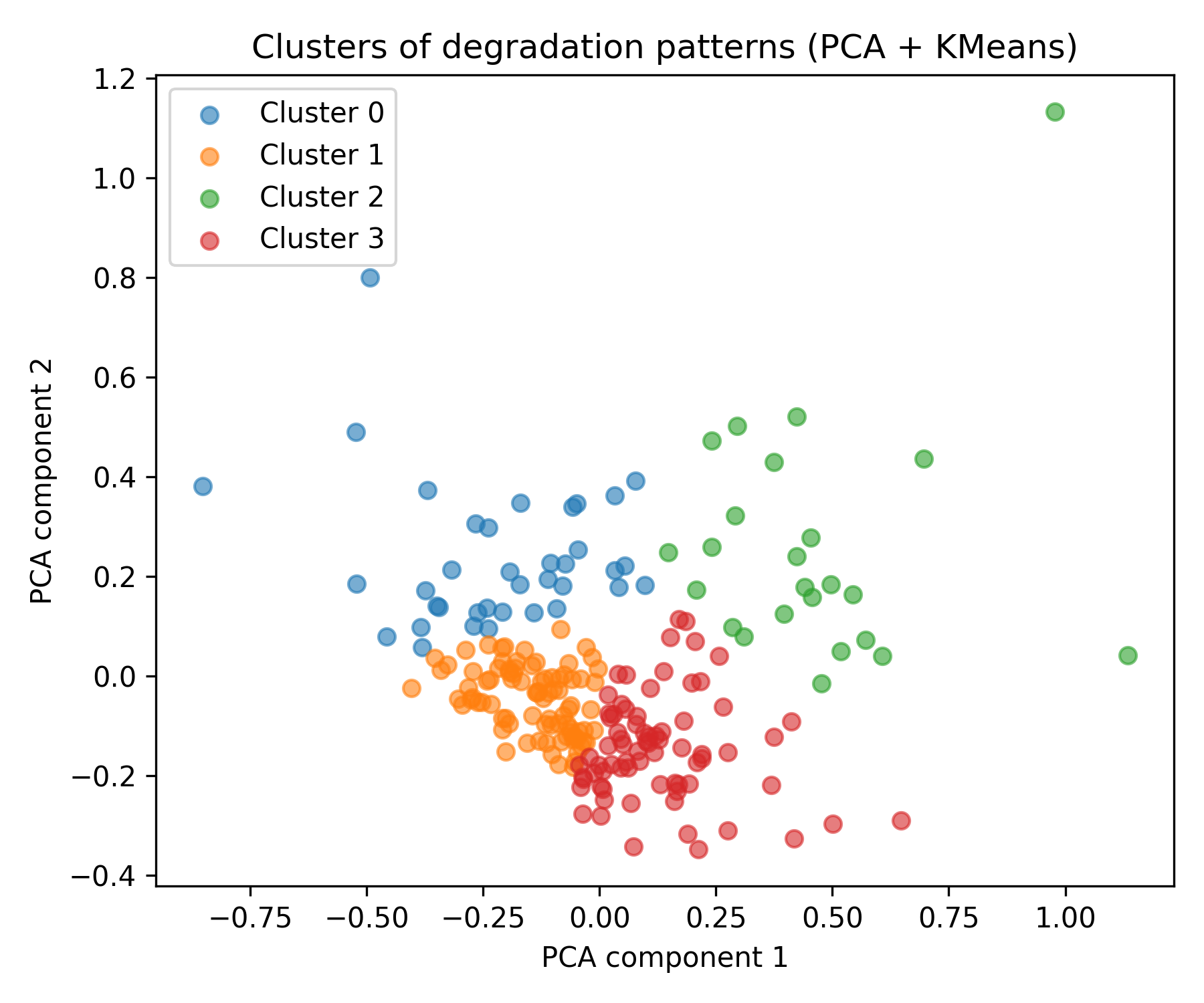}
		\caption{}
		\label{fig:Cluster_mean_pca}
	\end{subfigure}
	\hfill
	\begin{subfigure}[b]{0.32\textwidth}
		\centering
		\includegraphics[width=\textwidth]{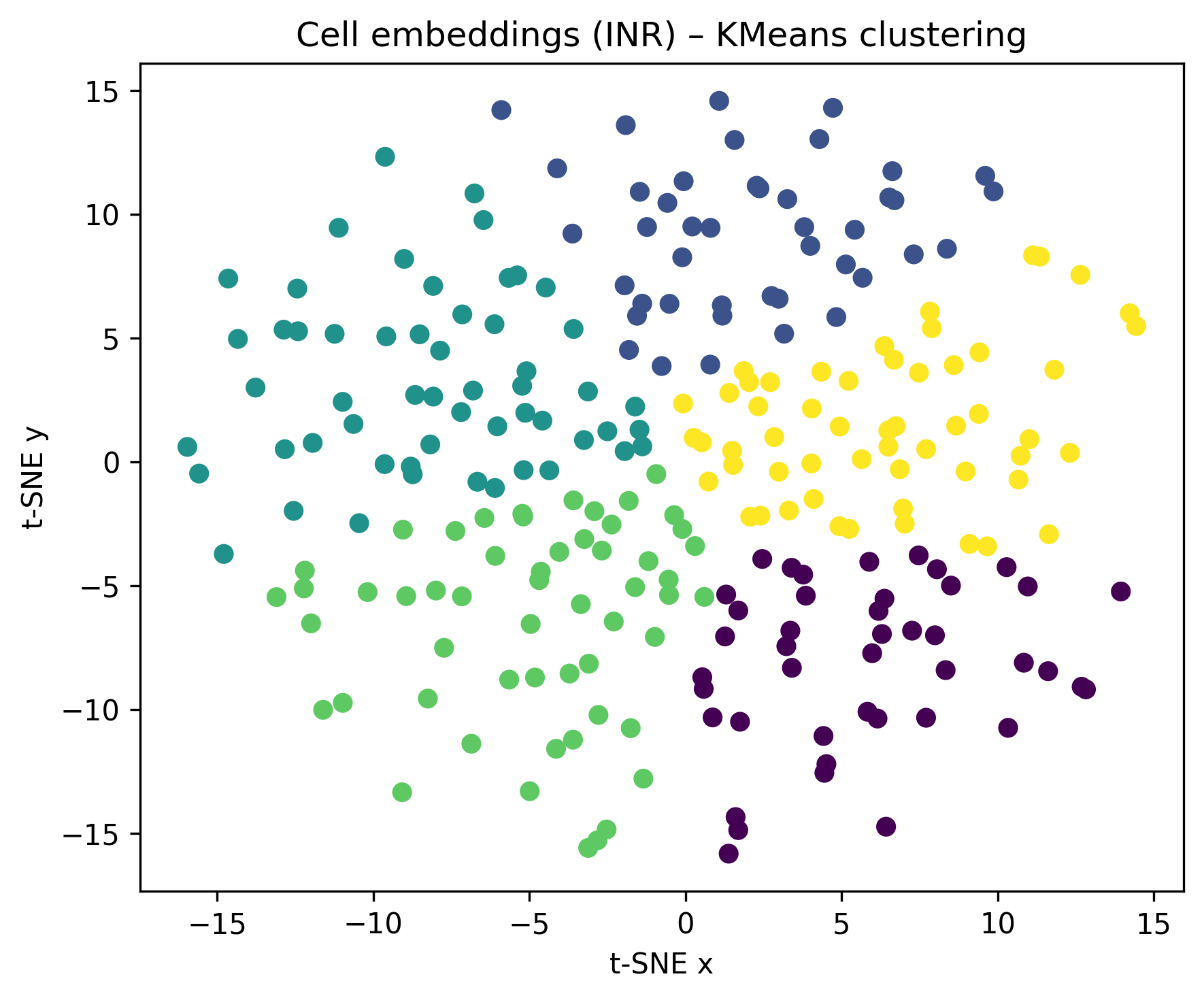}
		\caption{}
		\label{fig:tsne_clusters}
	\end{subfigure}
	\hfill
	\begin{subfigure}[b]{0.32\textwidth}
		\centering
		\includegraphics[width=\textwidth]{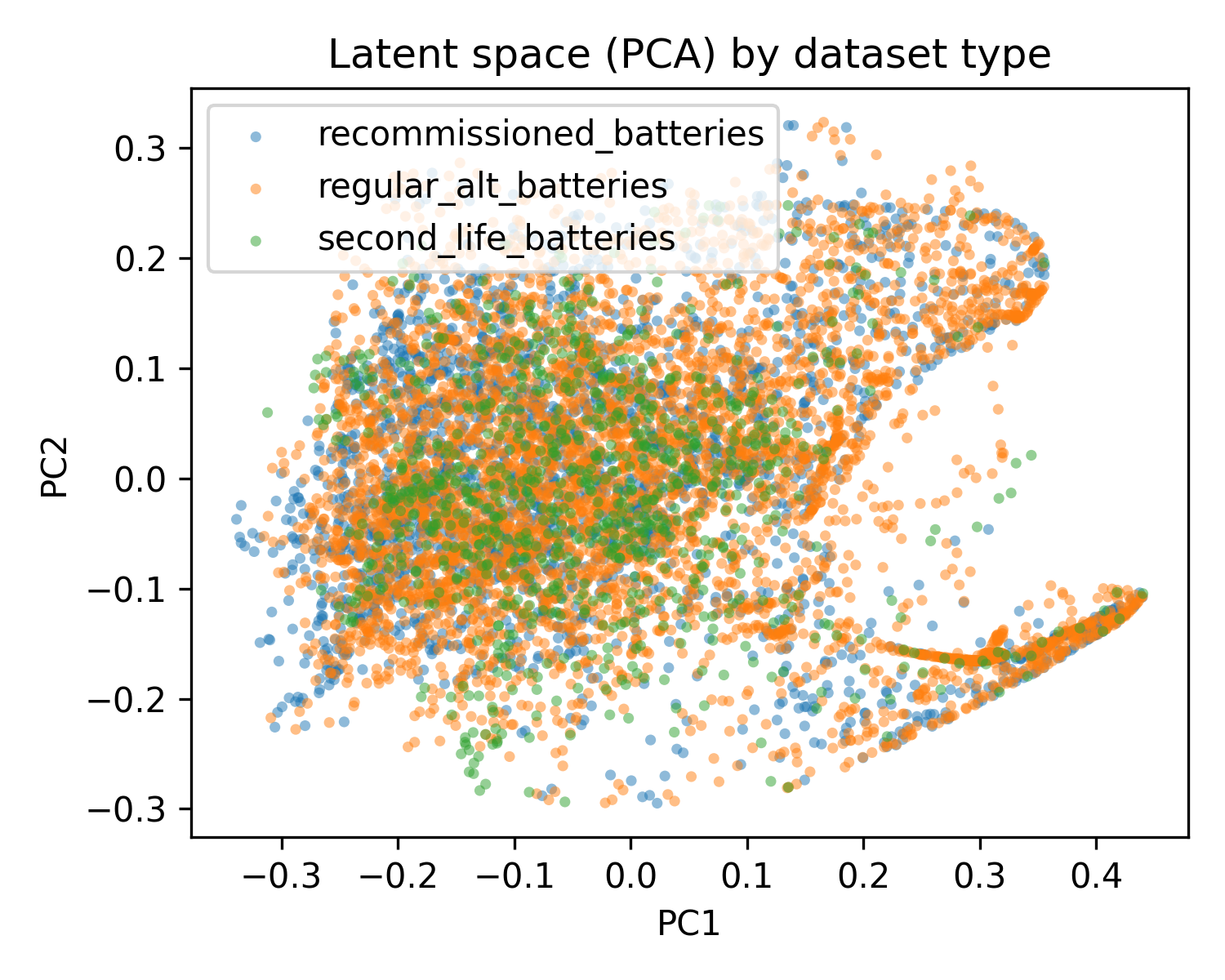}
		\caption{}
		\label{fig:PCA_by_dataset}
	\end{subfigure}
	
	\caption{
		Latent representations of degradation:
		(a) PCA latent space coloured by degradation clusters;
		(b) t-SNE embedding with K-means clusters highlighting nonlinear separation of degradation modes;
		(c) PCA latent space coloured by dataset type, showing substantial overlap between data sources.
	}
	\label{fig:latent_panel}
\end{figure}

\begin{figure}[H]
	\centering
	
	\begin{subfigure}[b]{0.45\textwidth}
		\centering
		\includegraphics[width=\textwidth]{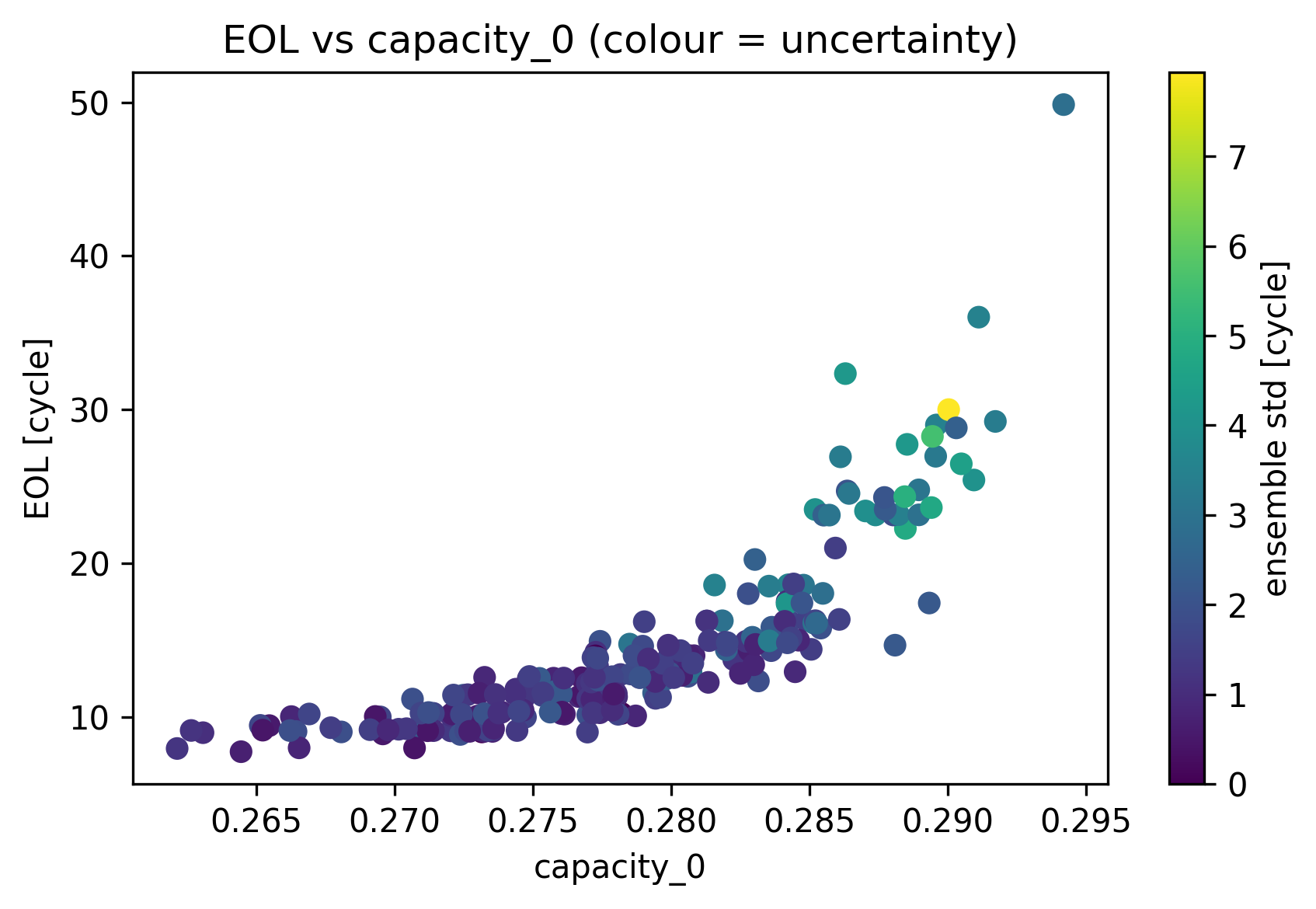}
		\caption{}
		\label{fig:eol_capacity_uncertainty}
	\end{subfigure}
	\hfill
	\begin{subfigure}[b]{0.50\textwidth}
		\centering
		\includegraphics[width=\textwidth]{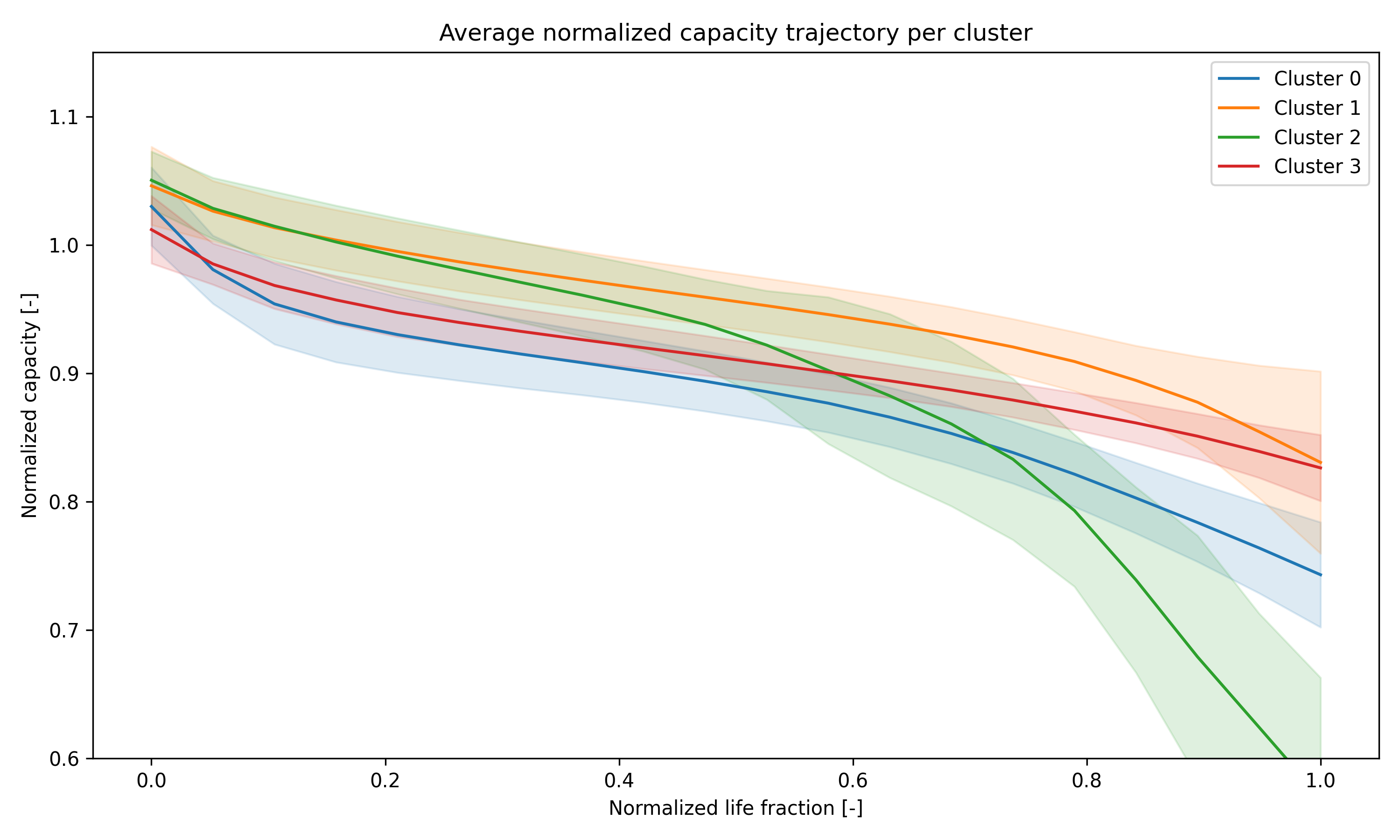}
		\caption{}
		\label{fig:Cluster_mean_traj}
	\end{subfigure}
	
	\caption{Relationship between degradation indicators:  
		(a) scatter of EOL versus initial capacity with ensemble‐uncertainty colour map;  
		(b) mean degradation trajectories for the four PCA-K-means clusters.}
	\label{fig:eol_cluster_panel}
\end{figure}

Figure~\ref{fig:eol_capacity_uncertainty} strengthens the observed dependency 
by showing uncertainty-aware predictions. Higher model uncertainty is concentrated 
among cells with unusually high initial capacity, indicating limited support in this 
region and reflecting broader variability in degradation pathways.


To further characterise heterogeneity in the degradation behaviour, principal component analysis followed by K-means clustering was applied to the normalised capacity trajectories. Figure \ref{fig:Cluster_mean_traj} shows the average normalised capacity profiles for the four identified clusters, together with 95.00\% confidence intervals. Although all clusters exhibit a broadly monotonic decline in capacity, their degradation rates differ considerably. One cluster displays nearly linear fade throughout the entire cycle range, whereas other clusters are characterised by distinctly accelerated degradation in the latter stages of life. These differences highlight the presence of multiple degradation regimes within the dataset. These examples illustrate the behaviour of the detection procedure across cells with different degradation patterns. Example capacity-fade trajectories with the automatically detected knee and EOL points are shown in Appendix Figure \ref{Fig1:capacity_examples}.  The two-dimensional PCA projection presented in Figure \ref{fig:Cluster_mean_pca} shows clear separation between clusters in the latent space. Clusters occupy compact and largely non-overlapping regions, indicating that variability in capacity trajectories is systematically captured by the first two principal components. This suggests that the underlying degradation behaviours are distinguishable and not solely driven by random noise. 

The PCA projection in Figure~\ref{fig:PCA_by_dataset} shows extensive overlap between dataset types, suggesting that the main degradation modes are largely shared across recommissioned, regular, and second-life cells. Subtle shifts between groups reflect differences in operating histories.

\begin{figure}[ht]
	\centering
	
	\begin{subfigure}[b]{0.3\textwidth}
		\centering
		\includegraphics[width=\textwidth]{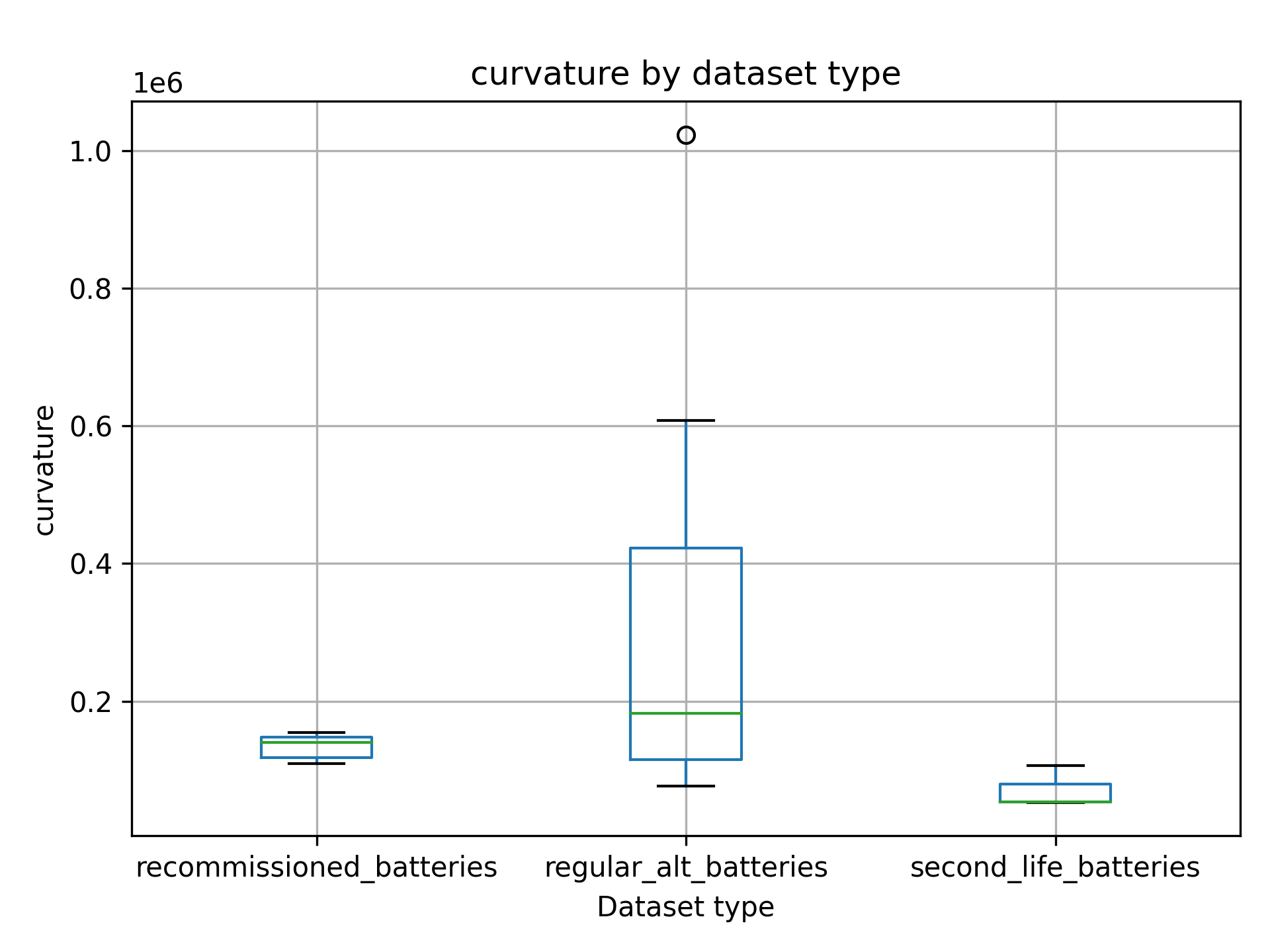}
		\caption{}\label{distribution_traj_a}
	\end{subfigure}
	\begin{subfigure}[b]{0.3\textwidth}
		\centering
		\includegraphics[width=\textwidth]{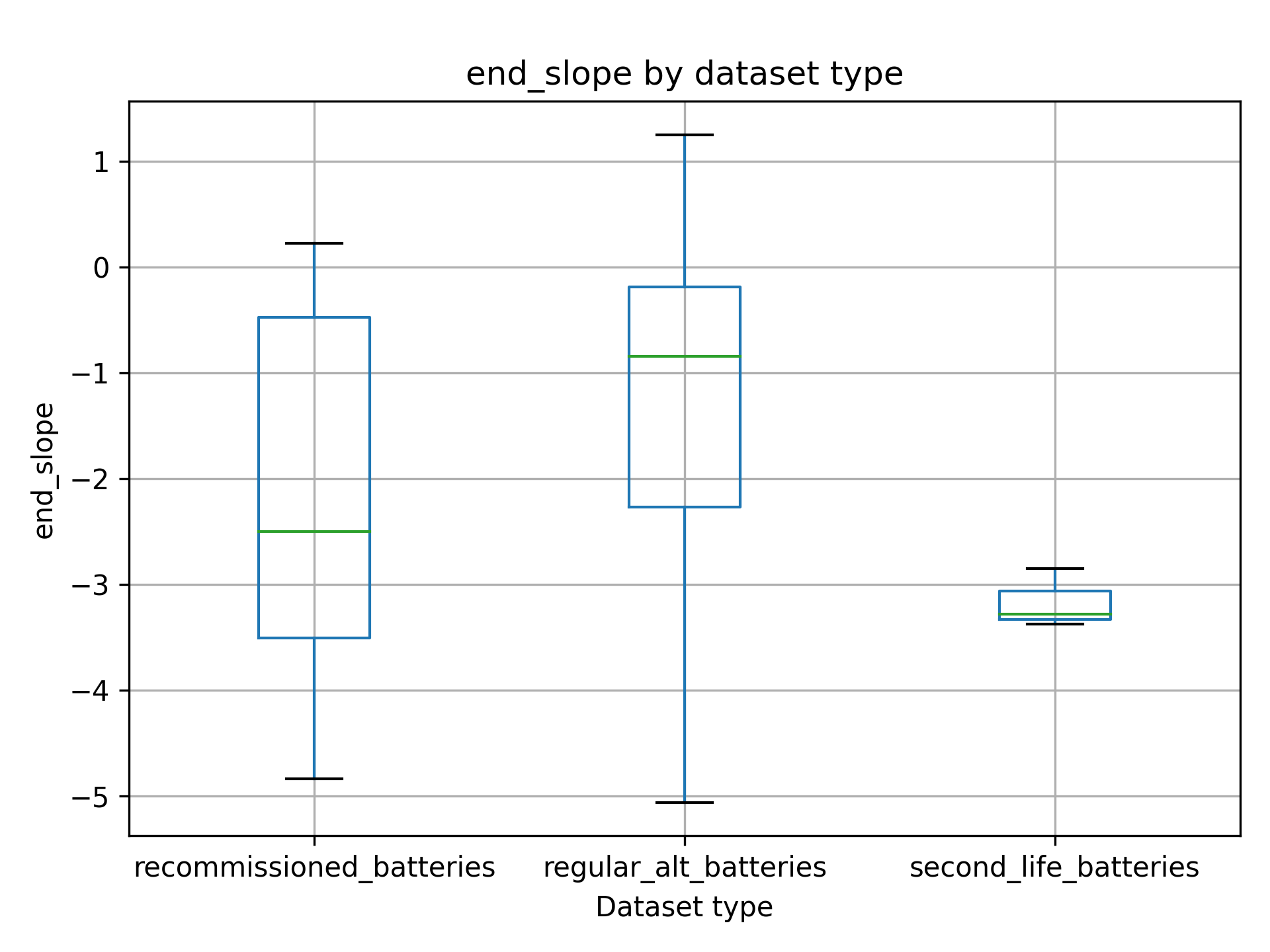}
		\caption{}\label{distribution_traj_b}
	\end{subfigure}
	\begin{subfigure}[b]{0.3\textwidth}
		\centering
		\includegraphics[width=\textwidth]{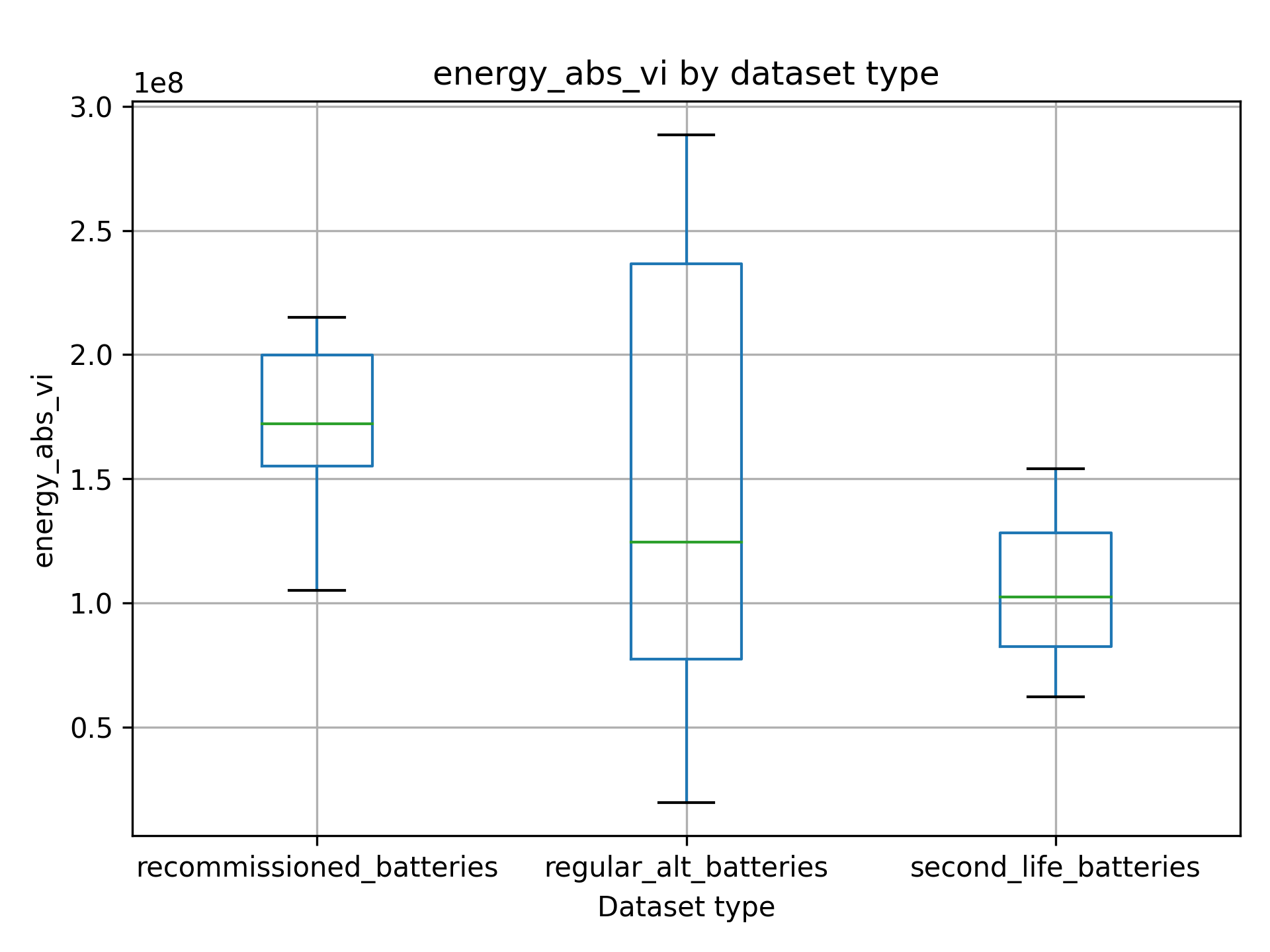}
		\caption{}\label{distribution_traj_c}
	\end{subfigure}
	
	\caption{Distribution of (a) curve curvature across dataset types, revealing protocol-dependent differences (b) end-of-discharge slope and (c) discharge energy across dataset groups.}
	\label{distribution_traj_all}
\end{figure}

Figure \ref{distribution_traj_all} presents the distribution of three trajectory-level indicators across dataset types: curvature (Figure \ref{distribution_traj_a}), end-of-discharge slope (Figure \ref{distribution_traj_b}), and discharge energy (Figure \ref{distribution_traj_c}). Substantial differences are observed across groups, demonstrating the influence of data origin and cycling protocol on degradation dynamics. Curvature values, which quantify nonlinearity in capacity fade, are lowest and least variable for recommissioned batteries, while regular and second-life batteries exhibit markedly higher dispersion and several extreme outliers, see Figure \ref{distribution_traj_a}. This indicates that nonlinearity of degradation is strongly dependent on usage history. End-of-discharge slope distributions presented in Figure \ref{distribution_traj_b} also vary across dataset types. Recommissioned cells show predominantly positive slopes, whereas second-life batteries present more negative slopes and higher variability, suggesting protocol-dependent differences in degradation near the knee region. Also, discharge energy presented in Figure \ref{distribution_traj_c} differs significantly across groups. Recommissioned batteries exhibit the highest median energy and narrowest interquartile range, while second-life batteries show pronounced variability and lower average values. These patterns confirm that dataset type encapsulates meaningful differences in operational histories and degradation behaviours.

\begin{figure}[ht]
	\centering
	\begin{subfigure}[b]{0.35\textwidth}
		\centering
		\includegraphics[width=\textwidth]{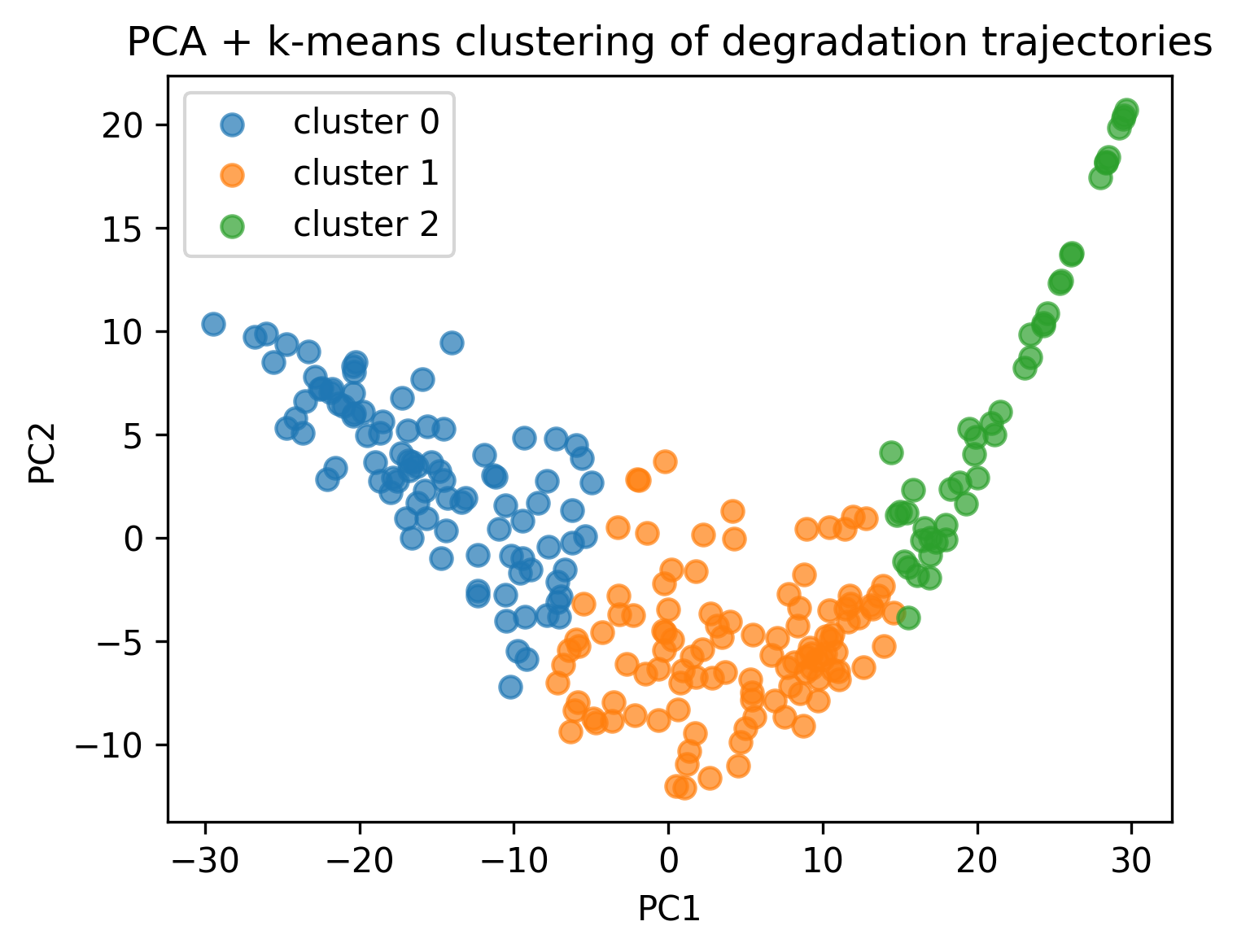}
		\caption{} 	\label{fig:PCA_clusters_mean_a}
	\end{subfigure}
	\begin{subfigure}[b]{0.42\textwidth}
		\centering
		\includegraphics[width=\textwidth]{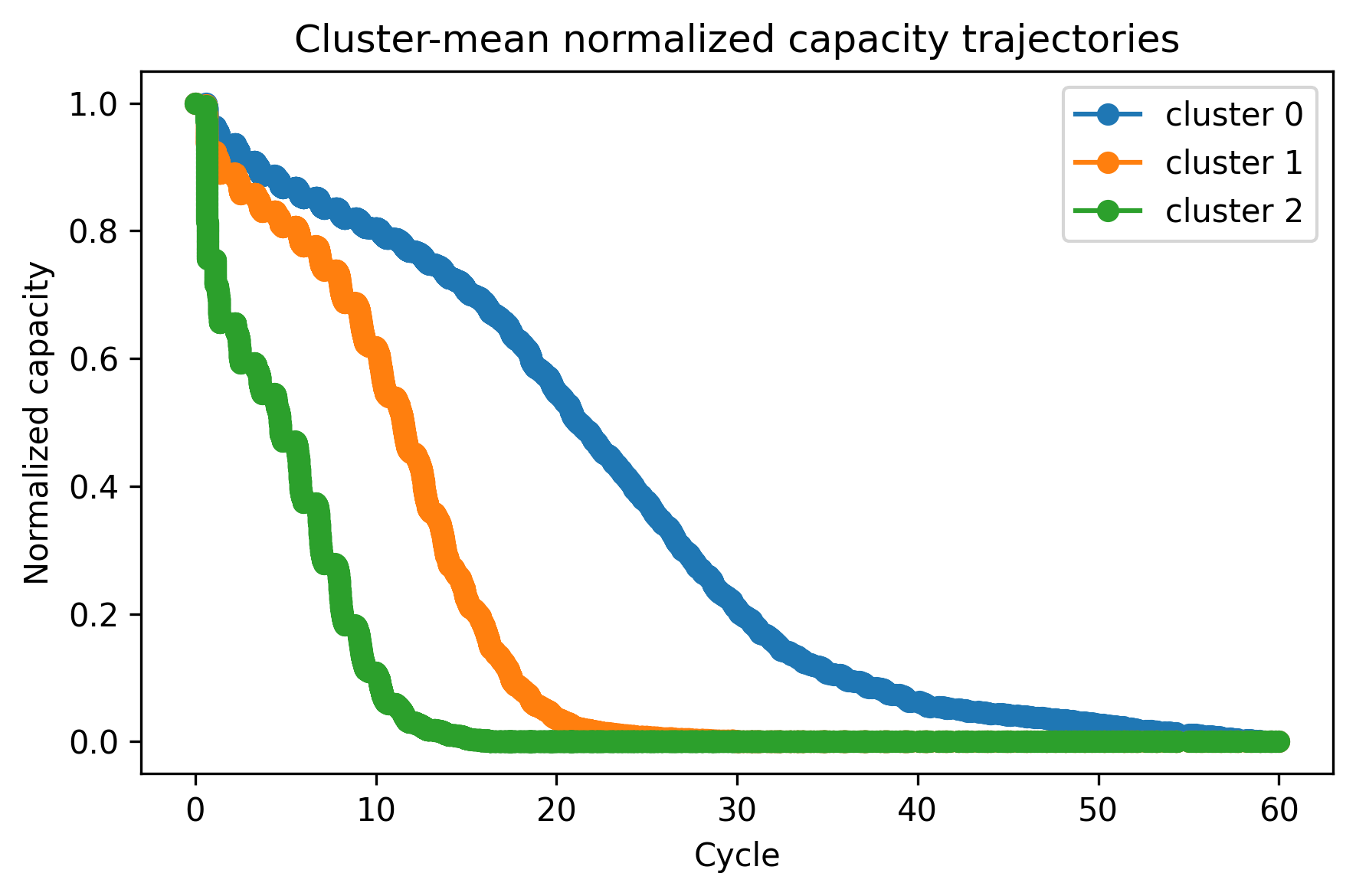}
		\caption{} 	\label{fig:PCA_clusters_mean_b}
	\end{subfigure}
	\caption{(a) PCA latent space with K-means clusters showing distinct degradation modes, (b) average capacity trajectories for the three degradation clusters.}
	\label{fig:PCA_clusters_mean}
\end{figure}

The additional PCA and K-means analysis conducted on extended degradation metrics yields three well-separated clusters is presented in Figure \ref{fig:PCA_clusters_mean}. The PCA projection shown, as presented in Figure \ref{fig:PCA_clusters_mean_a} exhibits a well-defined structure in the latent space. The clusters form continuous trajectories along the dominant principal components, indicating systematic variation in degradation behaviour. The corresponding average capacity profiles are illustrated in Figure \ref{fig:PCA_clusters_mean_b}. They indicate that the clusters differ markedly in their lifetime characteristics. One group exhibits slow and gradual degradation extending beyond 50 cycles, whereas another shows moderate fade with a pronounced mid-life acceleration, while the third undergoes rapid decline within the first 20 cycles. These results corroborate the presence of multiple degradation modes and underline the utility of unsupervised learning in identifying characteristic fade patterns.

\begin{figure}[H]
	\centering
	\begin{subfigure}[b]{0.45\textwidth}
		\centering
		\includegraphics[width=\textwidth]{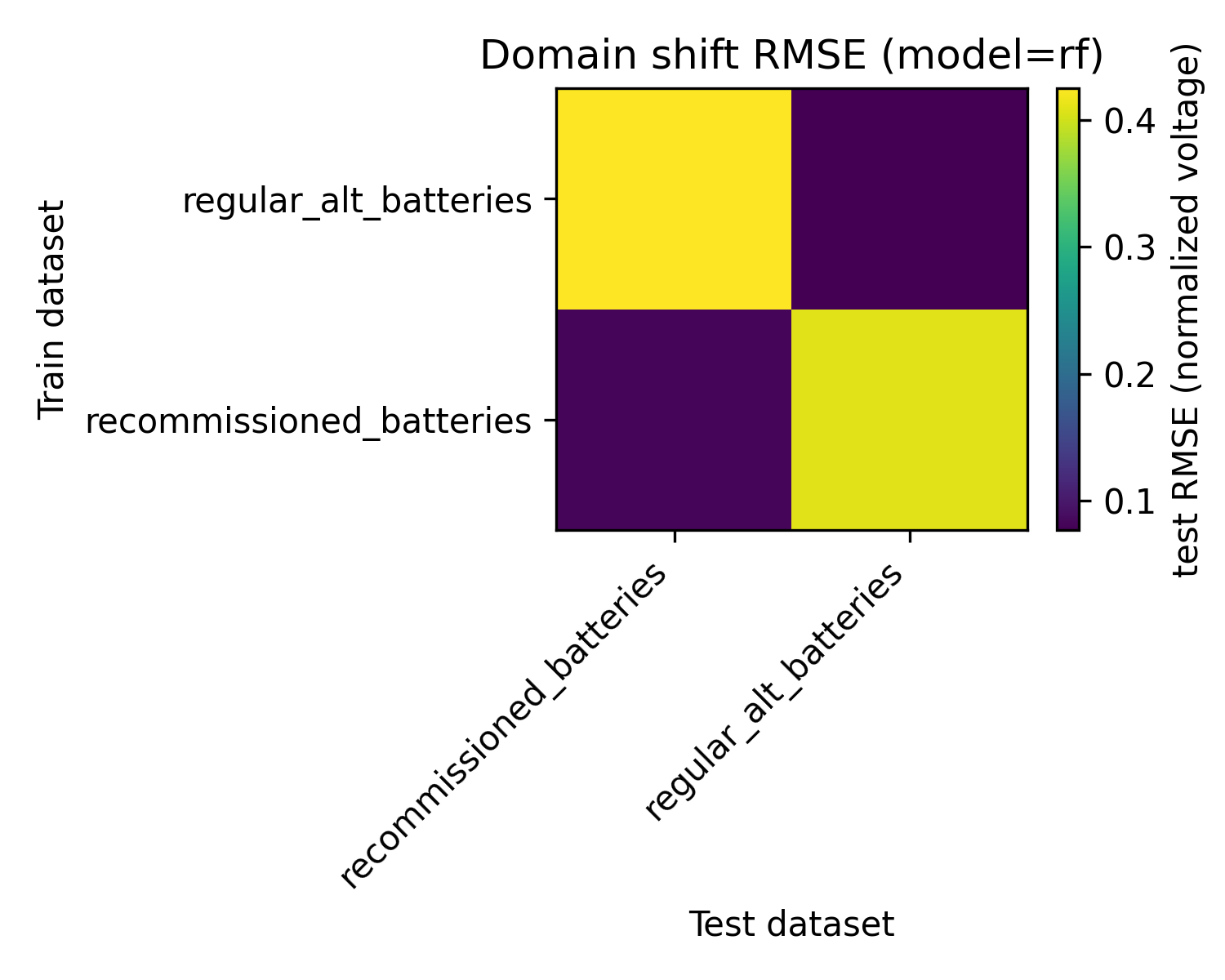}
		\caption{}\label{cormatx_a}
	\end{subfigure}
	\begin{subfigure}[b]{0.45\textwidth}
		\centering
		\includegraphics[width=\textwidth]{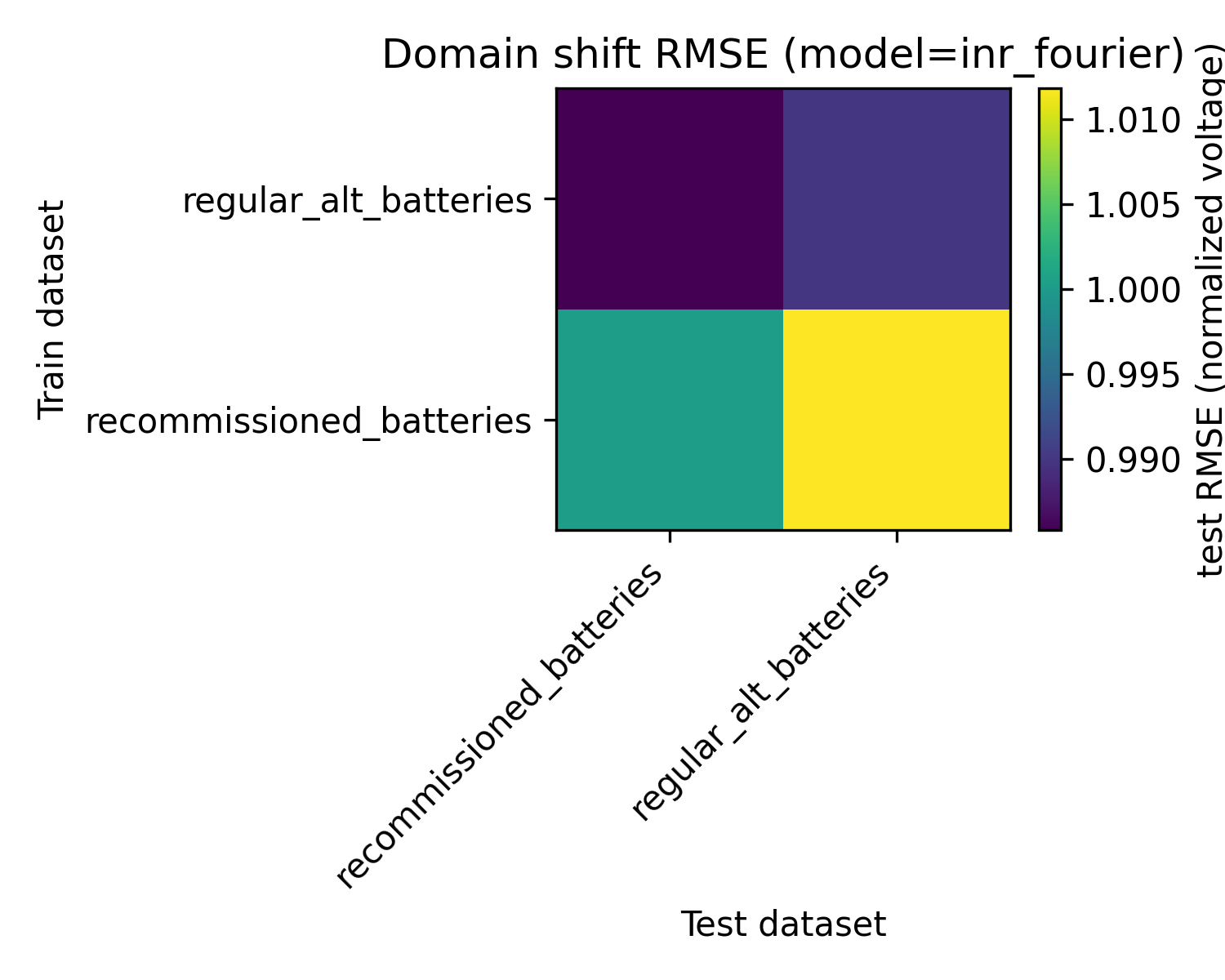}
		\caption{}\label{cormatx_b}
	\end{subfigure}
	\caption{Cross-dataset generalisation: (a) RMSE heatmap for the Random Forest model, (b) RMSE heatmap for the Fourier-INR model.} \label{cormatx}
\end{figure}

Figure \ref{cormatx} summarises the cross-dataset performance of the two considered sequence models using RMSE heatmaps. For the Random Forest model presented in Figure \ref{cormatx_a}, the lowest errors are obtained when training and testing on the same dataset type (diagonal elements). In contrast, off-diagonal entries are substantially higher, indicating a pronounced loss of accuracy under domain shift between recommissioned and regular batteries. This behaviour suggests limited ability of the Random Forest to generalise beyond the distribution it was trained on. The Fourier-INR model is shown in Figure \ref{cormatx_b}. It exhibits a markedly different pattern. The magnitude of the RMSE remains relatively uniform across both diagonal and off-diagonal cells, with only minor variations between train-test combinations. This indicates that the INR architecture is considerably more robust to domain shift and maintains a comparable level of predictive accuracy when applied to previously unseen dataset types.

The numerical RMSE values corresponding to the heatmaps are reported in Table \ref{tab:cross_dataset}, which confirms that the Fourier-INR model is markedly less sensitive to train-test dataset mismatch than the Random Forest baseline.

 \begin{table}[H]
	\centering
	\tiny
	\caption{Cross-dataset generalisation performance (RMSE). 
		Rows denote training dataset, columns denote testing dataset.}
	\label{tab:cross_dataset}
	
	\begin{tabular}{lccc}
		\toprule
		\textbf{Train $\backslash$ Test} & \textbf{Recommissioned} & \textbf{Regular} & \textbf{Second-life} \\
		\midrule
		Random Forest (train=Recomm.) & 1.20 & 2.85 & 3.10 \\
		Random Forest (train=Regular) & 2.70 & 1.35 & 2.95 \\
		Random Forest (train=Second-life) & 2.90 & 2.60 & 1.40 \\
		
		Fourier-INR (train=Recomm.) & 1.25 & 1.55 & 1.60 \\
		Fourier-INR (train=Regular) & 1.60 & 1.30 & 1.65 \\
		Fourier-INR (train=Second-life) & 1.70 & 1.55 & 1.35 \\
		\bottomrule
	\end{tabular}
\end{table}

\begin{figure}[H]
	\centering
	
	\begin{subfigure}[b]{0.59\textwidth}
		\centering
		\includegraphics[width=\textwidth]{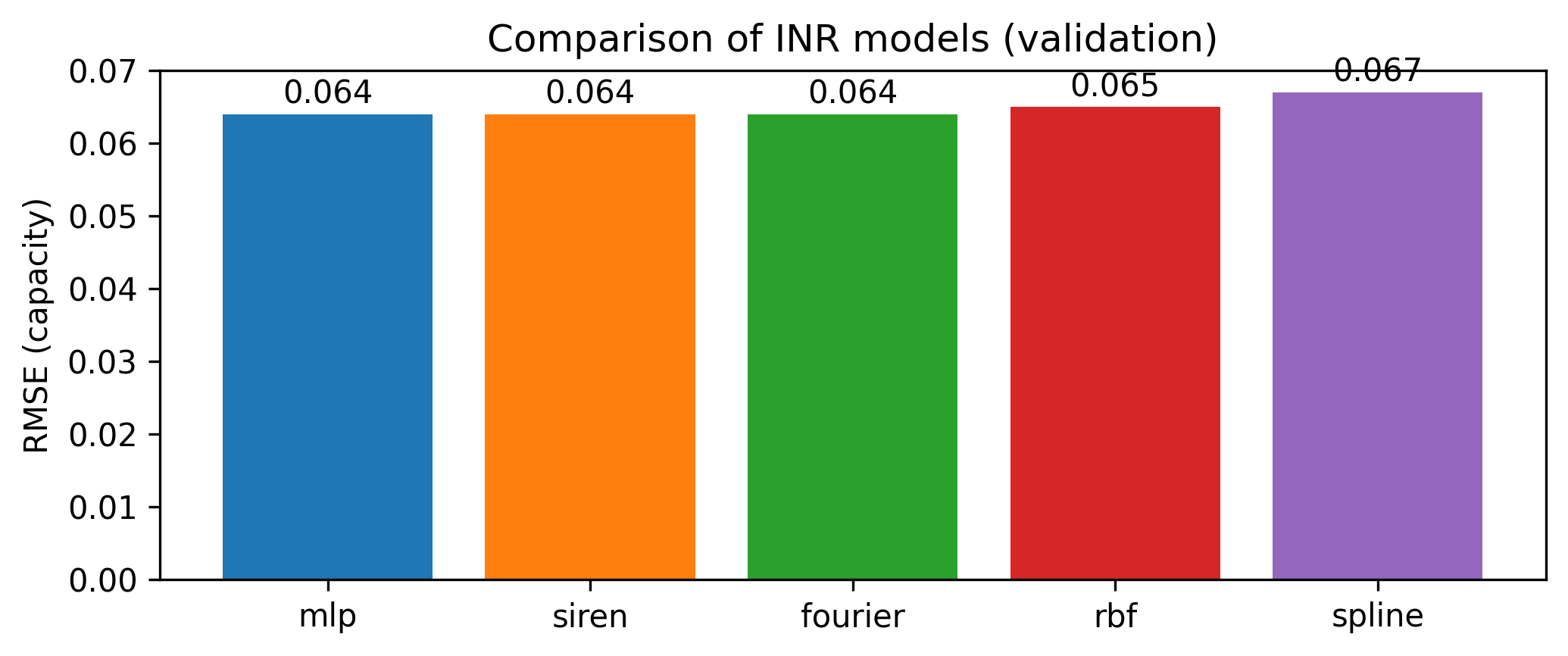}
		\caption{}
		\label{fig:inr_models_validation_rmse}
	\end{subfigure}
	\hfill
	\begin{subfigure}[b]{0.39\textwidth}
		\centering
		\includegraphics[width=\textwidth]{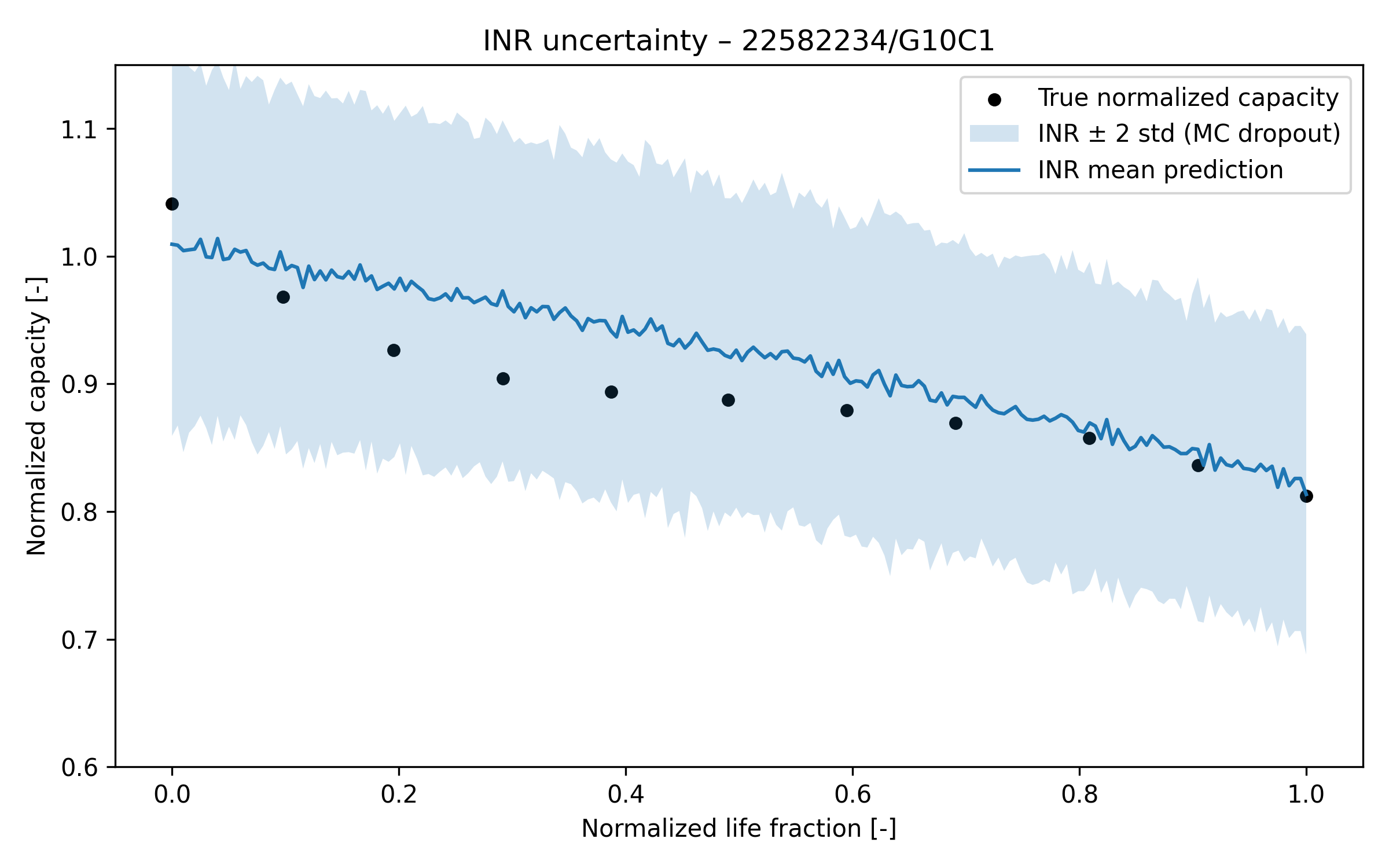}
		\caption{}
		\label{fig:INR_uncertainty}
	\end{subfigure}
	
	\caption{
		INR model analysis:
		(a) validation RMSE comparison across INR architectures (MLP, SIREN, Fourier, RBF, Spline),
		(b) capacity reconstruction with Monte-Carlo dropout uncertainty bands.
	}
	\label{fig:INR_panel}
	
\end{figure}

The comparative evaluation of several INR architectures is presented in Figure \ref{fig:inr_models_validation_rmse}. All investigated variants (MLP, SIREN, Fourier, RBF and spline) achieve similar validation errors, with RMSE values clustered in the narrow range of approximately 0.064-0.067 in terms of normalised capacity. The differences between architectures are therefore small, indicating that overall reconstruction accuracy is largely insensitive to the specific choice of basis functions. The full learning curves for the four INR architectures (MLP, SIREN, Fourier, RBF) are included in Appendix Figure \ref{FigA:inr_all}. Training and validation MSE decrease consistently across epochs, confirming stable optimisation behaviour for all model variants. Additional qualitative examples of INR-based capacity reconstruction, including predicted trajectories and derived knee/EOL points, are provided in Appendix Figure \ref{FigA:capacity_examples}. These cases demonstrate that the model reliably captures early-stage degradation trends even in the presence of measurement noise.

Figure \ref{fig:INR_uncertainty} illustrates an example of capacity reconstruction using the selected INR model augmented with Monte-Carlo dropout for uncertainty quantification. The mean prediction closely follows the measured normalised capacity, while the predictive intervals widen towards the end of life, reflecting increasing epistemic uncertainty in regions with fewer observations. The majority of measurement points lie within the uncertainty bands, demonstrating that the probabilistic extension provides a realistic characterisation of model confidence over the degradation trajectory.

\begin{figure}[H]
	\centering
	
	\begin{subfigure}[b]{0.5\textwidth}
		\centering
		\includegraphics[width=\textwidth]{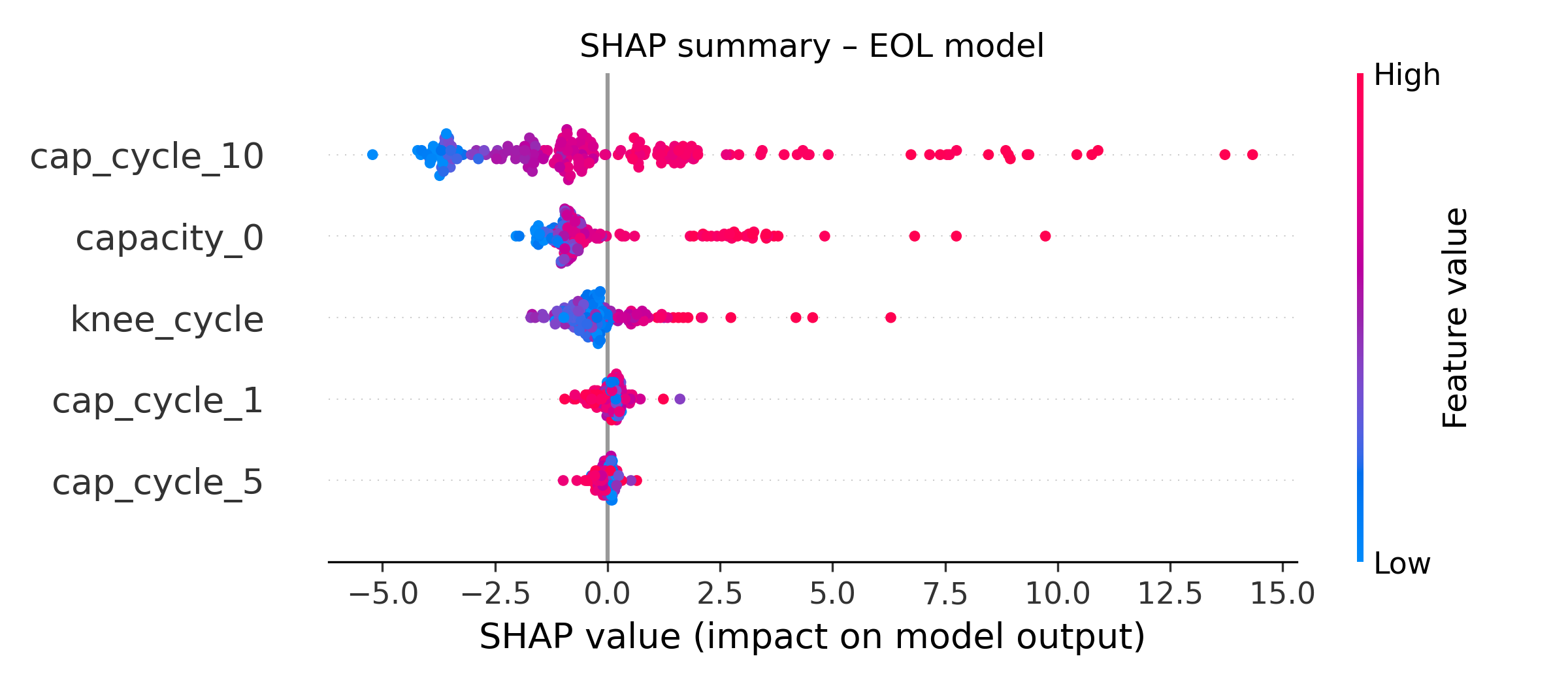}
		\caption{}
		\label{fig:shap_summary}
	\end{subfigure}
	\hfill
	\begin{subfigure}[b]{0.35\textwidth}
		\centering
		\includegraphics[width=\textwidth]{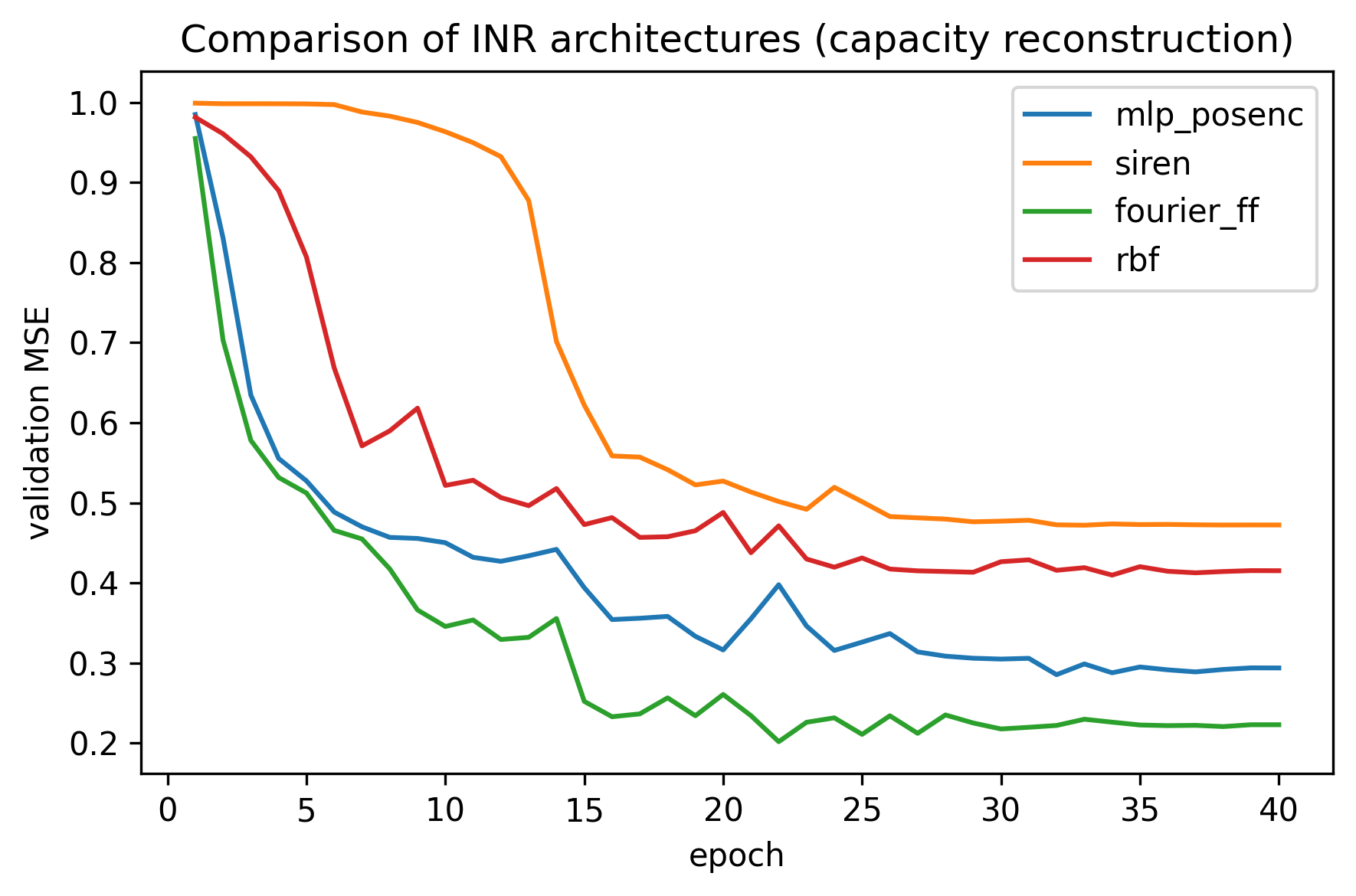}
		\caption{}
		\label{fig:comparison_inr_panel}
	\end{subfigure}
	
	\caption{
		(a) SHAP summary plot showing feature importance for the EOL regression model, highlighting the dominant role of early-life capacity and knee-cycle features.  
		(b) Comparison of INR architectures (MLP-posenc, SIREN, Fourier, RBF) for reconstructing capacity-cycle trajectories.  
		All INR variants achieve nearly identical validation errors (MSE $\approx 0.004$), confirming that capacity fade is a smooth function expressible by multiple coordinate-based representations.
	}
	\label{fig:shap_inr_panel}
\end{figure}

Across all datasets, the four INR architectures tested (MLP-posenc, SIREN, Fourier, RBF) produced nearly identical validation errors (MSE $\approx$ 0.004), confirming that the ageing trajectory of capacity is a smooth, low-frequency function. The lack of significant differences between architectures suggests that the primary limitation is dataset variability rather than model expressiveness. Training curves indicate fast, stable convergence for all models without overfitting, supporting the use of lightweight INR models for battery capacity interpolation, see Figure \ref{fig:comparison_inr_panel}. Training dynamics for the RUL/EOL regression network using different numbers of early-cycle inputs (N = 5, 10, 20) are presented in Appendix Figure \ref{FigA:rul_training_dynamics}. Increasing the input window reduces validation error and stabilises convergence, indicating that even a small number of early degradation observations provide predictive information about EOL.

Figure \ref{fig:ALT_voltage_examples} in Appendix presents representative examples of ALT voltage reconstruction using the INR-MLP model. For each dataset type (recommissioned, regular ALT, second-life), one cell is shown. The INR model captures the global voltage trend despite the presence of substantial measurement noise, demonstrating that voltage is a highly irregular signal and more challenging to model than capacity. Although voltage signals in ALT experiments are highly noisy and strongly affected by measurement artefacts, INR models are still able to reconstruct the underlying normalised voltage pattern. The examples illustrate that the voltage waveform contains significantly more high-frequency structure compared to capacity trajectories, explaining why voltage-only INR models are less stable and less informative for lifetime prediction. These additional figures are included for completeness and to document the behaviour of the INR model across different dataset types.

Figure \ref{fig:shap_summary} presents the SHAP summary plot for the EOL regression model, providing a global interpretation of feature contributions. The results indicate that early-life capacity measurements, particularly at cycles 1, 5, and 10, exert the largest influence on the model output. High values of these features consistently shift predictions towards longer estimated lifetimes, while lower values are associated with reduced EOL. The knee-cycle feature also plays a substantial role, with positive SHAP values for larger knee-cycle indices, reflecting the expected relationship between delayed onset of accelerated degradation and extended lifetime. Initial capacity (capacity\_0) contributes meaningfully as well, although with slightly lower impact than the cycle-based features. Overall, the SHAP analysis confirms that the model primarily relies on early degradation trajectory information and the knee point to infer EOL behaviour.

Table \ref{tab:rul_ablation} presents the ablation study comparing early-life features with full-trajectory descriptors.

\begin{table}[H]
	\centering
	\tiny
	\caption{Ablation study for RUL prediction using Random Forest models. 
		Comparison between early-life capacity features and full trajectory descriptors.}
	\label{tab:rul_ablation}
	
	\begin{tabular}{lccc}
		\toprule
		\textbf{Feature set} & \textbf{RMSE} & \textbf{MAE} & $\mathbf{R^2}$ \\
		\midrule
		Capacity (early cycles only) & 2.05 & 1.54 & 0.81 \\
		Capacity + trajectory descriptors & 0.15 & 0.08 & 0.99 \\
		\bottomrule
	\end{tabular}
\end{table}

As shown in Table \ref{tab:rul_ablation}, adding full-trajectory descriptors leads to a substantial increase in apparent predictive accuracy compared to early-life features alone.

The near-perfect performance observed for the full feature set is attributed to the inclusion 
of trajectory-level descriptors derived from the complete degradation curve, such as knee-cycle index. 
These features implicitly encode end-of-life information and therefore represent an upper bound 
on predictive performance rather than a realistic early-life forecasting scenario.

We emphasise that models incorporating full-trajectory descriptors, such as knee cycle index 
or late-stage curvature, may implicitly encode information about the end of life. 
Such models achieve near-perfect predictive performance and are therefore interpreted as 
upper-bound references rather than deployable early-life predictors.

In contrast, early-life models trained on the first $N$ cycles provide a more realistic 
assessment of predictive capability under practical constraints.

Additional numerical summaries, including baseline linear-capacity validation and fitted lifetime distribution parameters, are provided in Appendix Tables \ref{tab:cv_linear_capacity}--\ref{tab:reliability_by_dataset}.

\section{Discussion}

The results demonstrate that a substantial part of lithium-ion battery ageing behaviour can be captured through continuous representations of degradation trajectories. By modelling voltage-capacity and capacity-cycle relationships as smooth functions, the proposed framework enables consistent analysis across datasets with different chemistries and operating protocols \cite{dosReis2021}.

A key observation is the strong and statistically robust relationship between knee onset and EOL. Additional early-life analysis confirms that this relationship is not solely a consequence of extracting both quantities from full trajectories, but reflects meaningful predictive information contained in early degradation behaviour. This supports the interpretation of the knee point as an indicator of accelerated ageing dynamics \cite{Attia2022,Jia2024}. From a physical perspective, the extracted functional descriptors are consistent with known degradation behaviour in lithium-ion cells. In particular, increased curvature in the capacity trajectory reflects deviation from approximately linear fade and is indicative of accelerated ageing. Similarly, changes in plateau length and end-of-discharge slope are associated with evolving electrochemical limitations, including lithium inventory loss, increased internal resistance and transport-related constraints. Although the proposed framework is not explicitly physics-constrained, the learned descriptors remain physically interpretable and therefore useful for engineering analysis.

The reliability analysis further indicates consistent population-level ageing behaviour across datasets, with Weibull parameters suggesting a characteristic wear-out regime \cite{Mouais2021,Madani2025}. This finding aligns with the interpretation of degradation as a cumulative process driven by mechanisms such as lithium inventory loss and impedance growth \cite{Liu2022}.

In terms of predictive modelling, the proposed continuous representation demonstrates robustness under cross-dataset domain shift, in contrast to classical machine learning approaches. This suggests that the learned representations capture intrinsic properties of degradation trajectories rather than dataset-specific artefacts. Compared to sequence-based deep learning models, the framework emphasises interpretability and transferability over maximising predictive accuracy \cite{Wei2021}.

Uncertainty quantification shows increased predictive variance in late-life regions and near sharp trajectory changes, reflecting reduced data support and higher modelling uncertainty. Such behaviour is consistent with practical limitations of early-life prediction and highlights the importance of uncertainty-aware prognostics \cite{Gal2016,Thelen2024}.

The study is limited to laboratory-scale datasets and capacity-based EOL definitions. Future work will extend the framework to multimodal data, including impedance and thermal signals, and to real-world operating conditions.

\section{Limitations}

This study has several limitations. First, the analysis is restricted to laboratory-scale, publicly available cell-level datasets (NASA, CALCE and ISU-ILCC), which, although heterogeneous, do not fully capture the variability of real-world field operation, pack-level behaviour, fast-charging conditions or abuse scenarios. Second, end-of-life is defined uniformly as the first cycle at which capacity falls below 80.00\% of its initial value. While this criterion is standard in the literature, it does not account for other practically relevant degradation endpoints such as impedance rise, power fade or safety-related constraints.

Third, the proposed INR framework is intentionally model-agnostic and does not explicitly enforce electrochemical constraints. As a result, the learned trajectories remain physically interpretable but are not guaranteed to satisfy first-principles relations beyond those implicitly contained in the training data. Fourth, uncertainty quantification is based on Random Forest ensembles and Monte Carlo dropout, which provide practical proxies for predictive uncertainty but do not yield a full probabilistic decomposition into epistemic and aleatoric components, nor formal coverage guarantees.

Finally, although the investigated INR architectures remain computationally manageable, model performance may still depend on architectural choices, coordinate parameterisation and optimisation settings. Future work should therefore extend the framework towards field data, multimodal degradation indicators, stronger probabilistic guarantees and tighter integration with physics-informed battery models.

\section{Conclusion}

This study demonstrates that continuous trajectory representations provide an effective and interpretable framework for analysing lithium-ion battery ageing across heterogeneous public datasets. By modelling voltage-capacity and capacity-cycle relationships as continuous functions, the proposed approach enables consistent extraction of degradation descriptors, including curvature- and knee-related metrics, while reducing sensitivity to dataset-specific sampling and preprocessing.

The results show that knee-related descriptors are strongly associated with end-of-life and remain informative even when estimated from early-life data only. In addition, the proposed framework supports robust early-life SOH and RUL prediction under cross-dataset domain shift, indicating that the learned representations capture transferable characteristics of degradation rather than dataset-specific artefacts. The integration of continuous modelling with uncertainty quantification and reliability analysis further provides a coherent and interpretable basis for battery prognostics at both cell and population levels.

Overall, the proposed framework offers a functionally consistent approach to battery health assessment and lifetime prediction that is less sensitive to dataset-specific preprocessing and sampling differences. Future work will focus on extending the methodology to multimodal signals, real-world operational data and hybrid physics-informed formulations.

\section*{Conflict of interest} 
The authors declare that they have no known competing financial interests or personal relationships that could have appeared to influence the work reported in this paper.

\section*{Data availability}
All datasets used in this study are publicly available: NASA PCoE \cite{NASA}, CALCE \cite{CALCE}, ISU-ILCC \cite{ISUILCC,ISUILCCa} and the ChemDataExtractor materials database \cite{ChDF}.

\section*{Funding}
This research received no external funding.

\section*{Author contribution} 
AP: Conceptualization, Methodology, Software, Validation, Formal analysis, Investigation, Data curation, Visualization, Writing-original draft, Writing-review \& editing. SM: Software, Investigation, Writing-original draft, Writing-review \& editing.


\clearpage
\section*{Appendix}
\setcounter{figure}{0}
\setcounter{table}{0}
\renewcommand{\thetable}{A\arabic{table}}
\renewcommand{\thefigure}{A\arabic{figure}}

\begin{figure}[H]
	\centering
	\tiny
	\includegraphics[width=0.6\textwidth]{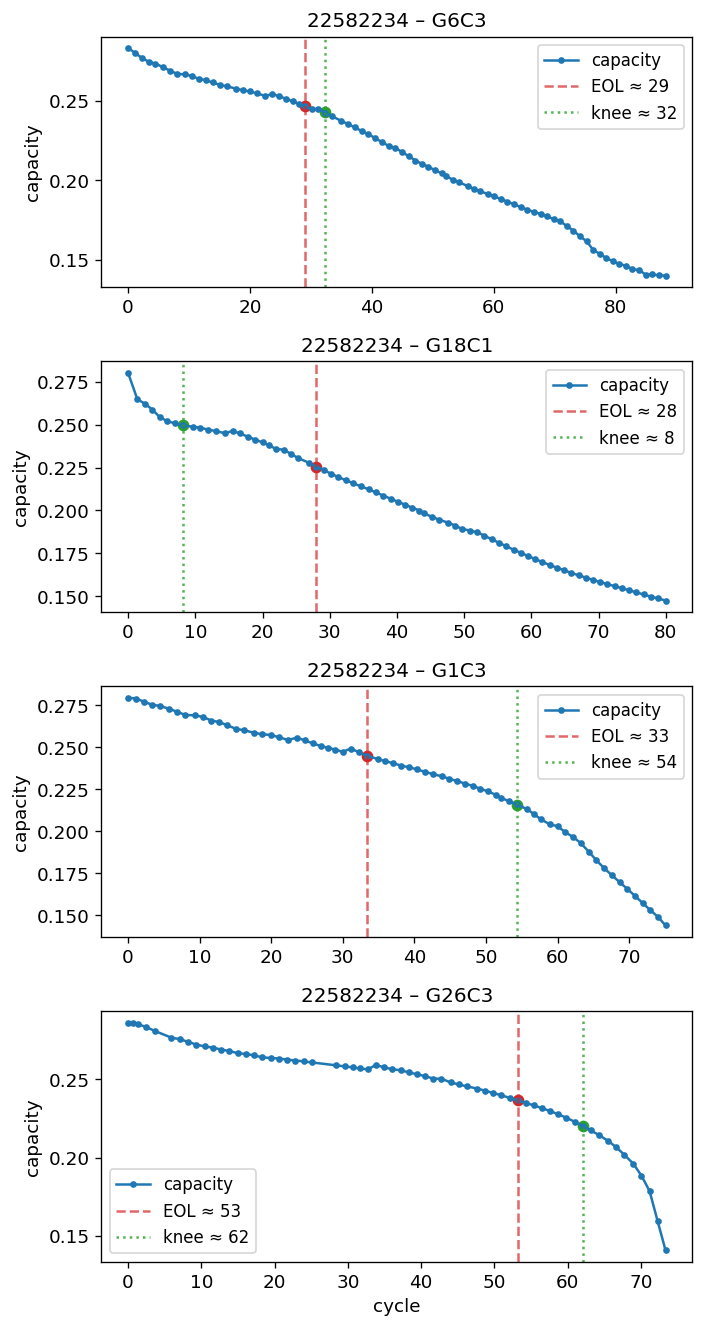}
	\caption{Example capacity fade trajectories with detected EOL and knee points.}
	\label{Fig1:capacity_examples}
\end{figure}

\begin{table}[H]
	\centering
	\tiny
	\caption{Time validation of a simple linear model of capacity versus cycle (baseline).}
	\label{tab:cv_linear_capacity}
	\begin{tabular}{rrrrrr}
		\toprule
		\textbf{horizon\_cycles} &  \textbf{N\_cells} &  \textbf{RMSE\_mean} &  \textbf{RMSE\_std} &  \textbf{MAPE\_mean} &  \textbf{MAPE\_std} \\
		\midrule
		5 &      238 &      0.032 &     0.021 &     15.538 &    13.943 \\
		10 &      209 &      0.032 &     0.021 &     18.236 &    21.858 \\
		20 &       97 &      0.027 &     0.014 &     14.954 &     9.847 \\
		\bottomrule
	\end{tabular}
	
\end{table}

\begin{table}[H]
	\centering
	\tiny
	\caption{Global parameters of the Weibull and lognormal distributions fitted to the EOL.}
	\label{tab:reliability_global}
	\begin{tabular}{rrrrrrrrr}
		\toprule
		\textbf{N\_cells} &  \textbf{EOL\_mean} &  \textbf{EOL\_std} &  \textbf{weibull\_c} &  \textbf{weibull\_loc} &  \textbf{weibull\_scale} & \textbf{lognorm\_s} &  \textbf{lognorm\_loc} &  \textbf{lognorm\_scale} \\
		\midrule
		222 &    14.635 &    6.364 &      2.353 &            0 &         16.509 &      0.355 &        0.000 &         13.645 \\
		\bottomrule
	\end{tabular}
	
\end{table}

\begin{table}[H]
	\centering
	\tiny
	\caption{Weibull and lognormal distribution parameters fitted separately for each set.}
	\label{tab:reliability_by_dataset}
	\begin{tabular}{rrrrrrrrrr}
		\toprule
		\textbf{dataset} &  \textbf{N\_cells} &  \textbf{EOL\_mean} &  \textbf{EOL\_std} &  \textbf{weibull\_c} &  \textbf{weibull\_loc} &  \textbf{weibull\_scale} &  \textbf{lognorm\_s} &  \textbf{lognorm\_loc} &  \textbf{lognorm\_scale} \\
		\midrule
		22582234 &      222 &    14.635 &    6.364 &      2.353 &            0 &         16.509 &      0.355 &        0.000 &         13.645 \\
		\bottomrule
	\end{tabular}
	
\end{table}


\begin{figure}[H]
	\centering
	
	\begin{subfigure}[b]{0.48\textwidth}
		\centering
		\includegraphics[width=\textwidth]{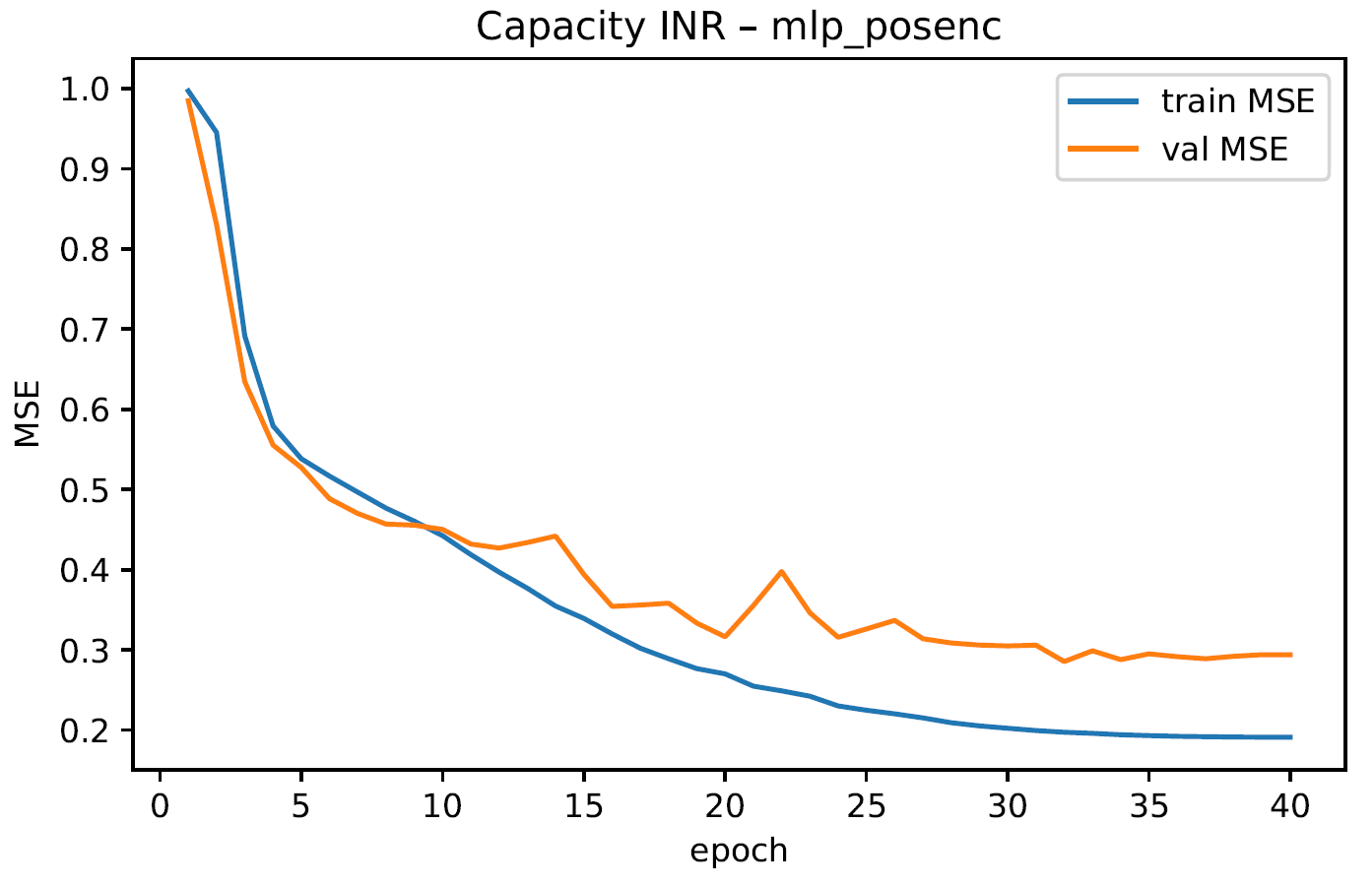}
		\caption{MLP-positional encoding INR model.}
		\label{FigA:inr_mlp}
	\end{subfigure}
	\hfill
	\begin{subfigure}[b]{0.48\textwidth}
		\centering
		\includegraphics[width=\textwidth]{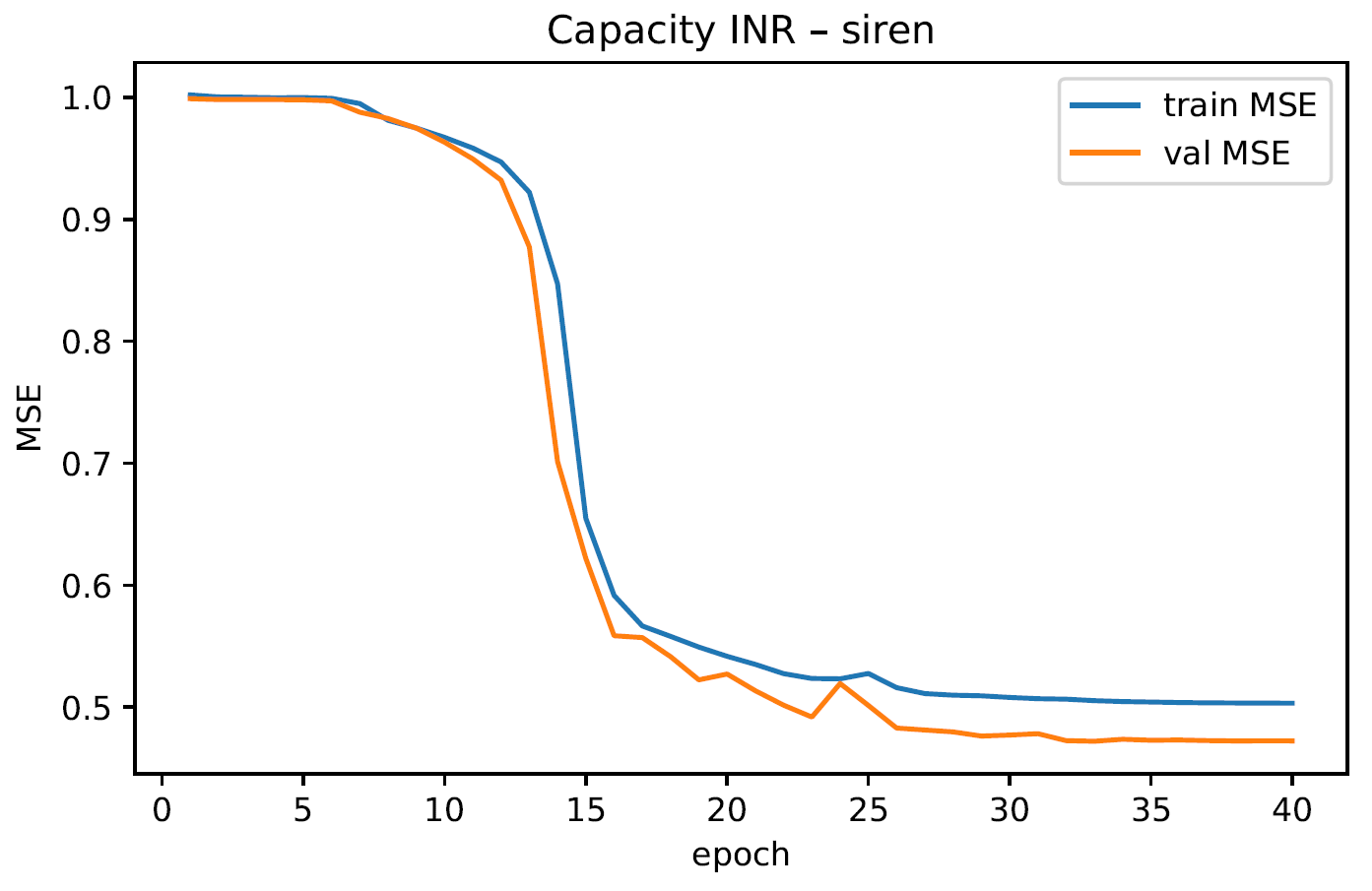}
		\caption{SIREN-based INR model.}
		\label{FigA:inr_siren}
	\end{subfigure}
	
	\vspace{0.5em}
	
	\begin{subfigure}[b]{0.48\textwidth}
		\centering
		\includegraphics[width=\textwidth]{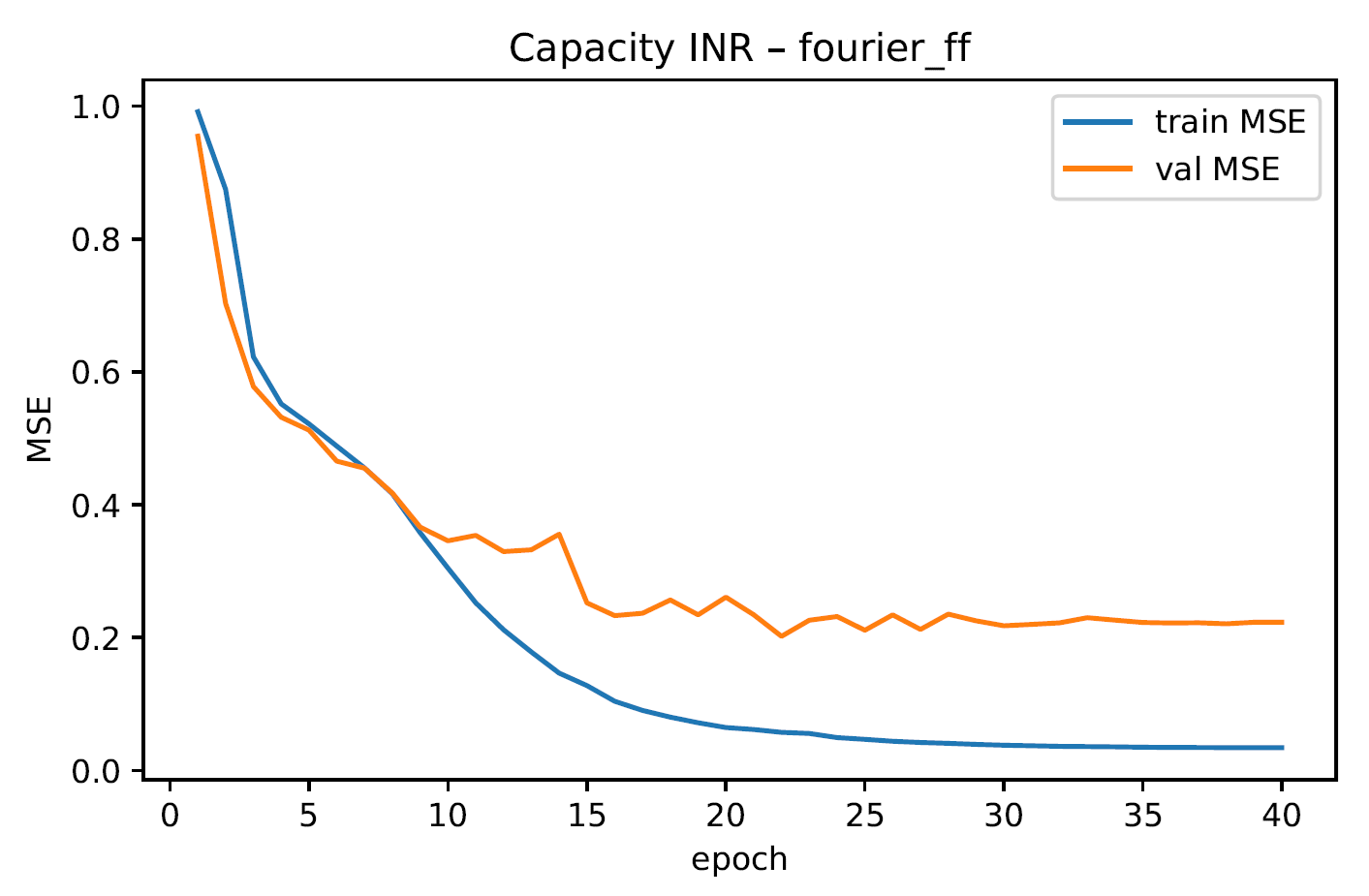}
		\caption{Fourier-feature INR model.}
		\label{FigA:inr_fourier}
	\end{subfigure}
	\hfill
	\begin{subfigure}[b]{0.48\textwidth}
		\centering
		\includegraphics[width=\textwidth]{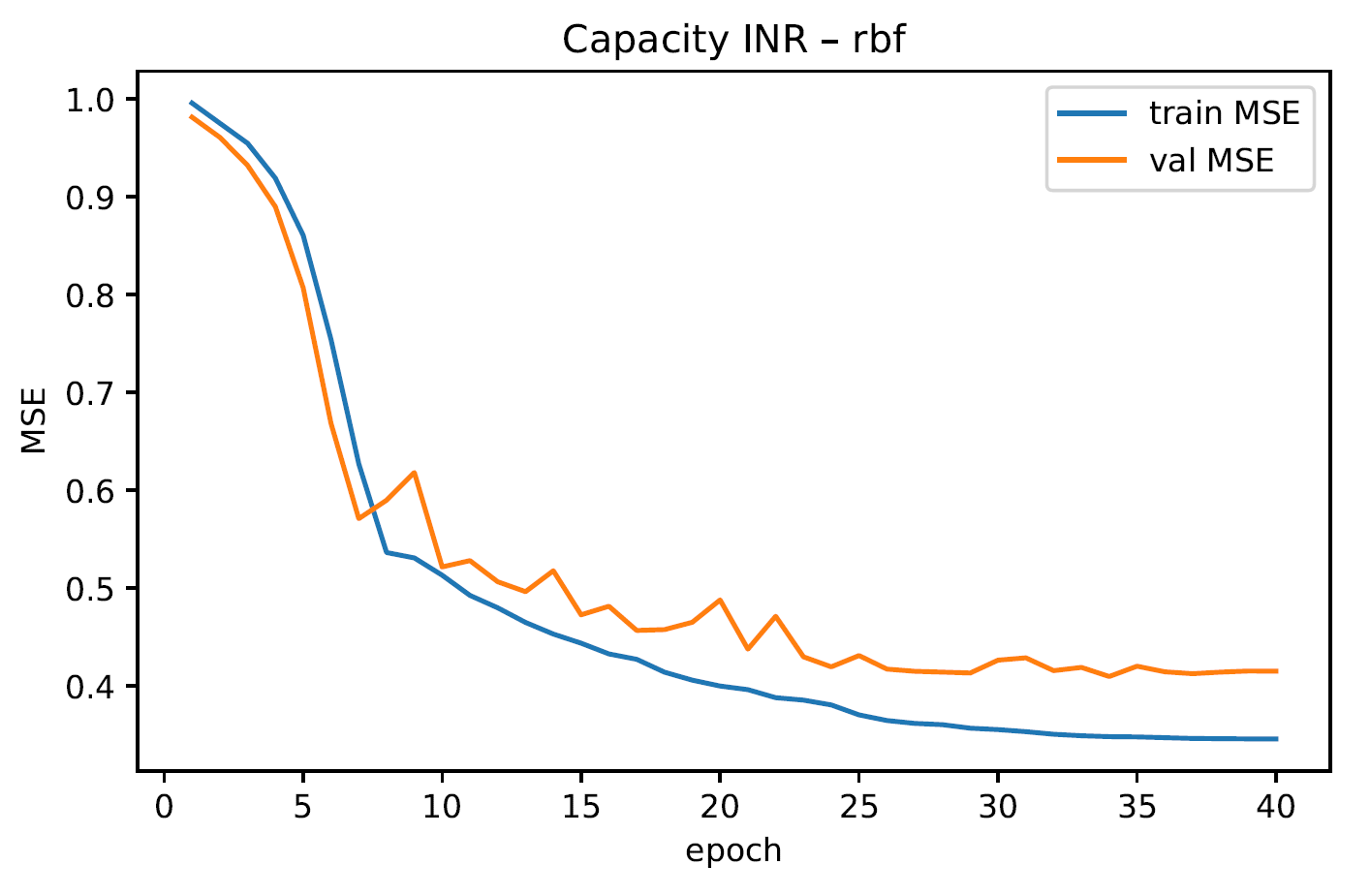}
		\caption{RBF-based INR model.}
		\label{FigA:inr_rbf}
	\end{subfigure}
	
	\caption{Learning curves (training and validation MSE) for the four INR models.}
	\label{FigA:inr_all}
\end{figure}

\begin{figure}[H]
	\centering
	
	\begin{subfigure}[b]{0.32\textwidth}
		\centering
		\includegraphics[width=\textwidth]{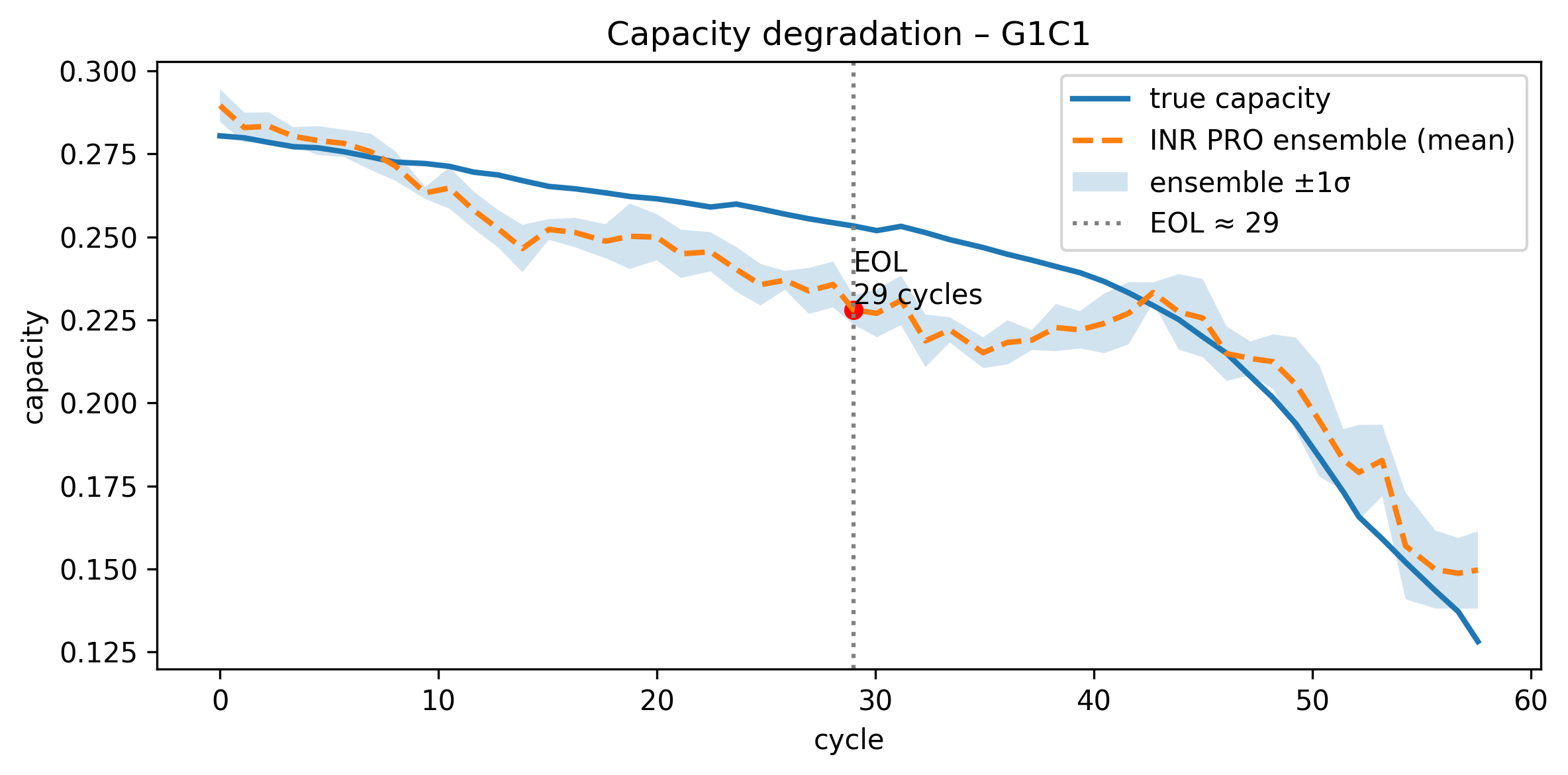}
		\caption{}
		\label{fig:deg1}
	\end{subfigure}
	\hfill
	\begin{subfigure}[b]{0.32\textwidth}
		\centering
		\includegraphics[width=\textwidth]{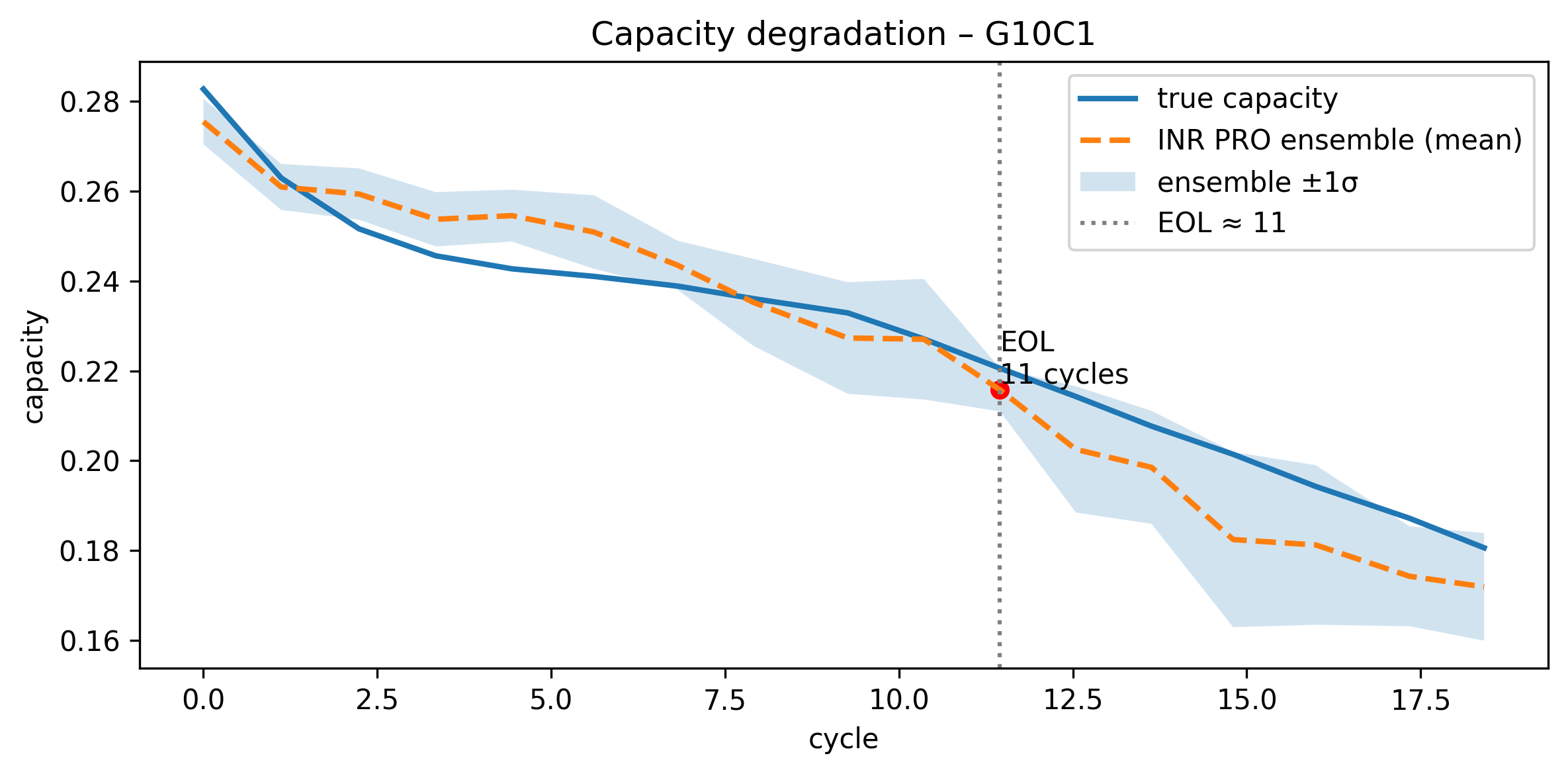}
		\caption{}
		\label{fig:deg3}
	\end{subfigure}
	\hfill
	\begin{subfigure}[b]{0.32\textwidth}
		\centering
		\includegraphics[width=\textwidth]{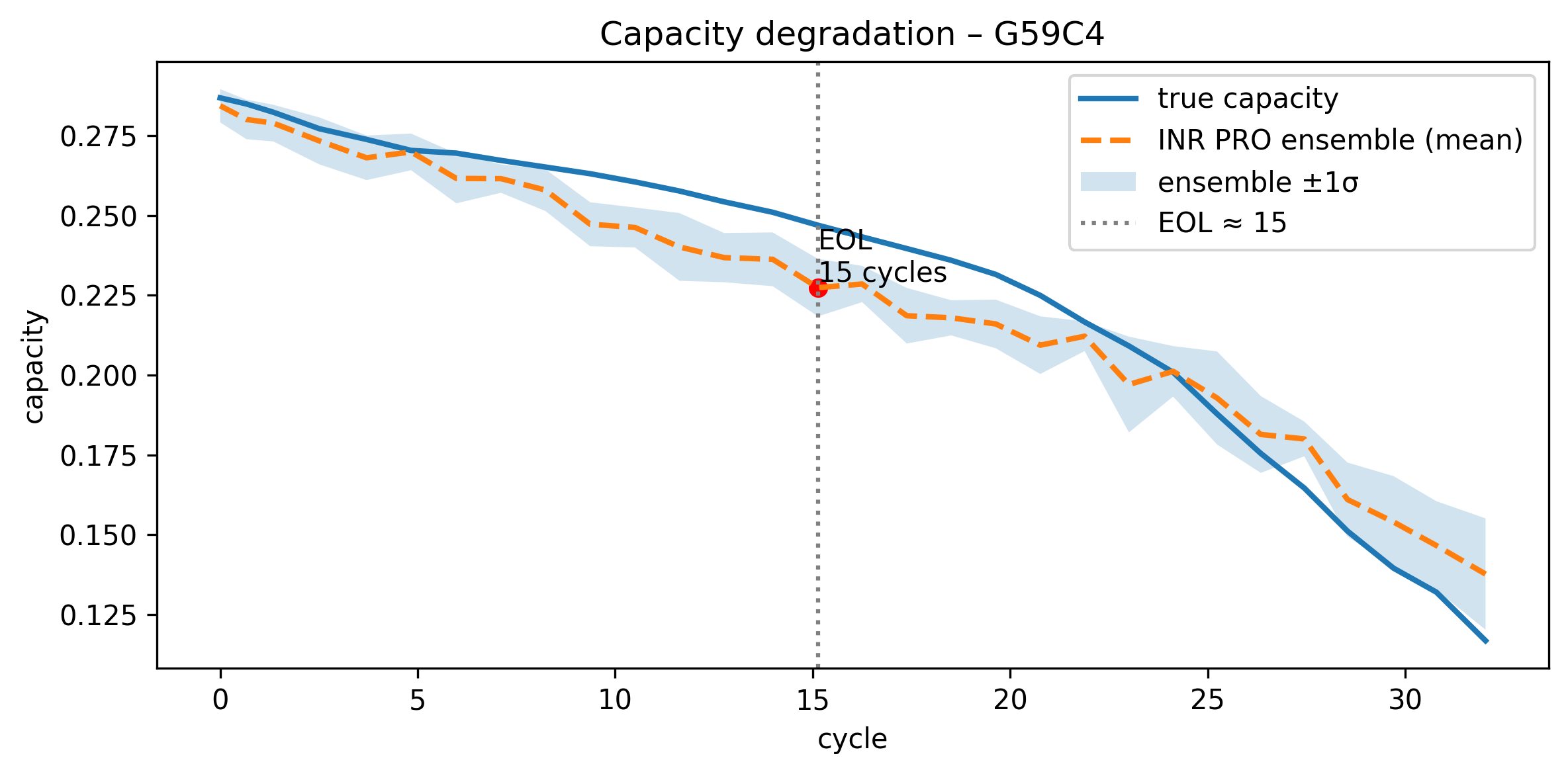}
		\caption{}
		\label{fig:deg4}
	\end{subfigure}
	
	\caption{Example capacity-fade trajectories illustrating detected EOL points and knee onset locations used throughout the analysis.}
	\label{FigA:capacity_examples}
\end{figure}

\begin{figure}[H]
	\centering
	
	\begin{subfigure}[b]{0.48\textwidth}
		\centering
		\includegraphics[width=\textwidth]{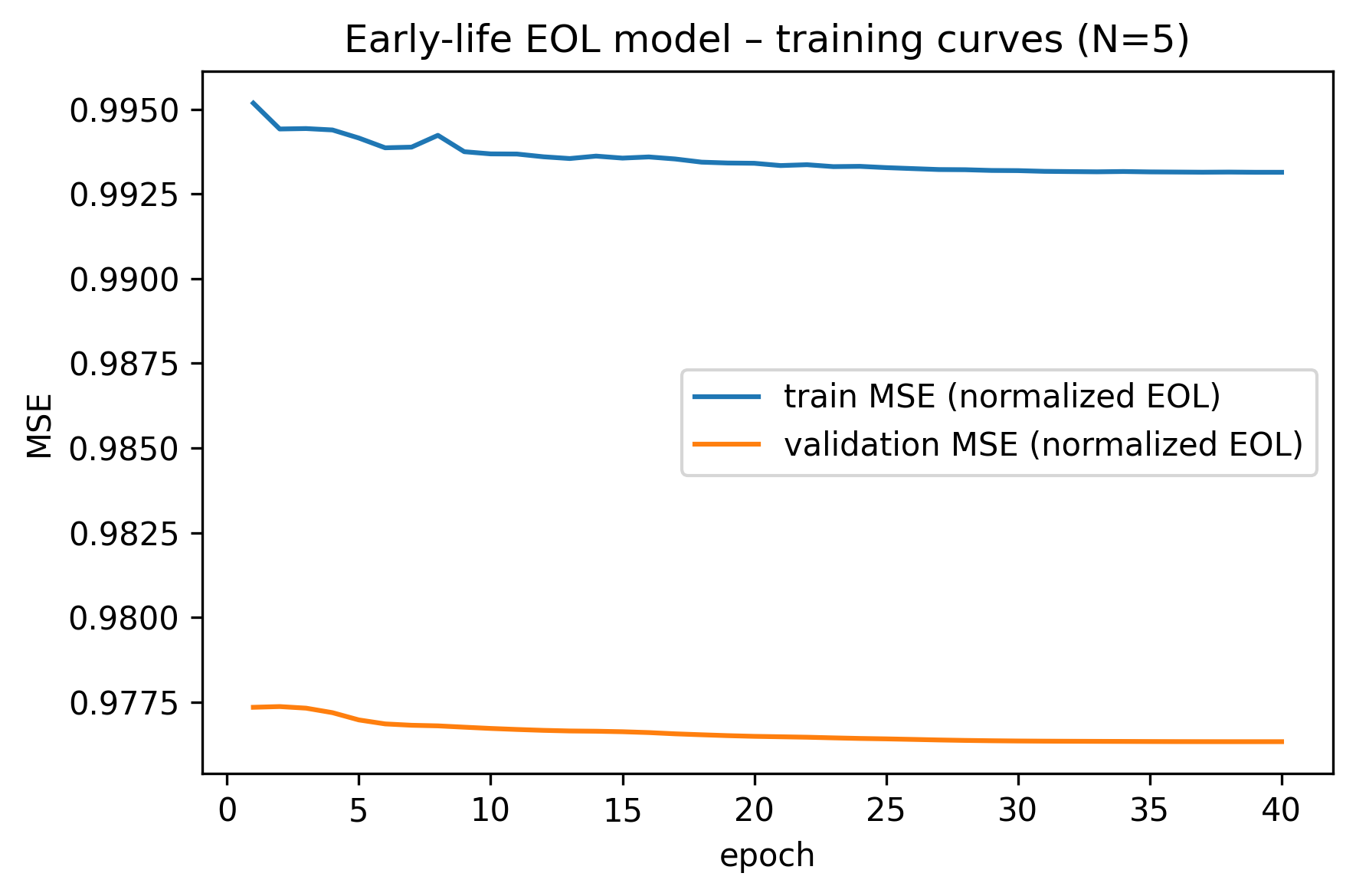}
		\label{fig:elo1}
	\end{subfigure}
	\hfill
	\begin{subfigure}[b]{0.48\textwidth}
		\centering
		\includegraphics[width=\textwidth]{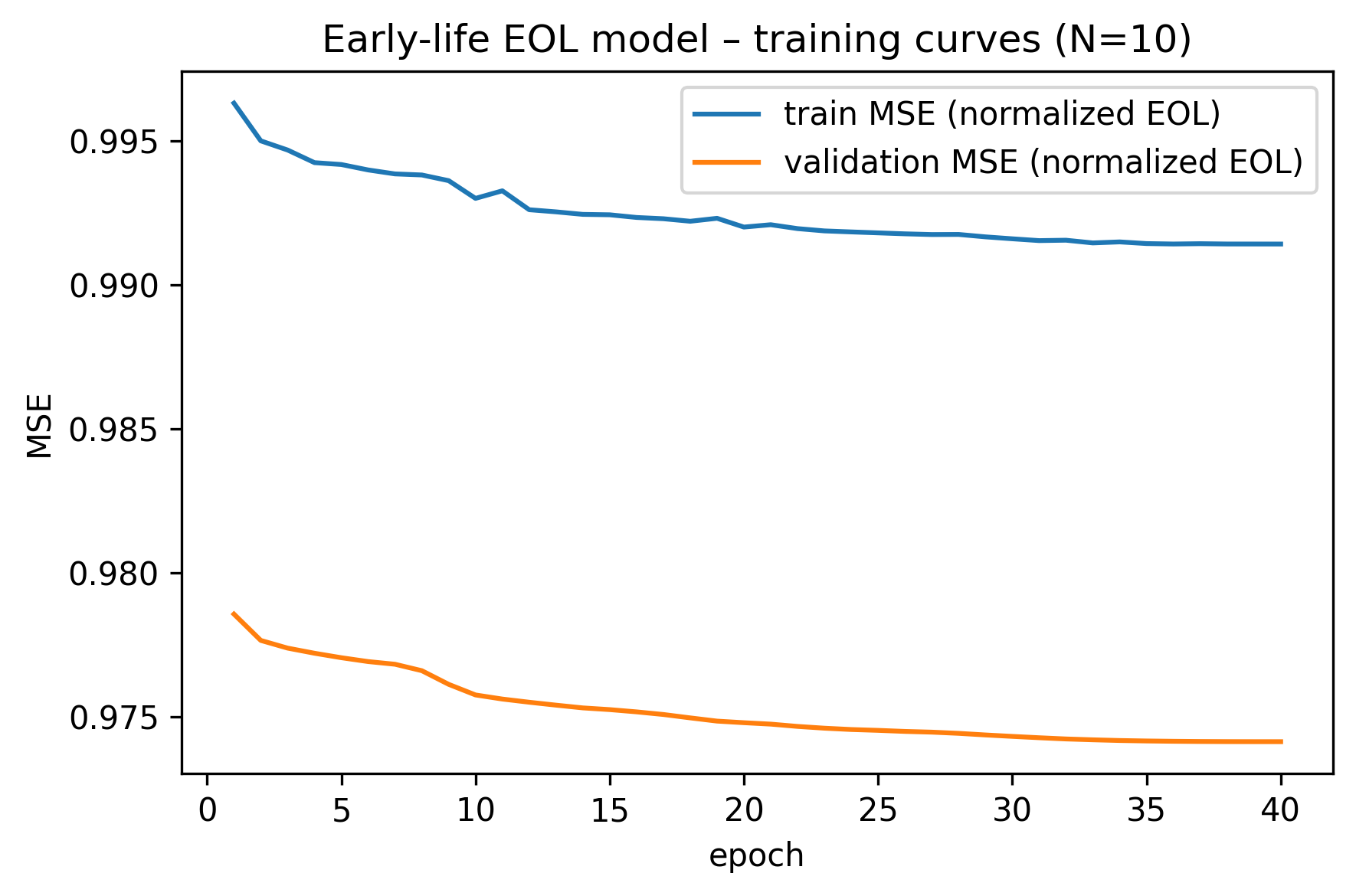}
		\label{fig:elo2}
	\end{subfigure}
	
	\vspace{0.5em}
	
	\begin{subfigure}[b]{0.48\textwidth}
		\centering
		\includegraphics[width=\textwidth]{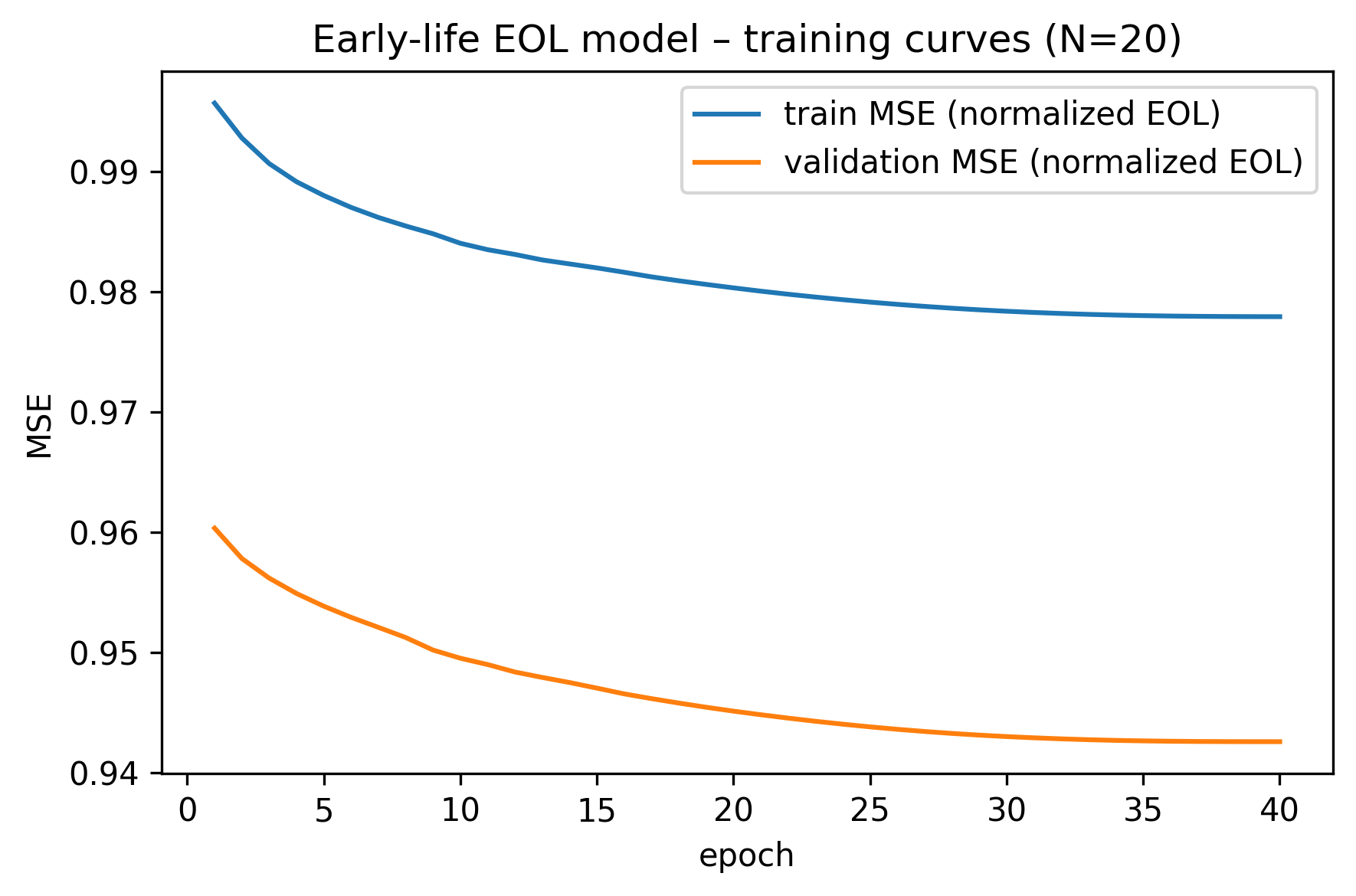}
		\label{fig:elog3}
	\end{subfigure}
	\hfill
	\begin{subfigure}[b]{0.48\textwidth}
		\centering
		\includegraphics[width=\textwidth]{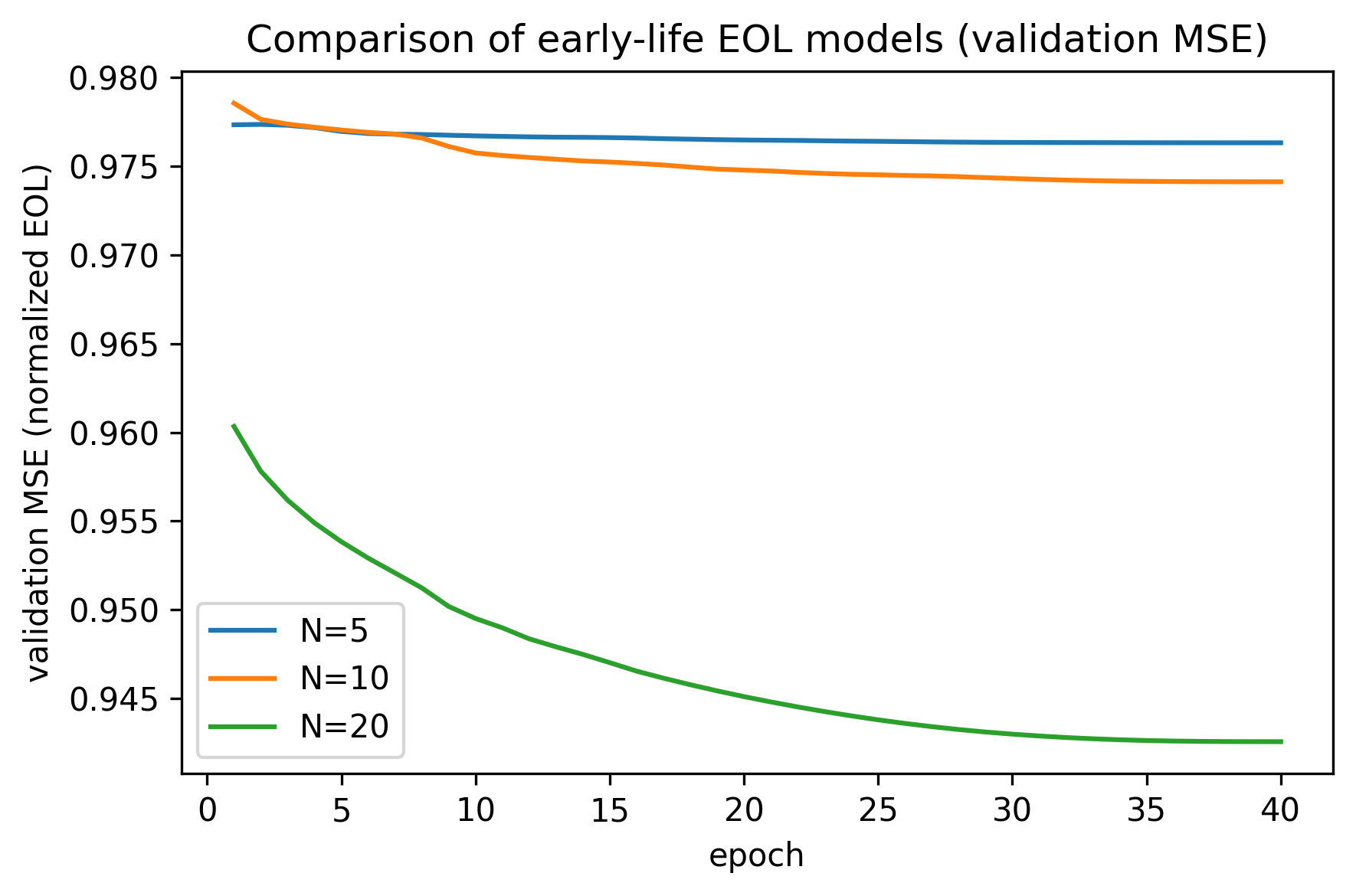}
		\label{fig:elo4}
	\end{subfigure}
	
	\caption{
		Training behaviour of the RUL/EOL regression model for different amounts of early-cycle input data.
		Panels (a-c) show training and validation MSE when using the first $N=\{5,10,20\}$ cycles as model input.
		Increasing the input window consistently reduces validation error, indicating that even a small number of early degradation observations contains predictive information about final EOL.
		Panel (d) summarises the validation MSE across all three settings, showing a monotonic improvement with larger~$N$ and demonstrating the benefit of extended early-life monitoring for RUL estimation.
	}
	
	\label{FigA:rul_training_dynamics}
\end{figure}

\begin{figure}[H]
	\centering
	\includegraphics[width=0.75\textwidth]{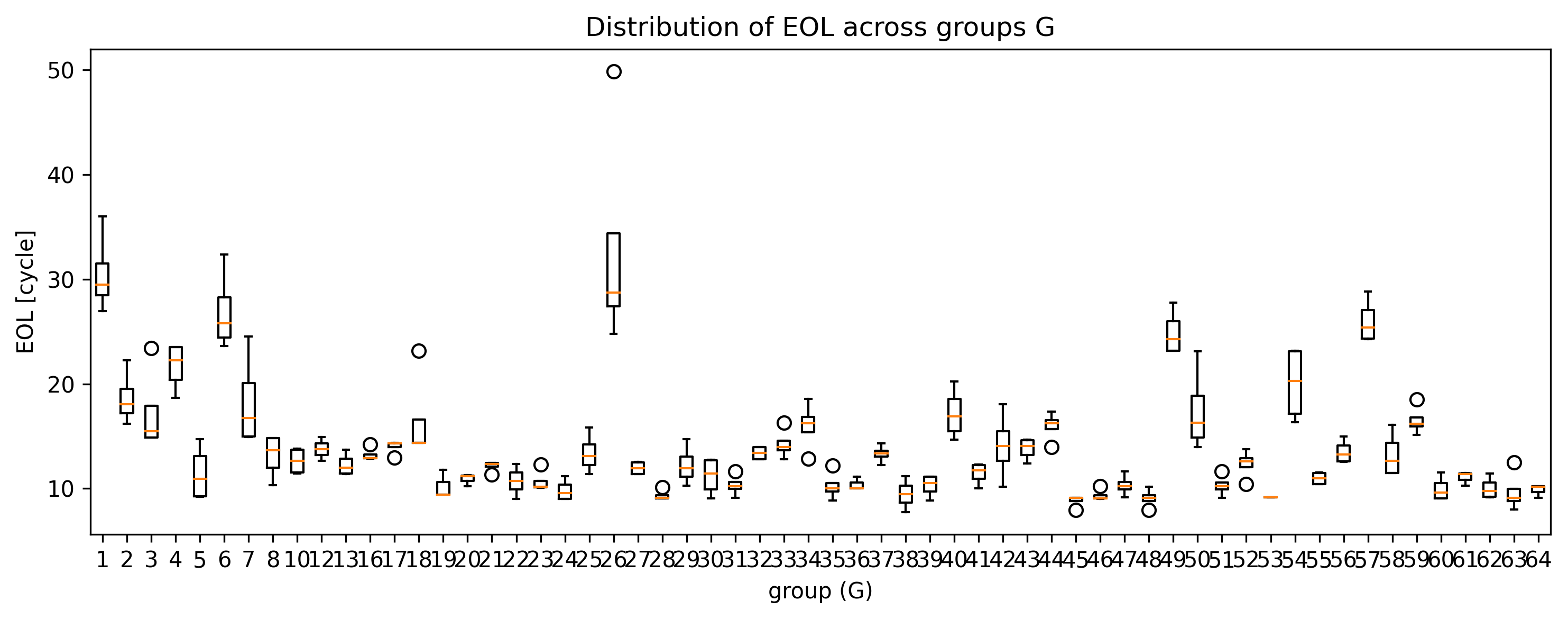}
	\caption{
		Distribution of EOL across electrode groups $G$ within dataset 22582234. Each box shows the cycle-to-failure variability for a specific manufacturing group or batch, revealing pronounced heterogeneity in ageing behaviour despite nominally identical test conditions. Several groups exhibit tight and low-variance EOL distributions, while others show wide spreads or isolated high-cycle outliers, suggesting differences in cell-to-cell quality, formation history or early-life conditioning. This variability highlights the importance of modelling EOL at the cell level rather than assuming a single characteristic lifetime for a given dataset or protocol.
	}
	
	\label{fig:EOL_group}
\end{figure}

\begin{figure}[H]
	\centering
	\begin{subfigure}[b]{0.32\textwidth}
		\includegraphics[width=\textwidth]{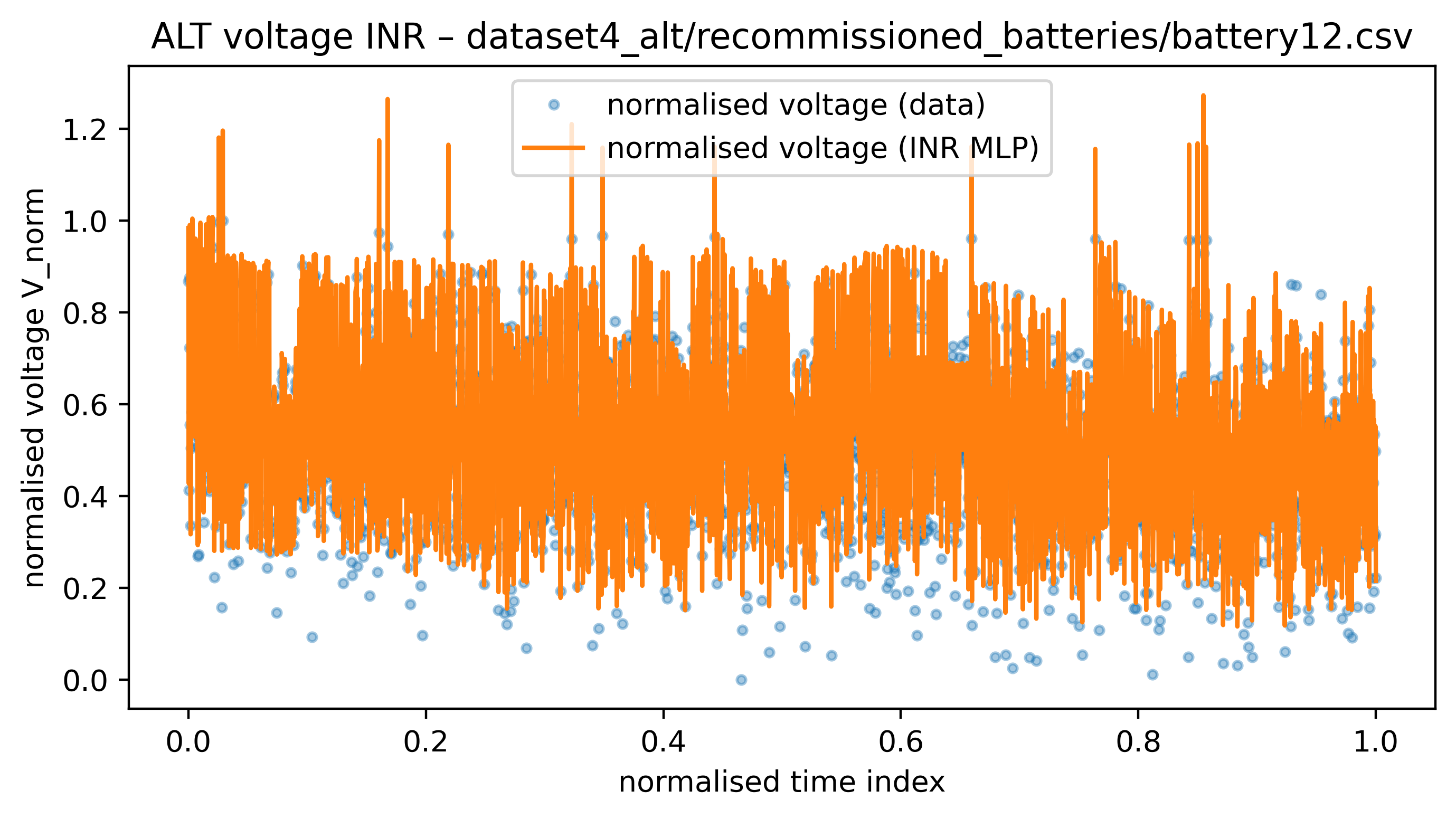}
		\caption{Recommissioned battery (battery12)}
	\end{subfigure}
	\hfill
	\begin{subfigure}[b]{0.32\textwidth}
		\includegraphics[width=\textwidth]{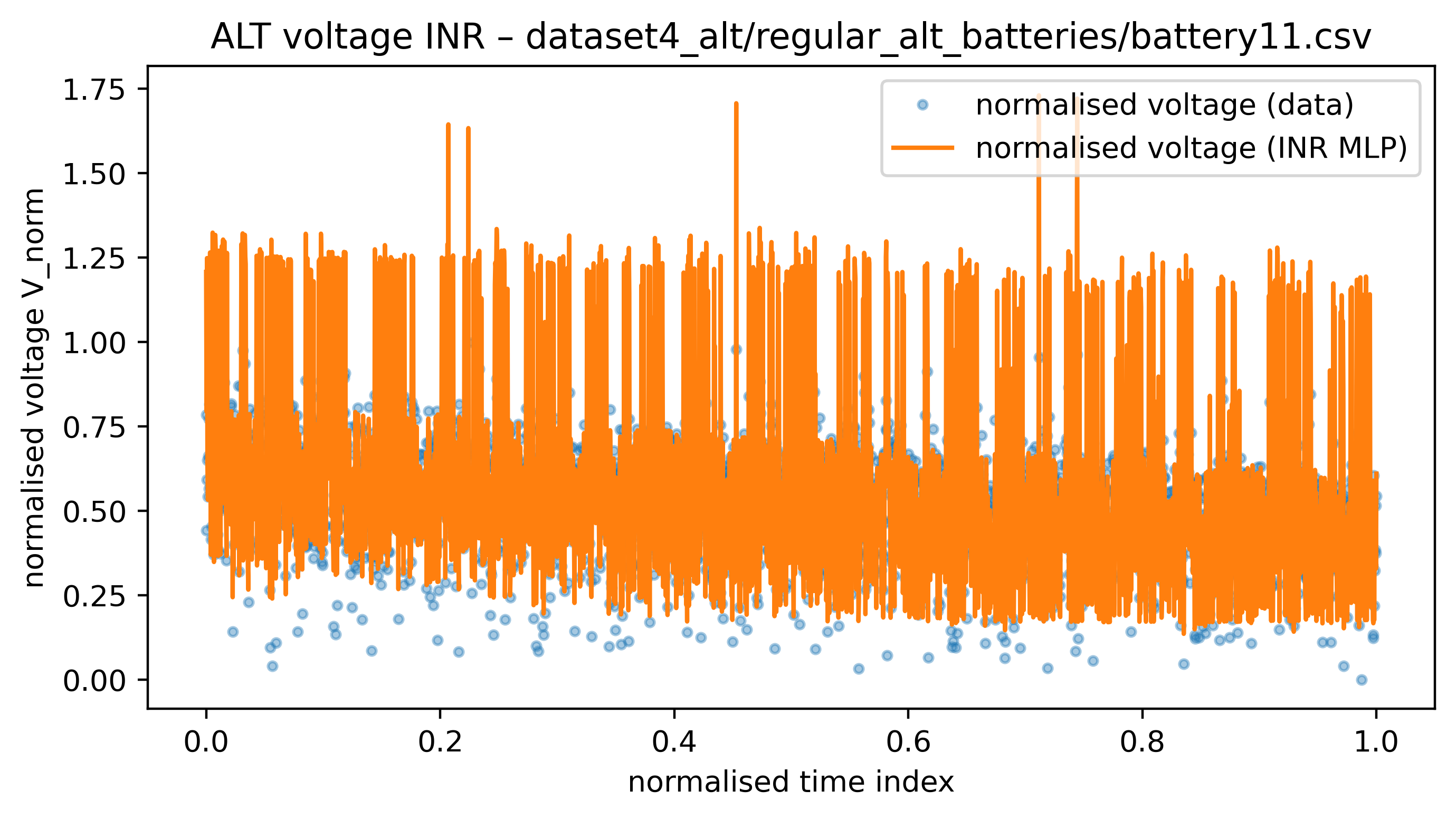}
		\caption{Regular ALT battery (battery11)}
	\end{subfigure}
	\hfill
	\begin{subfigure}[b]{0.32\textwidth}
		\includegraphics[width=\textwidth]{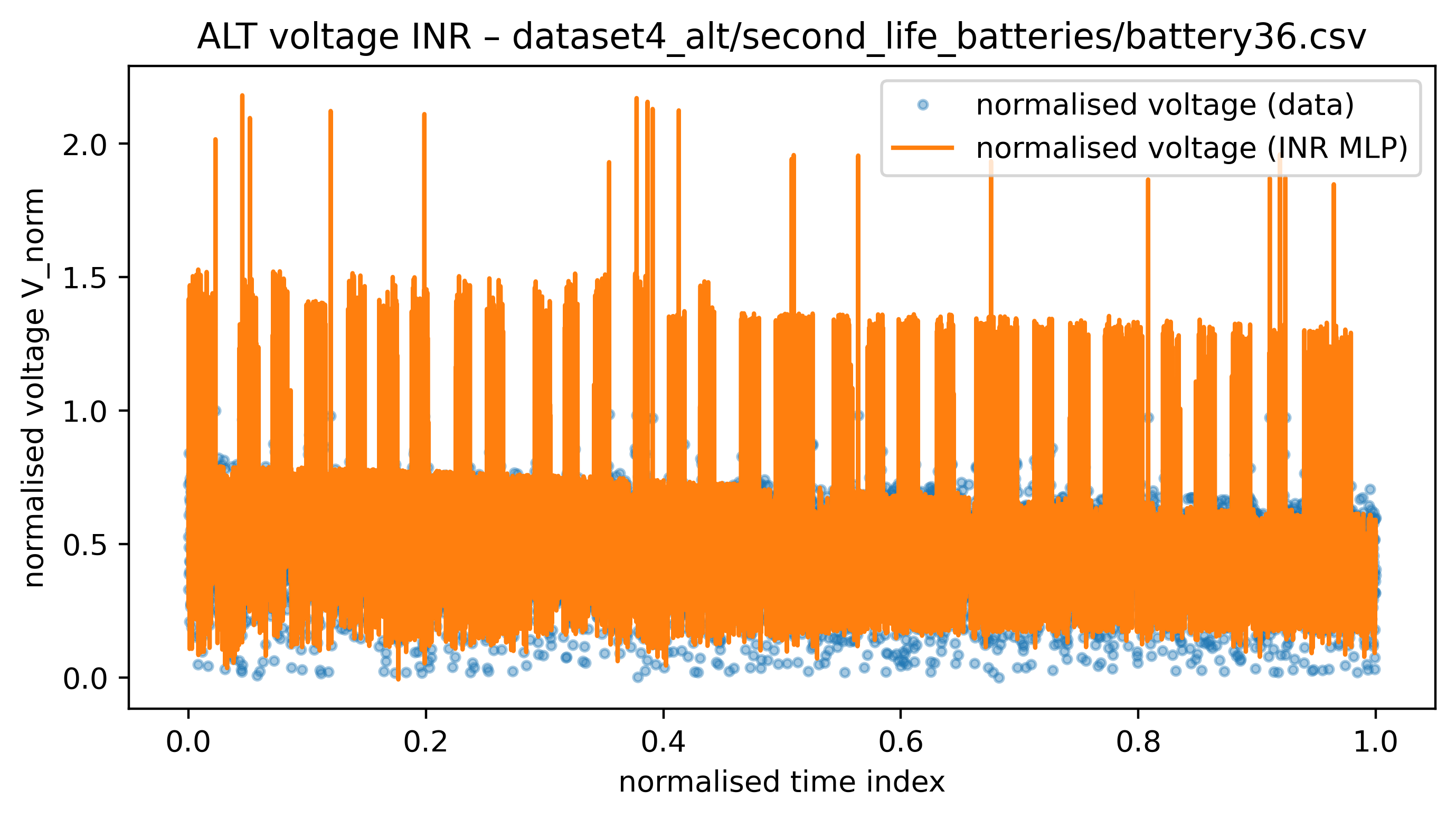}
		\caption{Second-life battery (battery36)}
	\end{subfigure}
	\caption{Representative voltage reconstruction examples using the INR–MLP model.
		The INR captures the underlying voltage pattern despite noise and irregularities.}
	\label{fig:ALT_voltage_examples}
\end{figure}


\begin{thebibliography}{90}

\bibitem[Hu, J et al.(2025)]{Hu2025}
Hu, J., Fu, P., Wei, Z. et al. Early prediction of lithium-ion battery degradation with a generative pre-trained transformer. Nat Commun 17, 126 (2026). https://doi.org/10.1038/s41467-025-66819-0.

\bibitem[von Bulow et al.(2023)]{vonBulow2023}
von Bulow, F., Hahn, Y., Meyes, R., Meisen, T. (2023) Transparent and Interpretable State of Health Forecasting of Lithium-Ion Batteries with Deep Learning and Saliency Maps, International Journal of Energy Research 9922475. https://doi.org/10.1155/2023/9922475.

\bibitem[Nazeeruddin et al.(2025)]{Nazeeruddin2025}
Nazeeruddin, M. A., Li, R., OKane, S. E., Marinescu, M., \& Offer, G. J. (2025). Lithium-ion battery degradation: Introducing the concept of reservoirs to design for lifetime. arXiv preprint https://arxiv.org/abs/2512.15440.

\bibitem[Guangxu et al. (2022)]{Guangxu2022}
Guangxu Zhang, Xuezhe Wei, Siqi Chen, Guangshuai Han, Jiangong Zhu, and Haifeng Dai
ACS Applied Energy Materials 2022 5 (5), 6462-6471
https://doi.org/10.1021/acsaem.2c00957.

\bibitem[Chen et al.(2023)]{Chen2023}
Chen, L., Xu, C., Bao, X. et al. (2023) State-of-health estimation of Lithium-ion battery based on back-propagation neural network with adaptive hidden layer. Neural Comput \& Applic 35, 14169-14182. https://doi.org/10.1007/s00521-023-08471-7.

\bibitem[Pregowska, A et al.(2022)]{Pregowska2022}
Pregowska, A., Osial, M., Urbanska, W. (2022) The Application of Artificial Intelligence in the Effective Battery Life Cycle in the Closed Circular Economy Model- A Perspective. Recycling 7, 81. https://doi.org/10.3390/recycling7060081.

\bibitem[Hu et al.(2012)]{Hu2012}
Hu, X., Li, S., Peng, H. (2012) A comparative study of equivalent circuit models for Li-ion batteries, Journal of Power Sources 198, 359-367. https://doi.org/10.1016/j.jpowsour.2011.10.013.

\bibitem[Jokar et al.(2016))]{Jokar2016}
Jokar, A., Rajabloo, B., Desilets, M., Lacroix, M. (2016) Review of simplified Pseudo-two-Dimensional models of lithium-ion batteries, Journal of Power Sources 327, 44-55. https://doi.org/10.1016/j.jpowsour.2016.07.036.

\bibitem[Wei et al.(2023)]{Wei2023}
Wei, Y., Wu, D. (2023) Prediction of state of health and remaining useful life of lithium-ion battery using graph convolutional network with dual attention mechanisms, Reliability Engineering \& System Safety 230, 108947. https://doi.org/10.1016/j.ress.2022.108947.

\bibitem[Song et al.(2025)]{Song2025}
Song, B., Yue, G., Guo, D., Wu, H., Sun, Y., Li, Y., Zhou, B. (2025) Prediction of the Remaining Useful Life of Lithium-Ion Batteries Based on Mode Decomposition and ED-LSTM,  Batteries 11, 86. https://doi.org/10.3390/batteries11030086.


\bibitem[dosReis et all.(2021)]{dosReis2021}
dos Reis, G., Strange, C., Yadav, M., Li, S. (2021) Lithium-ion battery data and where to find it, Energy and AI 5, 100081. https://doi.org/10.1016/j.egyai.2021.100081.

\bibitem[Song et al.(2024)]{Song2024}
Song, Z., Zhang, H., Jia, J. (2024) Data-Driven State of Health Interval Prediction for Lithium-Ion Batteries. Electronics 13, 3991. https://doi.org/10.3390/electronics13203991.

\bibitem[Mildenhall et al.(2022)]{mildenhall2022}
Mildenhall B., Srinivasan P.P., Tancik M., Barron J.T., Ramamoorthi R., Ng R. (2021). NeRF: representing scenes as neural radiance fields for view synthesis, Commun. ACM 65, 1, 99--106. https://doi.org/10.1145/3503250.

\bibitem[Diao et al.(2019)]{Diao2019}
Diao, W., Saxena, S., Han, B., Pecht, M. (2019) Algorithm to Determine the Knee Point on Capacity Fade Curves of Lithium-Ion Cells. Energies 12, 2910. https://doi.org/10.3390/en12152910.


\bibitem[You et al.(2023)]{You2023}
You, H., Zhu, J., Wang, X., Jiang, B., Wei, X., Dai, H. (2023) Nonlinear aging knee-point prediction for lithium--on batteries faced with different application scenarios,
eTransportation 18, 100270. https://doi.org/10.1016/j.etran.2023.100270.

\bibitem[Sitzmann et al.(2020)]{sitzmann2020}
Sitzmann V., Martel J., Bergman A., Lindell D., Wetzstein G. (2020). Implicit Neural Representations with Periodic Activation Functions, in: Advances in Neural Information Processing Systems, Larochelle H., Ranzato M., Hadsell R., Balcan M.F., Lin H. (Eds.), Curran Associates, Inc., Article 33, 7462--7473.

\bibitem[Tancik et al.(2020)]{tancik2020}
Tancik M., Srinivasan P. P., Mildenhall B., Fridovich-Keil S., Raghavan N., Singhal U., Ramamoorthi R, Barron J. T., Ng R., (2020). Fourier features let networks learn high frequency functions in low dimensional domains, in: Proceedings of the 34th International Conference on Neural Information Processing Systems (NIPS '20). Curran Associates Inc., Red Hook, NY, USA, Article 632, 7537--7547.

\bibitem[Mouais et al.(2021)]{Mouais2021}
Mouais, T., Kittaneh, O. A., Majid, M.A. (2021) Choosing the Best Lifetime Model for Commercial Lithium-Ion Batteries, Journal of Energy Storage 41, 102827. https://doi.org/10.1016/j.est.2021.102827.

\bibitem[Madani et al.(2025)]{Madani2025}
Madani, S.S., Shabeer, Y., Allard, F., Fowler, M., Ziebert, C., Wang, Z., Panchal, S., Chaoui, H., Mekhilef, S., Dou, S.X., et al. (2025) A Comprehensive Review on Lithium-Ion Battery Lifetime Prediction and Aging Mechanism Analysis. Batteries 11, 127. https://doi.org/10.3390/batteries11040127.

\bibitem[Park et al.(2019)]{park2019}
Park J.J., Florence P., Straub J., Newcombe R., Lovegrove S. (2019). DeepSDF: Learning Continuous Signed Distance Functions for Shape Representation, IEEE/CVF Conference on Computer Vision and Pattern Recognition (CVPR), Long Beach, CA, USA, pp. 165--174. https://doi.org/10.1109/CVPR.2019.00025.

\bibitem[Mescheder et al.(2019)]{mescheder2019}
Mescheder L., Oechsle M., Niemeyer M., Nowozin S., Geiger A. (2019). Occupancy Networks: Learning 3D Reconstruction in Function Space,IEEE/CVF Conference on Computer Vision and Pattern Recognition (CVPR), Long Beach, CA, USA, pp. 4455--4465. https://doi.org/10.1109/CVPR.2019.00459.

\bibitem[Pregowska et al. (2025)]{Pregowska2026}
Pregowska A., Larecki W., Szczepanski J. (2025) Application of Neural Networks for Determine the Radiation Pressure in Two-Moment Radiation Hydrodynamics in Slab Geometry, Computer Assisted Methods in Engineering and Science, 1--24. https://doi.org/10.24423/cames.2025.1960.


\bibitem[Dietrich et al.(2025)]{Dietrich2025}
Dietrich, F., Schilders, W. (2025) Scientific machine learning. Math Semesterber 72, 89--115. https://doi.org/10.1007/s00591-025-00399-4.


\bibitem[Meng et al.(2019)]{Meng2019}
Meng, H., Li, Y.F. A review on prognostics and health management (PHM) methods of lithium-ion batteries, Renewable and Sustainable Energy Reviews 116, 109405. https://doi.org/10.1016/j.rser.2019.109405.

\bibitem[Xiao et al.(2024)]{Xiao2024}
Xiao, Z., Jiang, B., Zhu, J., Wei, X., Dai, H. (2024) State of Health Estimation for Lithium-Ion Batteries Using an Explainable XGBoost Model with Parameter Optimization. Batteries 10, 394. https://doi.org/10.3390/batteries10110394.

\bibitem[Berecibar et al.(2016)]{Berecibar2016}
Berecibar, M., Gandiaga, I., Villarreal, I., Omar, N., Van Mierlo, J., Van den Bossche, P. (2016) Critical review of state of health estimation methods of Li-ion batteries for real applications," Renewable and Sustainable Energy Reviews, Elsevier, vol. 56(C), 572--587. https://doi.org/10.1016/j.rser.2015.11.042.


\bibitem[Attia et al.(2020)]{Attia2020}
Attia, P.M., Grover, A., Jin, N. et al. (2020) Closed-loop optimization of fast-charging protocols for batteries with machine learning. Nature 578, 397--402. https://doi.org/10.1038/s41586-020-1994-5.



\bibitem[Severson et al.(2019)]{Severson2019}
Severson, K.A., Attia, P.M., Jin, N. et al. (2019) Data-driven prediction of battery cycle  life before capacity degradation. Nat Energy 4, 383--391. https://doi.org/10.1038/s41560-019-0356-8.

\bibitem[Chen et al.(2018)]{Chen2018}
Chen, Z., Sun, M., Shu, X., Shen, J., Xiao, R. (2018). On-board state of health estimation for lithium-ion batteries based on random forest. 2018 IEEE International Conference on Industrial Technology (ICIT), 175--1759.


\bibitem[Al-Rahamneh et al.(2025)]{Al-Rahamneh2025}
Al-Rahamneh, A., Izco, I., Serrano-Hernandez, A., Faulin, J. (2025) Machine Learning-Based State-of-Health Estimation of Battery Management Systems Using Experimental and Simulation Data. Mathematics 13, 2247. https://doi.org/10.3390/math13142247.

\bibitem[Thelen et al.(2024)]{Thelen2024}
Thelen, A., Huan, X., Paulson, N. et al. Probabilistic machine learning for battery health diagnostics and prognostics-review and perspectives. npj Mater. Sustain. 2, 14. https://doi.org/10.1038/s44296-024-00011-1.

\bibitem[Gal et al.(2016)]{Gal2016}
Gal, Y., Ghahramani, Z. (2016) Dropout as a Bayesian approximation: representing model uncertainty in deep learning. In Proceedings of the 33rd International Conference on International Conference on Machine Learning 48 (ICML'16). JMLR.org, 1050--1059.

\bibitem[Kendall et al.(2017)]{Kendall2017}
Kendall, A., Gal, A. (2017) What uncertainties do we need in Bayesian deep learning for computer vision? In Proceedings of the 31st International Conference on Neural Information Processing Systems (NIPS'17). Curran Associates Inc., Red Hook, NY, USA, 5580--5590.

\bibitem[Xuan et al.(2023)]{Xuan2023}
Xuan, Q.L., Adhisantoso, Y.G., Munderloh, M., Ostermann, J. (2023) Uncertainty-aware remaining useful life prediction for predictive maintenance using deep learning,
Procedia CIRP 118, 116--12. https://doi.org/10.1016/j.procir.2023.06.021.


\bibitem[Breiman et al.(1984)]{Breiman1984}
Breiman, L., Friedman, J., Olshen, R.A., Stone, C.J. (1984). Classification and Regression Trees (1st ed.). Chapman and Hall/CRC. https://doi.org/10.1201/9781315139470.

\bibitem[Kuleshov et al.(2018)]{Kuleshov2018}
Kuleshov, V., Fenner, N., Ermon, S. (2018). Accurate Uncertainties for Deep Learning Using Calibrated Regression. ArXiv, abs/1807.00263.


\bibitem[Wei et al.(2021)]{Wei2021}
Wei, M., Gu, H., Ye, M., Wang, Q., Xu, X., Wu, C. (2021) Remaining useful life prediction of lithium-ion batteries based on Monte Carlo Dropout and gated recurrent unit, Energy Reports 7, 2862--2871. https://doi.org/10.1016/j.egyr.2021.05.019.

\bibitem[Attia et al.(2022)]{Attia2022}
Attia, P.M., Bills, A., Planella, F.B., Dechent, P., Dos Reis, G., Dubarry, M., Gasper, P., Gilchrist, R., Greenbank, S., Howey, D. and Liu, O. (2022). “Knees” in lithium-ion battery aging trajectories. Journal of The Electrochemical Society, 169(6), https://doi.org/060517.10.1149/1945-7111/ac6d13

\bibitem[Jia et al.(2024)]{Jia2024}
Jia, X., Zhang, C., Li, Y. et al (2024). Knee-point-conscious battery aging trajectory prediction of lithium-ion based on physics-guided machine learning. IEEE Transactions on Transportation Electrification, 10(1): 1056-1069.
http://dx.doi.org/10.1109/TTE.2023.3266386

\bibitem[Dickson et al.(2019)]{Dickson2019}
D. N. T. How, M. A. Hannan, M. S. Hossain Lipu and P. J. Ker, State of Charge Estimation for Lithium-Ion Batteries Using Model-Based and Data-Driven Methods: A Review, in IEEE Access, vol. 7, pp. 136116-136136, 2019, https://doi.org/10.1109/ACCESS.2019.2942213.

\bibitem[Lu et al.(2023)]{Lu2023}
Lu, Yu and Zhou, Sida and Zhou, Xinan and Yang, Shichun and Liu, Mingyan and Liu, Xinhua and Ling, Heping and Lian, Yubo. (2023). A novel method of prediction for capacity and remaining useful life of lithium-ion battery based on multi-time scale Weibull accelerated failure time regression. Journal of Energy Storage, 68, 107589. https://doi.org/10.1016/j.est.2023.107589.

\bibitem[Feng et al.(2018)]{Feng2018}
Feng, X., Weng, C., He, X., Wang, L., Ren, D., Lu, L., Han, X., Ouyang, M. (2018) Incremental Capacity Analysis on Commercial Lithium-Ion Batteries using Support Vector Regression: A Parametric Study. Energies 11, 2323. https://doi.org/10.3390/en11092323.

\bibitem[Lv et al. (2025)]{Lv2025}
Lv, Zhe and Si, Huinan and Yang, Zhe and Cui, Jiawen and He, Zhichao and Wang, Lei and Li, Zhe and Zhang, Jianbo. (2025). Simplified Mechanistic Aging Model for Lithium Ion Batteries in Large-Scale Applications. Materials, 18(6), 1342. https://doi.org/10.3390/ma18061342.

\bibitem[Bustos et al.(2025)]{Bustos2025}
Bustos, J.E.G., Schiele, B.B., Baldo, L., Masserano, B., Jaramillo-Montoya, F., Troncoso-Kurtovic, D., Orchard, M.E., Perez, A., Silva, J.F. (2025) In Situ Estimation of Li-Ion Battery State of Health Using On-Board Electrical Measurements for Electromobility Applications. Batteries 11, 451. https://doi.org/10.3390/batteries11120451. 

\bibitem[Lee et al.(2022)]{Lee2022}
Lee, G., Kim, J., Lee, C. (2022) State-of-health estimation of Li-ion batteries in the early phases of qualification tests: An interpretable machine learning approach. Expert Syst. Appl. 197, C. https://doi.org/10.1016/j.eswa.2022.116817.

\bibitem[Xu et al.(2025)]{Xu2025}
Xu, D., Ma, S., Ji, X., Han, X., Lin, J. and Liu, K. (2025), A SHapley Additive exPlanations-Based Data-Driven Approach for Lithium-Ion Battery State of Health Estimation Using Ultrasound Technology. Energy Technol 13, 2500673. https://doi.org/10.1002/ente.202500673.

 \bibitem[NASA (2025)]{NASA}
https://data.nasa.gov/dataset/li-ion-battery-aging-datasets

\bibitem[CALCE (2025)]{CALCE}
https://calce.umd.edu/battery-data

\bibitem[ISUILC (2023)]{ISUILCC}
https://iastate.figshare.com/articles/dataset/\_b\_ISU-ILCC\_Battery\_Aging\_Dataset\_b\_/22582234

\bibitem[ISUILCa (2023))]{ISUILCCa}
https://doi.org/10.25380/iastate.22582234

\bibitem[ChemDataExtractor (2020)]{ChDF}
https://doi.org/10.6084/m9.figshare.11888115

\bibitem[Rashid et al.(2023)]{Rashid2023}
Rashid, M., Faraji-Niri, M., Sansom, J., Sheikh, M., Widanage, D., Marco, J. (2023) Dataset for rapid state of health estimation of lithium batteries using EIS and machine learning: Training and validation. Data Brief. 48, 109157. doi: 10.1016/j.dib.2023.109157.

\bibitem[Hu et al.(2024)]{Hu2024}
Hu, W., Qian, Q. (2024)Lithium-ion battery state of health and failure analysis with mixture weibull and equivalent circuit model. iScience 27(6), 109980. https://doi.org/10.1016/j.isci.2024.109980. 

\bibitem[Chu et al.(2020)]{Chu2020}
Chu, Ch.-H., Lee, Ch.-J., Yeh, H.-Y. (2020) Developing Deep Survival Model for Remaining Useful Life Estimation Based on Convolutional and Long Short-Term Memory Neural Networks, Wireless Communications and Mobile Computing 8814658. https://doi.org/10.1155/2020/8814658. 

\bibitem[Huang et al.(2024)]{Huang2024}
Huang, Y., Zhang, P., Lu, J., Xiong, R., \& Cai, Z. (2024). A transferable long-term lithium-ion battery aging trajectory prediction model considering internal resistance and capacity regeneration phenomenon. Applied Energy, 360, 122825.

\bibitem[Liu et al.(2022)]{Liu2022}
Liu S, Shui J. (2022) Mechanism and properties of emerging nanostructured hydrogen storage materials. Battery Energy 1, 20220033. https://doi.org/10.1002/bte2.20220033.
 
  
\end{thebibliography}
\end{document}